%% file: main.tex
\documentclass{article}

\usepackage{times}
\usepackage[letterpaper,textwidth=5.5in,textheight=9in,centering]{geometry}
\usepackage[authoryear,round]{natbib}
\setcitestyle{citesep={;},aysep={,},yysep={;}}
\usepackage{essh_preprint}

\input{math_commands.tex}

\makeatletter
\@ifpackageloaded{amsmath}{}{\usepackage{amsmath}}
\@ifpackageloaded{amssymb}{}{\usepackage{amssymb}}
\@ifpackageloaded{booktabs}{}{\usepackage{booktabs}}
\@ifpackageloaded{array}{}{\usepackage{array}}
\@ifpackageloaded{multirow}{}{\usepackage{multirow}}
\@ifpackageloaded{graphicx}{}{\usepackage{graphicx}}
\@ifpackageloaded{xcolor}{}{\usepackage{xcolor}}
\@ifpackageloaded{algorithm2e}{}{\usepackage[ruled,vlined,linesnumbered]{algorithm2e}}
\usepackage{enumitem}
\makeatother

\usepackage{hyperref}
\usepackage{url}
\usepackage{soul}

\SetAlgoNlRelativeSize{0}
\SetNlSty{textbf}{\scriptsize}{}

\providecommand{\ours}{ESSH}

\providecommand{\refpub}{$^{\dagger}$}    %
\providecommand{\refspec}{$^{\diamond}$}   %

\providecommand{\refhyb}{$^{\star}$}
\definecolor{revblue}{HTML}{000000}
\newcommand{\rev}[1]{#1}

\definecolor{revred}{HTML}{000000}
\newcommand{\revr}[1]{#1}
\newenvironment{revblockr}{}{}
\newcommand{\editblue}[1]{#1}
\newenvironment{editblock}{}{}
\newcommand{\revcaption}[2]{\caption{#1}\label{#2}}
\newcommand{\revcaptionr}[2]{\caption{#1}\label{#2}}
\newcommand{\revfigcaptionr}[2]{\caption{#1}\label{#2}}
\newcommand{\revsubsection}[2]{\subsection{\editblue{#1}}\label{#2}}


\title{Elastic Selective Spectral Hybrids for \\ Train-Once, Export-Many Budgeted Inference}

\author{Dachuan Song\textsuperscript{1}\quad
Chuchu Chen\textsuperscript{2}\quad
Xuan Wang\textsuperscript{1,*}\\[0.5em]
\small\textsuperscript{1}George Mason University\\
\small\textsuperscript{2}George Washington University\\[0.3em]
\small\textsuperscript{*}Corresponding author:
\href{mailto:xwang64@gmu.edu}{\texttt{xwang64@gmu.edu}}}
\date{}

\usepackage{placeins}
\hypersetup{hidelinks,
  pdftitle={Elastic Selective Spectral Hybrids for Train-Once, Export-Many Budgeted Inference},
  pdfauthor={Dachuan Song, Chuchu Chen, Xuan Wang},
  pdfsubject={Elastic selective spectral models for budgeted inference},
  pdfkeywords={state space models, spectral filtering, elastic neural networks, budgeted inference}}
\begin{document}

\maketitle

\begin{abstract}
Deploying a language model under different computing and latency budgets calls for compact models with different quality--cost trade-offs. Towards this end, elastic spectral state space models provide ordered and temporally decomposed channels that can be truncated, but they are associated with linear time-invariant filters that cannot selectively preserve relevant past information or forget irrelevant information as the context evolves. To address this, we introduce the Elastic Selective Spectral Hybrid (ESSH), which realizes each Hankel spectral channel as an independent recurrent unit using fitted damped rotation modes. It also features an input-dependent decay and write/read gates that make temporal retention and state update input-dependent while preserving channel-wise truncation and a structured recurrence for efficient execution. ESSH combines these selective spectral mixers with sliding-window attention and jointly trains multiple capacities by reducing spectral-channel count and feed-forward width at different rates through a two-rate capacity map with full-model distillation. The resulting models support chunked parallel training and fused recurrent decoding while avoiding computation for discarded channels. \rev{A single 1.53B training run yields ten deployable models from 0.47B to 1.53B parameters, and the full model’s seven-task accuracy is within 0.4–2.6 points of independently trained models of the same size.} \rev{Smaller exports exhibit a smooth quality--size trade-off.} We validate the effectiveness of the proposed framework using language understanding, retrieval, cross-domain text, and DNA experiments, by assessing quality retention and the trade-off against independently trained and elastic baselines. \revr{Under one compilation and CUDA-graph policy at 1.53B, fused ESSH decoding has the lowest latency among the tested models on B300, with speedups of $1.12$--$1.45\times$ (batch sizes 1--512) over Mamba-2, $1.2$--$1.6\times$ over Mamba-3 SISO/MIMO and $1.2$--$6.6\times$ over Transformer++.} Code is available at \url{https://github.com/songdc98/ESSH}.

\end{abstract}

\section{Introduction}
\label{sec:intro}

Language models are deployed under different computing and latency limits. A model that fits one deployment setting may be too expensive in another. This motivates the need for models with different sizes that can operate under different deployment budgets. Training an independent model for every deployment budget is costly. One approach is to start from a pretrained model and obtain smaller models through post-training compression or adaptation~\citep{muralidharan2024minitron,taghibakhshi2025minitronssm}. However, different deployment sizes may require additional size-specific pruning, adaptation, or distillation. Instead, elastic models jointly train nested subnetworks of different capacities within a single training procedure~\citep{yu2019slimmable,cai2020ofa}.

A key design choice in elastic models is how model capacity should be reduced as the deployment budget decreases. Existing methods define capacity through different internal dimensions. MatFormer varies the width of feed-forward blocks~\citep{devvrit2024matformer}, while MatMamba varies the inner width and corresponding number of heads within the recurrent mixer~\citep{shukla2024matmamba}. A complementary approach is to define capacity through the temporal basis used for sequence mixing~\citep{NEURIPS2019_952285b9}. Compared with reducing generic internal dimensions, this directly controls the temporal components used to represent the input history. For example, Hankel spectral models provide an ordered spectral basis, making the number of retained channels a structured axis for elastic recurrent models~\citep{hazan2017spectral,agarwal2023spectral}.

Existing elastic spectral state space models (ES-SSM) exploit this ordering by varying the number of retained spectral channels across deployment budgets~\citep{song2026elastic}. However, their temporal filters remain linear time-invariant (LTI), which limits their ability to adapt how information is retained over time to the input content. In particular, 
the resulting temporal dynamics cannot selectively preserve relevant past information or forget irrelevant information as the context evolves. We therefore seek a new selective spectral recurrence that preserves the elastic spectral structure while making state update and retention input-dependent and keeping the recurrence efficient.

To do so, we represent each Hankel channel using damped rotation modes and introduce input-dependent mechanisms for controlling state retention. The resulting selective spectral mixer preserves independently truncatable channels and supports efficient parallel training and recurrent decoding. To further enhance model capability, we incorporate these mixers into a hybrid architecture combining state-space modeling with attention, and jointly train multiple capacities by varying both spectral-channel count and feed-forward width. Our contributions are:

\begin{itemize}[leftmargin=10pt]

\item \textbf{Truncatable selective spectral channels.}
We realize each Hankel spectral channel as an independent recurrent unit built from fitted damped rotation modes, with input-dependent decay and write/read gates. This allows state updates to be input-dependent while preserving whole-channel truncation and the structured transition for efficient execution.

\item \textbf{Two-rate elastic training: spectral + feed-forward.}
We jointly vary spectral-channel count and feed-forward width at different rates through a two-rate capacity map and train the resulting models with full-model distillation.
\revr{This widens the reduction range to smaller deployable models while encouraging compact subnetworks to maintain model quality across deployment budgets.}

\item \textbf{Efficient parallel training and recurrent decoding.}
To improve computational efficiency, we derive a chunked parallel scan for the selective recurrence and implement GPU kernels for forward and backward passes. During decoding, we further use a fused recurrent path that combines the projections, state updates, and readouts, thereby avoiding computation for discarded channels.

\item \textbf{Evaluation across quality, elasticity, and deployment cost.}
We evaluate full-model quality, performance across capacities, and deployment cost against independently trained and elastic baselines. Across language modeling, retrieval, longer-context, cross-domain, and DNA experiments, targeted ablations further examine the effects of initialization, selective dynamics, elastic training, and the implementation.
\end{itemize}

\section{Related Work}
\label{sec:related}

\paragraph{Train-once model families.}
Slimmable Networks and Once-for-All share parameters across subnetworks with different widths, depths, or resolutions, avoiding a separate training run for every deployment size~\citep{yu2019slimmable,yu2019universally,cai2020ofa}.
However, a smaller subnetwork retains only part of the shared computation, so the retained parameters must remain effective when evaluated independently.
Nested dropout and Matryoshka representation learning address this by encouraging leading feature dimensions to carry useful task information~\citep{rippel2014nested,kusupati2022mrl}.
MatFormer applies this idea to feed-forward width, while MatMamba applies it to the inner dimensions of Mamba-2 blocks~\citep{devvrit2024matformer,shukla2024matmamba}.
Matryoshka mixture-of-experts extends nesting to expert sets, and Flextron introduces elastic choices into a pretrained Transformer~\citep{wang2025matryoshkamoe,cai2024flextron}.
\revr{ESSH applies this train-once principle to independently truncatable spectral-channel groups and a separately controlled feed-forward axis. Hankel initialization supplies the initial temporal responses. Capacity-mixed training jointly trains the full-capacity and reduced-capacity predictors.}

\paragraph{Selective and spectral state space models.}
Mamba makes recurrent state update input-dependent, while Mamba-2 connects selective state-space computation to structured attention%
~\citep{gu2024mamba,dao2024mamba2}.
Mamba-3 further develops the discretization and complex-valued dynamics~\citep{lahoti2026mamba3}.
Gated DeltaNet combines gating with delta-rule state updates~\citep{ICLR2025_4904fad1}.
\revr{Hankel spectral methods approximate linear dynamics with ordered temporal filters}~\citep{hazan2017spectral,agarwal2023spectral,liu2025flashstu}.
Elastic spectral SSMs use this ordering to vary channel count across deployment budgets~\citep{song2026elastic}, but their temporal filters remain linear time-invariant and cannot adapt state retention to the input content.
\revr{LRU uses a learned, input-independent complex-diagonal transition~\citep{orvieto2023lru}. HGRN adds input-dependent forgetting with layer-wise lower bounds and rotation phases shared across time~\citep{qin2023hgrn}. ESSH shares this per-mode transition form and, like HGRN, gates its readout, but it learns the write gate independently of retention instead of tying it to the forget gate. HGRN2 expands a real-valued state through outer products~\citep{qin2024hgrn2}. ESSH groups oscillatory modes into independently truncatable channels and uses Hankel initialization for their initial temporal responses. Each channel reuses one projected input across its modes for efficient recurrent decoding. MatMamba nests Mamba-2 inner dimensions and heads~\citep{shukla2024matmamba}. ESSH applies this nesting principle to spectral-channel groups, with a separate FFN reduction rate.}

\paragraph{Elastic hybrid models.}
Hybrid models such as Jamba and Zamba combine recurrent or SSM layers with attention, using recurrent computation for efficient sequence modeling and attention for stronger token interactions and retrieval~\citep{lieber2024jamba,glorioso2024zamba}.
Making such models elastic is harder because recurrent state, feed-forward layers, and attention contribute differently to model size and deployment cost.
Minitron-SSM prunes SSM groups together with feed-forward width, embedding width, and depth, followed by distillation~\citep{taghibakhshi2025minitronssm}.
Nemotron Elastic trains nested models within a pretrained model by jointly reducing SSM and feed-forward dimensions with multi-budget distillation~\citep{taghibakhshi2025nemotronelastic}.
Star Elastic extends elastic training to hybrid mixture-of-experts models and separate budgets for thinking and answering~\citep{taghibakhshi2026starelastic}.
These approaches introduce or refine elasticity after obtaining a pretrained parent model. In contrast, ESSH incorporates multiple deployment capacities directly into pretraining by varying both spectral-channel count and feed-forward width.

\section{Preliminaries}
\label{sec:preliminaries}

A linear state space model maps inputs $u(t)\in\mathbb R^{d_{\mathrm{in}}}$ to outputs $y(t)\in\mathbb R^{d_{\mathrm{out}}}$ through
\begin{equation}
h(t)=Ah(t-1)+Bu(t),\qquad y(t)=Ch(t)+Du(t),
\end{equation}
where $h(t)\in\mathbb R^N$, $A\in\mathbb R^{N\times N}$, $B\in\mathbb R^{N\times d_{\mathrm{in}}}$, $C\in\mathbb R^{d_{\mathrm{out}}\times N}$, and $D\in\mathbb R^{d_{\mathrm{out}}\times d_{\mathrm{in}}}$. With $h(0)=0$, its temporal kernel is $G(\tau)=CA^\tau B$ and
$y(t)=Du(t)+\sum_{\tau=0}^{t-1}G(\tau)u(t-\tau)$.

For symmetric $A$ with eigenvalues in $[0,1)$, 
Hankel spectral filtering represents the resulting temporal kernels using an ordered basis of temporal filters. Specifically, define the weighted geometric sequence $\mu(\lambda)=(1-\lambda)[1,\lambda,\ldots,\lambda^{L-1}]^\top$ and its second-moment matrix $Z=\int_0^1\mu(\lambda)\mu(\lambda)^\top\,\mathrm d\lambda$. Let $(\sigma_k,\phi_k)$ be the eigenpairs of $Z$, ordered by decreasing $\sigma_k$, with $\|\phi_k\|_2=1$. The first $K$ eigenvectors minimize mean squared projection error for this weighted single-pole family, which yields
\begin{equation}\textstyle
G(\tau)\approx\sum_{k=1}^{K}M_k\phi_k(\tau),\qquad
\hat y(t)=\sum_{k=1}^{K}M_k(\Phi_k*u)(t),
\label{eq:spectral_base}
\end{equation}
where $M_k\in\mathbb R^{d_{\mathrm{out}}\times d_{\mathrm{in}}}$ and $\Phi_k$ is the causal filter with samples $\phi_k(\tau)$. Derivations of this approximation are given in Appendix~\ref{app:theory-pca}. The direct term $Du(t)$ is kept separately.

Each term $M_k(\Phi_k*u)(t)$ defines one spectral channel. The ordering of the eigenpairs therefore provides a natural truncation axis: a full model uses $\bar K$ spectral channels, while a smaller model with $K<\bar K$ retains only the first $K$ channels and discards channels
$k>K$. However, the
spectral ordering alone does not guarantee that the important information is aggregated in the low-index channels (after
truncation). Thus, obtaining accurate smaller models requires explicitly
training the retained channel sets to maintain task performance.

Elastic spectral SSMs~\citep{song2026elastic} build on this
truncation rule by jointly training models with different retained channel
counts. They additionally introduce a tokenwise two-layer GELU MLP to produce gate logits $s(t)=\operatorname{GateMLP}(u(t))$. With $K$ retained channels, their output is
\begin{equation}\textstyle
\hat y^{(K)}(t)=\sum_{k=1}^{K}g_k^{(K)}(t)M_k(\Phi_k*u)(t),\qquad
 g_k^{(K)}(t)=\frac{1}{\sqrt K}\operatorname{sigmoid}(s_k(t)).
\label{eq:prior_elastic}
\end{equation}
The gate changes the contribution of a fixed filtered feature, but the filter still depends only on lag $\tau$. 
In the following, we introduce a selective mixer that preserves channel-wise truncation structure while making the temporal dynamics within each channel input-dependent.

\section{Method}
\label{sec:method}
The proposed method aims to address the following challenges: (i) make the temporal dynamics input-dependent within spectral channels,
(ii) maintain model performance across reduced-capacity tiers, and (iii) enable efficient parallel training and recurrent decoding. We refer to each model configuration with a specified spectral-channel count
and feed-forward width as a capacity tier.
Towards this end, we first realize each spectral channel as an independent recurrent unit with input-dependent state updates (Sec.~\ref{sec:selective_mixer}).
We then combine channel truncation with feed-forward width reduction and jointly train the resulting capacity tiers with distillation from the full-capacity tier (Sec.~\ref{sec:elasticity}).
Finally, we use the same structured recurrence for a chunked parallel scan during training and a fused recurrent path during decoding (Sec.~\ref{sec:efficient_execution}).

\begin{figure}[t]
\centering
\includegraphics[width=0.81\linewidth]{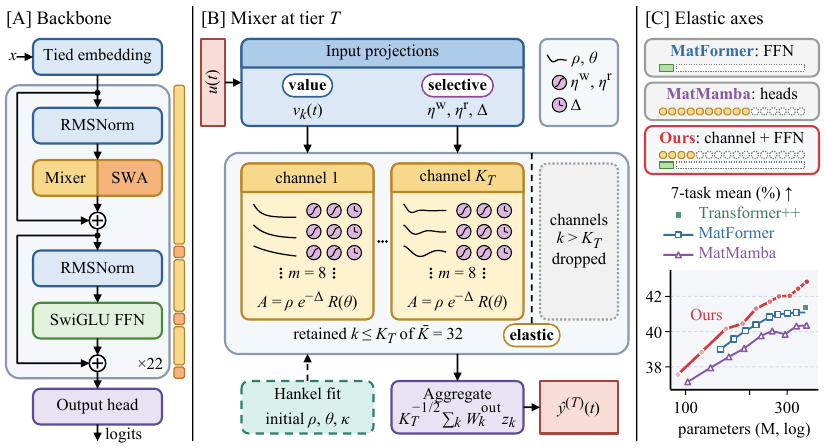}
\caption{
Overview of ESSH and its elastic axes.
(A) The hybrid backbone combines spectral mixers with three sliding-window attention layers.
(B) At tier $T$, the input produces channel values and token-dependent clock and write/read controls. Each retained spectral channel updates its $m$ recurrent modes; channels $k>K_T$ are removed before the outputs are aggregated.
(C) MatFormer varies feed-forward width, MatMamba varies mixer inner width, while ESSH varies both spectral-channel count and feed-forward width. The inset shows the quality--cost trade-off across capacities.
}
\label{fig:architecture}
\end{figure}

Figure~\ref{fig:architecture} represents the overall hybrid architecture:
Selective spectral mixers summarize inputs in recurrent states, while three sliding-window attention (SWA) layers provide direct access to recent token representations~\citep{beltagy2020longformer,jiang2023mistral}.
We vary capacity only inside the spectral mixers and feed-forward layers, and keep the residual width $d$ and attention dimensions fixed across capacity tiers.
The sliding-window configuration is specified in Appendix~\ref{app:setup-config}.

\subsection{Selective Spectral Mixer}
\label{sec:selective_mixer}
The output gate in \eqref{eq:prior_elastic} rescales a fixed filtered feature, but cannot change how an earlier input propagates through intervening positions.
To control this propagation within individual spectral channels, we approximate each channel using an independent recurrent realization of its fixed filter through damped rotation modes.
For each channel, a matrix-pencil fit~\citep{hua1990matrixpencil}
approximates the Hankel filter with $m$ damped rotation modes:
\begin{equation}\textstyle
\phi_k(\tau)\approx\sum_{i=1}^{m}\rho_{k,i}^{\,\tau}
\bigl(\kappa^{\mathrm c}_{k,i}\cos(\theta_{k,i}\tau)
-\kappa^{\mathrm s}_{k,i}\sin(\theta_{k,i}\tau)\bigr).
\label{eq:modal_fit}
\end{equation}
Each damped rotation mode admits a two-dimensional recurrent realization,
so this decomposition provides a compact recurrent realization of the fixed
spectral filter.
Here, $\rho_{k,i}\in(0,1)$ controls the intrinsic decay of the oscillator, $\theta_{k,i}$ is the rotation frequency, and $\kappa^{\mathrm c}_{k,i},\kappa^{\mathrm s}_{k,i}$ are mode readout coefficients.
The fitted parameters initialize the recurrence and remain trainable (with details in Appendix~\ref{app:modal_initialization}).
This recurrent realization approximates the temporal structure of the fixed
spectral filter, but its dynamics are still independent of the input content. 

To establish input dependence, output gating alone is not sufficient: it can suppress what is read at the current position, but cannot change what is propagated to future positions.
We therefore make the state transition input-dependent through its decay %
and introduce separate write and read gates for input injection and readout.
Let $u(t)\in\mathbb R^d$ be the normalized mixer input.
Each channel forms a value $v_k(t)\in\mathbb R^P$, shared by its $m$ modes, using the projection and channel-local normalization blend (details in Appendix~\ref{app:mixer-proj}).
Separate input projections produce clock increments $\Delta_{k,i}(t)>0$ and write/read gates $\eta^{\mathrm w}_{k,i}(t),\eta^{\mathrm r}_{k,i}(t)\in(0,2)$.
With mode state $H_{k,i}(t)\in\mathbb R^{2\times P}$, $e_1=(1,0)^\top$, and $\kappa_{k,i}=(\kappa^{\mathrm c}_{k,i},-\kappa^{\mathrm s}_{k,i})^\top$, each mode evolves as
\begin{equation}
\begin{aligned}
A_{k,i}(t)&=\rho_{k,i}\mathrm e^{-\Delta_{k,i}(t)}R(\theta_{k,i}),
\qquad H_{k,i}(0)=0,\\
H_{k,i}(t)&=A_{k,i}(t)H_{k,i}(t-1)
+\eta^{\mathrm w}_{k,i}(t)e_1v_k(t)^\top .
\end{aligned}
\label{eq:selective_mode}
\end{equation}
The planar rotation is
$R(\theta)=\left[\begin{smallmatrix}\cos\theta&-\sin\theta\\\sin\theta&\cos\theta\end{smallmatrix}\right]$.
The clock $\Delta_{k,i}(t)>0$ controls how much of the previous mode state is retained, the write gate controls how much of the current channel value enters the mode state, and the read gate controls how strongly that state contributes to the channel output.
Summing $\eta^{\mathrm r}_{k,i}(t)H_{k,i}(t)^\top\kappa_{k,i}$ over the $m$ modes gives the channel readout.
A zero-lag correction (Appendix~\ref{app:mixer-decay}) then gives the channel output $z_k(t)$.
Because rotation preserves the state norm, the clock acts through the decay factor
$\rho_{k,i}\mathrm e^{-\Delta_{k,i}(t)}<1$.
Larger clock increments suppress the previous mode state more strongly, while this factor approaches the intrinsic decay $\rho_{k,i}$ as the increment approaches zero.

A key design goal is to make the decay input-dependent while keeping the learned rotation frequency shared across positions.
Because scalar decay and planar rotation commute, propagation from position $s$ to $t$ retains the closed form
\begin{equation}
\prod_{\tau=s+1}^{t}A_{k,i}(\tau)
=\rho_{k,i}^{\,t-s}
\mathrm e^{-\sum_{\tau=s+1}^{t}\Delta_{k,i}(\tau)}
R\bigl(\theta_{k,i}(t-s)\bigr).
\label{eq:structured_transition}
\end{equation}
For $s=t$, the product is defined as the identity matrix.
In \eqref{eq:structured_transition}, intervening tokens modulate state retention while rotation depends only on token lag, preserving the structured transition product used for parallel training in Sec.~\ref{sec:efficient_execution}.
\revr{Each mode is a gated complex recurrence in real coordinates, as in HGRN~\citep{qin2023hgrn}. Its learned rotation angle is shared across time, while its decay depends on the input.}

Furthermore, because each channel computes its own value, clock, gates, and mode states from the shared normalized mixer input using channel-local operations, the channels remain independent.
If a truncated model keeps $1,\ldots,K$ channels and discards higher-indexed channels, the output becomes 
\[\textstyle
\hat y(t)=K^{-1/2}\sum_{k=1}^{K}W_k^{\mathrm{out}}z_k(t),
\qquad W_k^{\mathrm{out}}\in\mathbb R^{d\times P}.
\]
Channel truncation changes both the terms included in the sum and its
$K^{-1/2}$ normalization, so the truncated mixer does not generally produce
the same output as the full mixer. Next, we introduce methods to train these reduced-capacity models to maintain model performance after truncation.

\subsection{Two-Rate Elastic Training Across the Hybrid}
\label{sec:elasticity}

\textbf{Capacity tiers and model truncation.} 
We define a capacity tier $T$ by its retained spectral-channel count $K_T$ and feed-forward width $n_{\mathrm{ff},T}$.
These two quantities control different parts of deployment cost: the spectral-channel count determines recurrent-state size and mixer computation, whereas the feed-forward layer contains most of a spectral block's parameters (Table~\ref{tab:mixer-count}).
We reduce these two capacity axes at different rates, retaining
a larger fraction of spectral channels while reducing feed-forward width
more aggressively.
We encode the two rates with a budget variable $\xi_T\in(0,1]$:
\begin{equation}
K_T=\max\{1,\operatorname{round}(\bar K\xi_T^{1/2})\},\qquad
n_{\mathrm{ff},T}
=64\max\!\left\{1,\left\lfloor
\frac{d_{\mathrm{ff}}\xi_T^{0.8}}{64}
\right\rfloor\right\}.
\label{eq:two_rate}
\end{equation}
Here, $\bar K$ is the full spectral-channel count,
$K_T$ is the number of channels retained by tier $T$, $d_{\mathrm{ff}}$ is the full feed-forward width and $n_{\mathrm{ff},T}$ is its retained width.
At tier $T$, channel truncation retains spectral channels
$1,\ldots,K_T$ and discards channels $k>K_T$.
Similarly, each feed-forward layer retains the first
$n_{\mathrm{ff},T}$ units under the importance ordering.
The budget variable $\xi_T$ determines the retained tensor shapes rather than the exact exported parameter fraction.
Because $\xi_T^{1/2}>\xi_T^{0.8}$ for $0<\xi_T<1$, reduced-capacity tiers
retain a larger fraction of spectral channels than feed-forward units.
Since the feed-forward layer also contains most of a spectral block's
parameters, most of the parameter reduction occurs in the feed-forward
layers, while recurrent-state size and mixer computation decrease more
gradually. 
Attention, embeddings, and residual width do not vary with tier.
The exponents are empirical design choices evaluated in Appendix~\ref{app:el-nesting}.
Table~\ref{tab:el-map} gives the resulting model sizes.

\textbf{Capacity-mixed elastic training.} The two-rate map defines the spectral-channel count and feed-forward width of each tier, but it does not ensure that reduced-capacity models maintain model performance.
To address this, we introduce capacity-mixed training, in which each optimizer step combines full-capacity micro-batches with micro-batches assigned to sampled reduced capacities.
For a reduced-capacity micro-batch, the full-capacity tier $T_{10}$ is evaluated on the same input and supplies a stopped-gradient distillation target~\citep{hinton2015distillation}:
\begin{equation}
\mathcal L_T
=\mathrm{CE}(p_T,\mathrm{id}_{t+1})
+0.5\,\mathrm{KL}\bigl(
\operatorname{sg}[p_{T_{10}}]\,\|\,p_T
\bigr),
\label{eq:cross_capacity_distillation}
\end{equation}
where $p_T$ and $p_{T_{10}}$ are the next-token distributions produced by
tier $T$ and the full-capacity tier, respectively, and
$\operatorname{sg}[\cdot]$ denotes stop-gradient.
The cross-entropy (CE) term is the negative log-likelihood of the observed next token, while the KL divergence encourages the reduced-capacity distribution to match the full-capacity distribution. If $T=T_{10}$, the KL divergence is omitted.

Because each reduced-capacity tier retains channels $1,\ldots,K_T$,
lower-index channels are included in more sampled reduced-capacity
micro-batches.
Consequently, they are optimized under a broader range of reduced-capacity
configurations, which follows the nested-dropout principle
\citep{rippel2014nested,song2026elastic}.
For retained feed-forward weights, gradients from reduced-capacity
micro-batches are additionally scaled by $\alpha_{\mathrm{ff}}=0.5$,
which reduces their contribution to the shared parameter update without
changing the forward computation.
Together with the two-rate capacity map, our elastic-training recipe uses budget dropout (the capacity sampling above) over capacity tiers, full-tier distillation for reduced-capacity models, feed-forward importance ordering, and reduced-capacity feed-forward gradient scaling.
Appendices \ref{app:elastic-sort} and~\ref{app:elastic-mix} give the details for these. After pretraining and a brief shared refinement stage (Appendix~\ref{app:elastic-export}), each tier is exported as a standalone model by slicing its spectral-channel and feed-forward tensors.

\subsection{Efficient Parallel Training and Recurrent Decoding}
\label{sec:efficient_execution}

\begin{figure}[t]
\centering
\includegraphics[width=.85\linewidth]{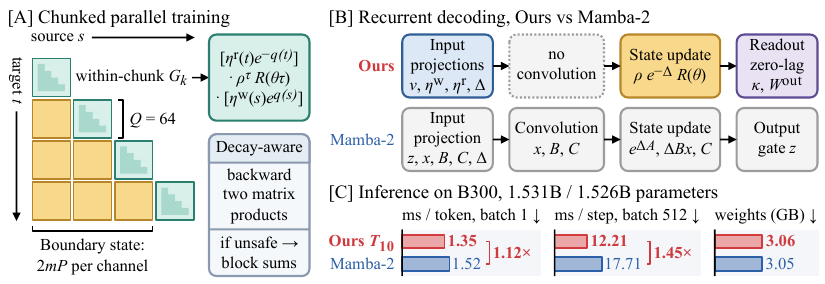}
\caption{Efficient execution of the selective mixer.
(A) A chunked parallel scan evaluates within-chunk contributions in parallel, while mode states connect consecutive chunks.
(B) Mixer-level recurrent decoding is compared with Mamba-2.
(C) Inference efficiency of ESSH versus parameter-matched Mamba-2 on B300\revr{, with torch.compile and CUDA graphs for both models}.
Full efficiency results: Appendices~\ref{app:systems_results} and~\ref{app:cross-device-final}.}
\label{fig:efficient_execution}
\end{figure}

The structured transition in \eqref{eq:structured_transition} supports both parallel training and recurrent decoding (Figure~\ref{fig:efficient_execution}) using the same selective recurrence.

\textbf{Chunked parallel training.}
During training, we divide each input sequence into contiguous chunks of at most $Q=64$ positions.
Within a chunk, the closed-form transition in
\eqref{eq:structured_transition} allows the contributions between all
source and target positions to be evaluated in parallel.
\rev{For channel $k$, stacking the chunk's channel values into $V_k$ gives the readouts $Z_k^{\mathrm{core}} = G_k V_k + B_k$. Here, $G_k$ is the causal within-chunk matrix and $B_k$ is the contribution from the entering mode states.}
Thus, the within-chunk computation is performed by matrix multiplication,
while only $2mP$ recurrent-state entries per channel are carried between
consecutive chunks~\citep{dao2024mamba2}.
The exact construction of $G_k$, the boundary-state update, and the backward
pass are given in Appendix~\ref{app:sys-fwd}--\ref{app:sys-bwd}.

\textbf{Recurrent decoding.}
During recurrent decoding, the same mode states are updated one token
at a time.
We use a fused recurrent path that combines the retained projections, state
updates, and readouts, without evaluating discarded channels.
At tier $T$, each spectral layer stores $2K_TmP$ recurrent-state entries,
while sliding-window attention contributes a fixed-size key--value cache.

\FloatBarrier
\section{Experiments}
\label{sec:experiments}

We evaluate ESSH on model quality, elastic quality--size trade-offs, and inference efficiency. 
We use a 1.5B-scale study to assess large-scale quality and deployment performance. The model is pretrained on 100B FineWeb-Edu tokens with the Llama-3.1 tokenizer~\citep{penedo2024fineweb,grattafiori2024llama3}. 
We also use a 370M-scale study for controlled comparisons with elastic baselines, independently trained models, and ablations. It uses 7.4B FineWeb-Edu tokens with the GPT-2 tokenizer. One-change ablations use 3.7B tokens.
Both models use a training context of 2048 tokens. Appendix~\ref{app:experimental_setup} gives configurations and protocols. Training costs are reported in Appendices~\ref{app:comp-amort} and~\ref{app:comp-external-cost}.
Evaluation includes FineWeb-Edu perplexity, a seven-task zero-shot language-understanding suite, NIAH retrieval, and a 2K--16K context-length sweep. We measure the training cost under four deployment sizes and benchmark decode latency and throughput, prefill throughput, and persistent state+KV memory.

\begin{table}[t]
\centering\small
\setlength{\tabcolsep}{1.5pt}
\renewcommand{\arraystretch}{1.0}
\caption{\editblue{Quality and inference cost: (a) full 1.5B models; (b) ESSH deployment tiers; (c) 370M models trained on 7.4B tokens; its lower block trains ESSH and the original ES-SSM mixer in the same 370M hybrid at full size only, on 3.7B tokens; (d) seven-task means of the $T_4$ and $T_1$ exports from (b) versus published models trained separately at $\mathord{\sim}$0.88B and $\mathord{\sim}$0.44B (TF++: Transformer++, GDN: Gated DeltaNet, M3: Mamba-3). Panels (a,b,d) use 100B FineWeb-Edu pretraining tokens. $\dagger$: published results from Table~1 of \citet{lahoti2026mamba3}. PPL in (a) uses each source's validation shard and is not ranked. Task scores are percentages in (a,b,d) and fractions in (c); retrieval uses 2K contexts. In (d), parentheses give baseline minus ESSH in percentage points; full task results are in Table~\ref{tab:compact-published-tasks}. ESSH retains 0.26B tied-embedding parameters. Inference in (b): B300, 2K context, bf16 weights, torch.compile and CUDA graphs; batch one unless marked. Bold compares task scores within (a) and full-size scores within the 7.4B block of (c). Protocols: Appendix~\ref{app:experimental_setup}.}}
\label{tab:exp-1p5b}
\label{tab:new-quality-latency-1p5b}
\label{tab:exp-370-summary}
\begin{tabular*}{\linewidth}{@{\extracolsep{\fill}}lrrrrrrrrrr@{}}
\toprule
\textbf{(a) Model} & Params (B) & PPL $\downarrow$ & LAMB & HellaS & PIQA & ARC-e & ARC-c & Wino & OBQA & Mean \\
\midrule
\textbf{ESSH} $T_{10}$ & 1.53 & 10.11 & 46.8 & 60.6 & 72.9 & 73.7 & 40.7 & 58.3 & 32.2 & 55.0 \\
Transformer++\refpub & $\mathord{\sim}1.5$ & 10.51 & 50.3 & 60.6 & 73.8 & 74.0 & 40.4 & 58.7 & 29.6 & 55.4 \\
Gated DeltaNet\refpub & $\mathord{\sim}1.5$ & 10.45 & 49.2 & 61.3 & 74.3 & 75.3 & 41.2 & 58.0 & 31.6 & 55.8 \\
Mamba-2\refpub & $\mathord{\sim}1.5$ & 10.47 & 47.8 & 61.4 & 73.6 & 75.3 & 41.8 & 57.5 & \textbf{32.6} & 55.7 \\
Mamba-3 SISO\refpub & 1.49 & 10.35 & 49.4 & 61.9 & 73.6 & 75.9 & 42.7 & 59.4 & 32.0 & 56.4 \\
Mamba-3 MIMO\refpub & 1.50 & 10.24 & \textbf{51.7} & \textbf{62.3} & \textbf{75.3} & \textbf{76.5} & \textbf{44.5} & \textbf{60.6} & \textbf{32.6} & \textbf{57.6} \\
\bottomrule
\end{tabular*}

\par
\begin{tabular*}{\linewidth}{@{\extracolsep{\fill}}lrrrrrrrrr@{}}
\textbf{(b) Tier} & \shortstack{Params\\(B)} & \shortstack{\revr{Weights}\\\revr{(GB) $\downarrow$}} & PPL $\downarrow$ & 7-task $\uparrow$ & \shortstack{Full\\key $\uparrow$} & \shortstack{Decode\\(ms) $\downarrow$} & \shortstack{\revr{Decode $B{=}512$}\\\revr{(ktok/s) $\uparrow$}} & \shortstack{Prefill\\(ktok/s) $\uparrow$} & \shortstack{State+KV\\(MB) $\downarrow$}\\
\midrule
$T_1$ & 0.474 & \revr{0.95} & 16.49 & 43.8 & 37.0 & \revr{1.25} & \revr{47.6} & \revr{127.6} & 51.15\\
$T_4$ & 0.828 & \revr{1.66} & 12.26 & 50.7 & 37.9 & \revr{1.34} & \revr{44.8} & \revr{113.6} & 52.17\\
$T_7$ & 1.168 & \revr{2.34} & 10.97 & 53.0 & 33.9 & \revr{1.53} & \revr{43.0} & \revr{107.4} & 52.89\\
$T_{10}$ & 1.531 & \revr{3.06} & 10.11 & 55.0 & 33.0 & \revr{1.35} & \revr{41.9} & \revr{102.0} & 53.61\\
\bottomrule
\end{tabular*}
\par
\begin{tabular*}{\linewidth}{@{\extracolsep{\fill}}lrrrrrr|rrr@{}}
\textbf{(c)} & \multicolumn{6}{c}{Shared training} & \multicolumn{3}{c}{Independent training}\\
\cmidrule(lr){2-7}\cmidrule(l){8-10}
Metric & \shortstack{ESSH\\$T_1$} & \shortstack{ESSH\\$T_4$} & \shortstack{ESSH\\$T_7$} & \shortstack{ESSH\\$T_{10}$} & \rev{\shortstack{MatFormer\\(full)}} & \rev{\shortstack{MatMamba\\(full)}} & \shortstack{Mamba-3\\+ SWA} & \shortstack{GDN\\+ SWA} & Transformer++\\
\midrule
Params (M) & 92.1 & 185.3 & 274.3 & 370.9 & \rev{365.6} & \rev{369.9} & 377.1 & 378.1 & 365.6\\
\revr{PPL $\downarrow$} & \revr{23.61} & \revr{17.68} & \revr{15.91} & \revr{14.71} & \revr{16.80} & \revr{16.98} & \revr{\textbf{14.10}} & \revr{14.31} & \revr{15.88}\\
7-task $\uparrow$ & 0.376 & 0.405 & 0.420 & 0.428 & \rev{0.411} & \rev{0.404} & \textbf{0.445} & 0.433 & 0.414\\
Full key $\uparrow$ & 0.241 & 0.370 & 0.366 & \textbf{0.389} & \rev{0.266} & \rev{0.004} & 0.230 & 0.100 & 0.350\\
\bottomrule
\end{tabular*}

\par
\begin{editblock}
\begin{tabular*}{\linewidth}{@{\extracolsep{\fill}}lrrrr@{}}
\revr{Matched dense hybrids} & Params (M) & PPL $\downarrow$ & 7-task $\uparrow$ & Full key $\uparrow$\\
\midrule
ESSH, matched control & 370.8 & 16.95 & 0.407 & 0.20\\
ES-SSM, original mixer & 370.5 & 19.10 & 0.390 & 0.13\\
\bottomrule
\end{tabular*}
\end{editblock}
\par
\begin{editblock}
\begin{tabular*}{\linewidth}{@{\extracolsep{\fill}}lrrrrrr@{}}
\textbf{(d) 7-task (\%) $\uparrow$} & ESSH & TF++$^\dagger$ & GDN$^\dagger$ & Mamba-2$^\dagger$ & M3 SISO$^\dagger$ & M3 MIMO$^\dagger$\\
\midrule
$T_4$ (0.83B) / $\sim$0.88B & 50.7 & 52.8 (+2.1) & 53.7 (+3.0) & 53.4 (+2.7) & 54.4 (+3.7) & 55.3 (+4.6)\\
$T_1$ (0.47B) / $\sim$0.44B & 43.8 & 49.6 (+5.8) & 49.7 (+5.9) & 49.6 (+5.8) & 49.8 (+6.0) & 51.0 (+7.2)\\
\bottomrule
\end{tabular*}
\end{editblock}
\end{table}

\subsection{{Main-model Evaluation}}
\phantomsection\label{sec:exp-1p5b}
As shown in Table~\ref{tab:exp-1p5b}(a), the 1.53B ESSH model achieves a seven-task mean of 55.03\%, \rev{within 0.4 points of the published Transformer++ reference and within 2.6 points of the strongest published model at this size, each of which is a separate model trained only at 1.5B}. \revr{Against Transformer++, ESSH matches HellaSwag at 60.6\% and scores higher on ARC-Challenge (40.7\% versus 40.4\%) and OpenBookQA (32.2\% versus 29.6\%).} 
ESSH also reaches a validation perplexity (PPL) of 10.11 on next-token prediction\rev{. Because each source reports PPL on its own validation shard, the PPL column is not a ranking across models}.

\revr{The same weights must support both the full model and its compact exports. After warm-up, 77.5\% of micro-batches use full capacity in expectation. In the controlled 370M study, allocating more updates to full capacity improves its PPL but worsens the smallest tier (Appendices~\ref{app:el-constants}--\ref{app:el-cost}).}

The 1.5B-scale run also produces standalone compact models.
The 0.828B ESSH-$T_4$ export retains a seven-task mean of 50.72\% with 45.9\% fewer parameters, while ESSH-$T_7$ reaches 53.01\% at 1.168B parameters (Table~\ref{tab:exp-1p5b}(b)).
\revr{In the separate 370M study, ESSH has a better PPL--size trade-off than the train-once MatFormer and MatMamba families at comparable sizes (Table~\ref{tab:exp-1p5b}(c), Figure~\ref{fig:main_frontier}(a)). In Table~\ref{tab:exp-1p5b}(d), we also compare the proposed model at $T_4$ (0.83B) and $T_1$ (0.47B) with comparable separately trained published models of 880M and 440M parameters, respectively. The seven-task mean of ESSH is 2.1--4.6 percentage points lower at $T_4$ and 5.8--7.2 points lower at $T_1$. %
These results illustrate a trade-off between compact-model accuracy and obtaining multiple deployment sizes from a single shared training run. Both ESSH exports come from the same run that produces ten deployment capacities.
The full per-task results are provided in Appendix Table~\ref{tab:compact-published-tasks}.}
Across the ten evaluated tiers, PPL increases smoothly as capacity decreases.

\subsection{The Elastic Frontier: Quality--Capacity Trade-offs}
\label{sec:exp-frontier}
\begin{figure}[htbp]
\centering
\includegraphics[width=\linewidth]{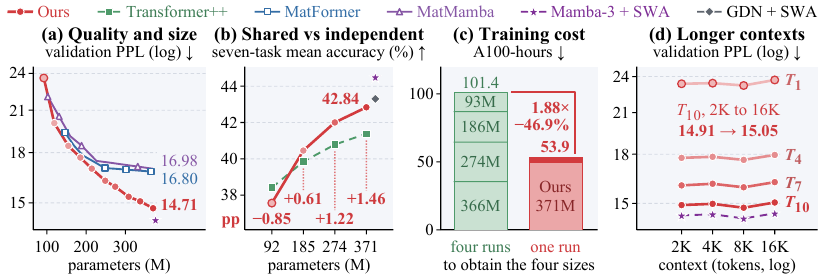}
\caption{Quality and training cost at the 370M scale. \revr{ESSH, MatFormer and MatMamba are train-once families; each Transformer++ model in (b) is trained independently.} $T_k$ denotes an ESSH tier. (a) PPL versus model size; (b) seven-task accuracy against independently trained Transformer++; (c) estimated A100-hours to obtain four sizes from four Transformer++ runs (green) or one shared 371M ESSH run (red), each labelled by size; (d) PPL at longer contexts. Protocols and full results: Appendices~\ref{app:comp-family-curves}, \ref{app:comp-external-cost}, and~\ref{app:ret-length}.}
\label{fig:main_frontier}
\end{figure}

\noindent\textbf{Comparison with elastic baselines (Figure~\ref{fig:main_frontier}(a)).}
At the 370M scale, we compare ESSH with elastic baselines MatFormer and MatMamba to evaluate quality as deployment capacity decreases. \rev{MatFormer and MatMamba are train-once families like ESSH, whereas the single-size models in Table~\ref{tab:exp-1p5b}(c) provide full-capacity quality references for the shared families.} From Figure~\ref{fig:main_frontier}(a), ESSH has a better PPL--size trade-off at intermediate sizes: the 214.5M ESSH-$T_5$ matches full MatMamba's PPL of 16.98 with 42.0\% fewer parameters. 
At nearly matched model sizes, the 274.3M ESSH-$T_7$ achieves a lower PPL of 15.91 than 16.98 for the 270.9M MatFormer. \revr{At full capacity, ESSH also has the highest seven-task mean and full-key accuracy among the three elastic families (Table~\ref{tab:exp-1p5b}(c)). Its seven-task mean is 1.7 percentage points below the independently trained Mamba-3+SWA.} \rev{The 92.1M $T_1$ export is smaller than either baseline at its minimum trained width (Appendix~\ref{app:comp-family-curves}). \revr{Removing every FFN unit still leaves 98.4M parameters, so reaching 92.1M in this backbone also requires spectral-channel reduction (Appendix~\ref{app:el-nesting}).}} \revr{With capacity sampling, distillation and refinement matched in the 3.7B-token controls, ESSH retains lower PPL than MatMamba at matched sizes. Against the Mamba-3+SWA family matched to all four ESSH sizes, ESSH has lower PPL at $T_1$ and $T_4$. At larger tiers, Mamba-3+SWA has up to 0.3 lower PPL, while ESSH uses less recurrent state (Table~\ref{tab:rev-aligned-m3}). ESSH stores a small bank of two-coordinate modes per channel, whereas the Mamba-3 control uses a larger expanded state.}

\noindent\textbf{Shared training versus independent models (Figure~\ref{fig:main_frontier}(b)--(c)).}
Next, we compare ESSH exports with independently trained Transformer++ models at corresponding deployment sizes in Figure~\ref{fig:main_frontier}(b).
ESSH has the higher seven-task mean at three of the four sizes (by up to $+1.46$ points) and trails only at $T_1$ ($-0.85$). \rev{At $T_1$ the independent model is also ahead in PPL (20.36 against 23.61), the size where sharing costs most (Table~\ref{tab:external-four-sizes}).}

\revr{Producing four ESSH sizes costs an estimated 53.9 A100-hours, 46.9\% less than the 101.4 A100-hours of four independent Transformer++ runs (Figure~\ref{fig:main_frontier}(c)). The saving comes from reusing one training run across deployment sizes, including additional tiers that require no retraining. Full-capacity ESSH trains more slowly per token, at 40.66 versus 57.87 ktok/s on A100 (Appendices~\ref{app:comp-amort}--\ref{app:comp-external-cost}).} \revr{Against independently trained ESSH specialists of the same shapes, shared exports have 0.27--3.82 higher PPL, with the largest gap at $T_1$. Their seven-task means differ by less than one percentage point (Table~\ref{tab:sharing-control}).}

\noindent\textbf{Retrieval and longer contexts (Figure~\ref{fig:main_frontier}(d) and Table~\ref{tab:exp-1p5b}(c)).}\phantomsection\label{sec:exp-retention}
We evaluate whether reduced-capacity exports retain retrieval capability and whether their degradation increases beyond the 2K training context.
As shown in Table~\ref{tab:exp-1p5b}(c), on the 2K NIAH retrieval test, \revr{the half-size $T_4$ retains full-key accuracy of 0.370 against 0.389 at full capacity. The unchanged attention layout preserves direct access to the supplied key as spectral-channel count and feed-forward width decrease.} Aggressive parameter reduction to $T_1$ lowers accuracy to 0.241. \revr{Retrieval beyond the receptive field of the stacked attention layers remains future work.}

Figure~\ref{fig:main_frontier}(d) further shows that capacity reduction does not substantially amplify PPL degradation beyond the training context.
From 2K to 16K tokens, the $T_4$--$T_{10}$ PPL gap remains nearly constant. Each tier carries its recurrent state across tokens while attention stays within the same local window.

\noindent\textbf{Mechanism ablations.}\phantomsection\label{sec:exp-ablations}
Matched 3.7B-token ablations isolate the two main components of ESSH
(Appendices~\ref{app:selective_ablations}--\ref{app:elastic_ablations}).
First, making the clock input-independent or neutralizing the write/read gates raises PPL at every tier, showing that selective state transition and
selective writing/readout contribute beyond the fixed spectral recurrence.
\revr{The two-seed joint LTI control removes input dependence from the clock and both gates and raises PPL by 1.1--1.8 at every tier. The matched 3.7B-token dense-hybrid comparison also favors ESSH, with PPL 16.95 versus 19.10 for the original ES-SSM mixer, whose temporal filters remain fixed (Table~\ref{tab:exp-1p5b}(c), Appendices~\ref{app:rev-init-s4d} and~\ref{app:rev-essm}).}
\revr{Second, capacity-mixed pretraining directly optimizes compact predictors through their own prediction losses and full-tier distillation. A model pretrained only at full capacity, then truncated without importance ranking and given the same shared refinement, reaches $T_1$ PPL 45.66 versus 25.32 for ESSH. Removing full-tier distillation increases $T_1$ PPL from 25.32 to 26.42.}
\rev{Ranked recovery matches $T_7$ PPL in the same-architecture comparison (Table~\ref{tab:comp-posthoc-twin}).} \revr{At $T_1$, jointly trained ESSH has 3.8--4.8 lower PPL after equal-budget recovery (Appendix~\ref{app:rev-t1-recovery}).}
\revr{Hankel initialization reaches $T_1$ PPL 50 with fewer training tokens than the tested alternatives (Appendix~\ref{app:rev-init-s4d}).}

\subsection{Inference Efficiency}
\label{sec:exp-efficiency}

\revr{B300 decoding and prefill use the trained ESSH layout and the same compilation policy for every model, with CUDA graphs for decoding (Appendix~\ref{app:setup-hw}).}

\begin{figure}[!htbp]
\centering
\includegraphics[width=\linewidth]{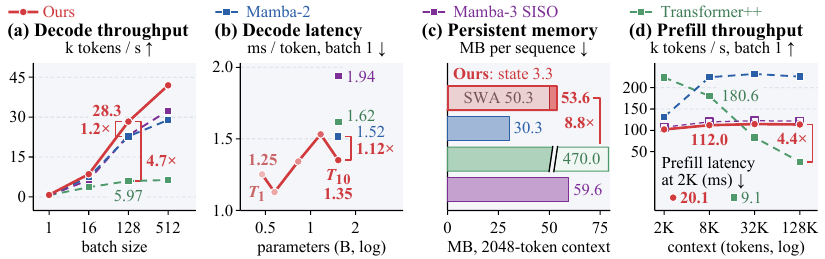}
\caption{B300 deployment at matched full-model parameter counts (1.526B--1.536B)\revr{, with torch.compile for every model and CUDA graphs for decoding (Appendix~\ref{app:setup-hw})}. (a) Decode throughput; (b) batch-one latency; (c) state+KV per sequence; (d) batch-one prefill. Decode uses a 2K context, and compact ESSH tiers use their sizes. Full measurements are in Appendix~\ref{app:cross-device-final}.}
\label{fig:exp-deploy}
\end{figure}

\textbf{Recurrent decoding.}
\revr{At matched parameter counts, full-capacity ESSH has the lowest measured decode latency on B300 at batch sizes 1, 16, 128 and 512. The speedups are $1.12$--$1.45\times$ over Mamba-2, $1.2$--$1.6\times$ over Mamba-3 SISO/MIMO and $1.2$--$6.6\times$ over Transformer++. The Mamba-2 ratio is smallest at batch one and largest at batch 512 (Table~\ref{tab:sys-b300-decode}).} \revr{ESSH computes each channel value once and shares it across modes, avoiding repeated value projections. The fused path combines projections, state updates and readout, which reduces intermediate tensor traffic.}
\revr{At batch one, the tier curve in Figure~\ref{fig:exp-deploy}(b) is non-monotonic: $T_{10}$ takes 1.35\,ms versus 1.53\,ms for $T_7$. Their compiled matrix kernels differ, so parameter count alone does not predict latency (Appendix~\ref{app:sys-b300-decode}).}

\textbf{Memory and long-context prefill.}
ESSH also requires substantially less persistent sequence memory than full attention. At a 2K context, the 1.53B model uses 53.61\,MB of recurrent state plus KV cache per sequence, compared with 469.99\,MB for Transformer++. \revr{Most of this memory is the fixed window-attention cache. Reducing spectral channels shrinks the recurrent state but leaves that cache unchanged.}
For prefill, Transformer++ is faster at 2K tokens, but its cost increases more rapidly with context length. \revr{At 128K tokens on B300, ESSH processes 113.6\,ktok/s compared with 26.0\,ktok/s for Transformer++}, because the spectral scan and sliding-window attention avoid full-prefix attention.
\revr{On B300, Mamba-2 and Mamba-3 SISO prefill faster than the full ESSH model (Appendix~\ref{app:cross-device-final}).}

\section{Conclusion}
\label{sec:conclusion}

ESSH extends LTI spectral filters to input-dependent recurrences while preserving channel-wise truncation. Capacity-mixed training with budget dropout and full-tier distillation maintains model quality across reduced-capacity tiers, while the two-rate capacity map jointly controls spectral-channel count and FFN width. Experiments show that compact exports retain substantial task and retrieval performance, and that the structured recurrence supports chunked parallel training and fused recurrent decoding with low latency on the evaluated GPUs. \revr{As a limitation, the two-rate map trades slightly lower quality at intermediate and larger tiers for a wider size range, smaller recurrent state, and stronger compact tiers. Retrieval is evaluated only within the attention window, up to 2K tokens.} The current framework varies capacity through spectral-channel count and FFN width while keeping the attention layout, depth, and residual width fixed. Extending elasticity to these shared components and developing capacity-selection strategies under model-quality and device constraints are natural directions for future work.

\subsection*{Reproducibility Statement}
Appendices~\ref{app:mixer_details} and~\ref{app:elastic_training} specify the model parameterization and training procedure, and Appendix~\ref{app:experimental_setup} describes the data, baselines, and evaluation protocols. Appendices~\ref{app:theory} and~\ref{app:systems_method} provide the derivations and systems algorithms. Appendix~\ref{app:reproducibility} records the hyperparameters, export procedure, evaluation environment, and compute accounting. Code for model implementation, training, ablations, evaluation, standalone export, and benchmarking is available at \url{https://github.com/songdc98/ESSH}.

\bibliographystyle{iclr2027_conference}
\bibliography{iclr2027_conference}
\clearpage
\appendix
\section*{Appendix}

\begin{tabular}{@{}p{.91\linewidth}r@{}}
\hyperref[app:notation]{\ref*{app:notation}\quad Notation} & \pageref{app:notation} \\
\hyperref[app:theory]{\ref*{app:theory}\quad Derivations} & \pageref{app:theory} \\
\hyperref[app:mixer_details]{\ref*{app:mixer_details}\quad Mixer Parameterization and Cost} & \pageref{app:mixer_details} \\
\hyperref[app:elastic_training]{\ref*{app:elastic_training}\quad Elastic Training} & \pageref{app:elastic_training} \\
\hyperref[app:experimental_setup]{\ref*{app:experimental_setup}\quad Experimental Protocols} & \pageref{app:experimental_setup} \\
\hyperref[app:complete_main_results]{\ref*{app:complete_main_results}\quad Complete Main Results} & \pageref{app:complete_main_results} \\
\hyperref[app:downstream_results]{\ref*{app:downstream_results}\quad Downstream Language Understanding} & \pageref{app:downstream_results} \\
\hyperref[app:retrieval_results]{\ref*{app:retrieval_results}\quad Retrieval and Length} & \pageref{app:retrieval_results} \\
\hyperref[app:synthetic]{\ref*{app:synthetic}\quad Synthetic Recall Suite} & \pageref{app:synthetic} \\
\hyperref[app:cross_domain]{\ref*{app:cross_domain}\quad Cross-Domain Language Modeling} & \pageref{app:cross_domain} \\
\hyperref[app:dna]{\ref*{app:dna}\quad DNA Sequence Modeling} & \pageref{app:dna} \\
\hyperref[app:selective_ablations]{\ref*{app:selective_ablations}\quad Selective-Dynamics Ablations} & \pageref{app:selective_ablations} \\
\hyperref[app:elastic_ablations]{\ref*{app:elastic_ablations}\quad Elasticity Ablations} & \pageref{app:elastic_ablations} \\
\hyperref[app:compression_baselines]{\ref*{app:compression_baselines}\quad Compression and Train-Once Baselines} & \pageref{app:compression_baselines} \\
\hyperref[app:systems_method]{\ref*{app:systems_method}\quad Systems: Algorithms and Kernels} & \pageref{app:systems_method} \\
\hyperref[app:systems_results]{\ref*{app:systems_results}\quad Systems Evaluation} & \pageref{app:systems_results} \\
\hyperref[app:reproducibility]{\ref*{app:reproducibility}\quad Reproducibility} & \pageref{app:reproducibility} \\
\hyperref[app:cross-device-final]{\ref*{app:cross-device-final}\quad Cross-Device Efficiency} & \pageref{app:cross-device-final} \\
\end{tabular}
\clearpage

\section[Notation]{Notation}
\label{app:notation}

Tables~\ref{tab:notation-a} and~\ref{tab:notation-b} summarize notation for the generic SSM, the selective spectral mixer, and the elastic capacity tiers. Channel values and channel outputs are column vectors in $\mathbb R^P$, while each mode state lies in $\mathbb R^{2\times P}$. The collection of retained mode states forms the recurrent state. Feed-forward network is abbreviated as FFN. Throughout, \emph{state update} refers to the recurrent dynamics, \emph{temporal retention} to the persistence of earlier contributions through those dynamics, and \emph{memory} to hardware storage.

The index $k$ denotes a spectral channel and $i$ a mode within that channel. A capacity tier $T$ retains channels $1,\ldots,K_T$ and feed-forward width $n_{\mathrm{ff},T}$. The two-rate map determines both quantities from the budget variable $\xi_T$, while the reported parameter fraction $r$ is measured from the resulting standalone export. Exporting the corresponding tensor slices produces a standalone model. In a metric name, $\Delta$ denotes a difference. Clock increments carry channel and mode indices. Indices introduced within a proof or algorithm are local to that construction.

\begin{table}[htbp]
\centering
\footnotesize
\setlength{\tabcolsep}{4pt}
\caption{Notation for the generic SSM and Hankel spectral construction.}
\label{tab:notation-a}
\begin{tabular}{@{}lp{0.70\linewidth}@{}}
\toprule
Symbol & Meaning \\
\midrule
$L$, $t,s$, $\tau=t-s$ & sequence length, target and source positions, and lag \\
$u(t)$, $h(t)$, $y(t)$ & input, latent state, and output of a generic state-space model \\
$A,B,C,D$ & state transition, input, readout, and direct input--output matrices \\
$N$, $j$ & LTI state dimension and eigenvalue index \\
$d_{\mathrm{in}},d_{\mathrm{out}}$ & input and output dimensions of the generic SSM \\
$\mathcal P_K$, $\Gamma_j$ & rank-$K$ Hankel projection and spectral residue matrix \\
$U,\Lambda$, $\lambda_j$ & eigenbasis, diagonal eigenvalue matrix, and one eigenvalue of $A$ \\
$G(\tau)$, $G(t,s)$ & LTI kernel and its time-varying counterpart \\
$\nu(\lambda)$, $\mu(\lambda)$, $Z$ & geometric response, weighted response, and Hankel matrix \\
$(\sigma_k,\phi_k)$, $\Phi_k$, $M_k$ & Hankel eigenpair, corresponding causal filter, and learned channel projection \\
$K$, $k$ & retained spectral-channel count and spectral-channel index \\
\bottomrule
\end{tabular}
\end{table}

\begin{table}[htbp]
\centering
\footnotesize
\setlength{\tabcolsep}{4pt}
\caption{Notation for the selective spectral mixer and elastic training.}
\label{tab:notation-b}
\begin{tabular}{@{}lp{0.66\linewidth}@{}}
\toprule
Symbol & Meaning \\
\midrule
$u(t)$ & mixer input after RMSNorm \\
$d$, $d_{\mathrm{ff}}$ & residual width and full FFN hidden width \\
$\bar K$, $K_T$ & full spectral-channel count and channel count retained by tier $T$, with $\bar K=32$ in the default recipe \\
$m$, $i$, $P$ & modes per channel, mode index, and channel value width; default language models use $m=8$ and $P=d/\bar K$ \\
$H_{k,i}(t)$ & mode state in $\mathbb R^{2\times P}$ \\
$v_k^0(t)$, $v_k(t)$ & projected channel value before and after the channel-local normalization blend \\
$z_k^{\mathrm{core}}(t)$, $z_k(t)$ & core channel readout and channel output after the zero-lag correction, both in $\mathbb R^P$ \\
$W_k^{\mathrm{in}}$, $W_k^{\mathrm{out}}$ & channel input and output projections, of shapes $P\times d$ and $d\times P$ \\
$e_1$, $\kappa_{k,i}$ & $(1,0)^\top$ and $(\kappa^{\mathrm c}_{k,i},-\kappa^{\mathrm s}_{k,i})^\top$ \\
$A_{k,i}(t)$, $R(\theta)$ & input-dependent mode transition and planar rotation \\
$\rho_{k,i}$, $\gamma_{k,i}=-\log\rho_{k,i}$, $1/\gamma_{k,i}$ & intrinsic decay factor, decay rate, and intrinsic e-folding horizon \\
$\theta_{k,i}$, $\kappa^{\mathrm c}_{k,i}$, $\kappa^{\mathrm s}_{k,i}$ & rotation frequency and modal readout coefficients \\
$\eta^{\mathrm w}_{k,i}(t)$, $\eta^{\mathrm r}_{k,i}(t)$ & write and read gates \\
$\Delta_{k,i}(t)$, $\mathrm e^{-\Delta_{k,i}(t)}$, $q_{k,i}(t)$ & clock increment, exponential factor, and accumulated clock, with $q_{k,i}(0)=0$ \\
$\Pi_{k,i}(t,s)$ & structured transition product from position $s$ to $t$ \\
$G_k$, $V_k$, $B_k$, $Z_k^{\mathrm{core}}$ & within-chunk causal interaction matrix, stacked values, entering-state contribution, and stacked core readouts \\
$\xi_T\in(0,1]$ & input budget variable of the two-rate capacity map for configuration $T$ \\
$r$ & parameter fraction of the standalone export relative to the full model \\
$T_1$--$T_{10}$, $T$ & the ten evaluated deployment tiers and a generic capacity configuration; $T_{10}$ is full capacity \\
$n_{\mathrm{ff},T}$ & FFN hidden width retained by tier $T$ \\
$f=0.75$, $\alpha_{\mathrm{ff}}=0.5$ & reserved full-capacity micro-batch fraction and reduced-capacity FFN gradient scale \\
$\alpha_{\mathrm{mix}}$ & reduced-capacity mixer-gradient scale used in the optimization ablation \\
$p_T$, $p_{T_{10}}$ & next-token distributions of tier $T$ and the full-capacity tier \\
$\Theta$, $\Omega$ & model parameters and optimizer state \\
$\beta_k$, $a_{k,i}$ & channel-local normalization-blend coefficient and mode zero-lag correction coefficient \\
$\widehat\phi_k^{(m)}(\tau)$, $\psi_{k,i}(\tau)$ & fitted channel response and individual damped mode response \\
$Q=64$, $b$, $e$ & maximum chunk length and first and last positions of a chunk \\
\bottomrule
\end{tabular}
\end{table}

\FloatBarrier
\section[Derivations]{Derivations}
\label{app:theory}

This section develops the mathematical results underlying the spectral construction, the selective recurrence, and channel truncation. These results characterize the fixed spectral representation and the realized selective dynamics. Prediction quality after training and truncation is evaluated empirically.

We use $z_k^{\mathrm{core}}(t)$ for the sum of the mode readouts and $z_k(t)$ for the channel output after the zero-lag correction in Equation~\ref{eq:zero_lag}. Both are column vectors in $\mathbb R^P$ before the output projection. With $H_{k,i}(0)=0$, $e_1=(1,0)^\top$, and $\kappa_{k,i}=(\kappa^{\mathrm c}_{k,i},-\kappa^{\mathrm s}_{k,i})^\top$, the core recurrence is
\begin{equation}
\begin{aligned}
H_{k,i}(t)&=\rho_{k,i}\mathrm e^{-\Delta_{k,i}(t)}
R(\theta_{k,i})H_{k,i}(t-1)
+\eta^{\mathrm w}_{k,i}(t)e_1v_k(t)^\top,\\
z_k^{\mathrm{core}}(t)&=
\sum_i\eta^{\mathrm r}_{k,i}(t)H_{k,i}(t)^\top\kappa_{k,i}.
\end{aligned}
\label{eq:th-rec}
\end{equation}
Here $0<\rho_{k,i}<1$, $\Delta_{k,i}(t)>0$, and the write/read gates lie in $(0,2)$. The intrinsic decay rate is
$\gamma_{k,i}=-\log\rho_{k,i}>0$.
Unless stated otherwise, the derivations below use the core output. Appendix~\ref{app:mixer-decay} shows how the zero-lag correction modifies the lag-zero term.

\subsection[Modal form of real recurrences]{Modal form of real recurrences}
\label{app:theory-modal}

A complex-conjugate pole pair in a real SSM can be represented by a two-dimensional damped rotation. Related SSM representations are discussed by \citet{lahoti2026mamba3}. It is sufficient here to consider the scalar temporal response, because each fitted mode applies the same temporal dynamics independently to the $P$ coordinates of a channel value. Here $\mathrm j$ denotes the imaginary unit, with $\mathrm j^2=-1$.

Let $A\in\mathbb R^{N\times N}$ be diagonalizable over $\mathbb C$, with spectral radius less than one, and let $B\in\mathbb R^N$ and $C\in\mathbb R^{1\times N}$. Write
$A=U\Lambda U^{-1}$ and define
\[
r_j=(CU)_j(U^{-1}B)_j.
\]
For integer $\tau\geq0$,
\[
CA^\tau B=\sum_j r_j\lambda_j^\tau.
\]
Because $A$, $B$, and $C$ are real, every nonreal eigenvalue and its residue occur with their complex conjugates. For a conjugate pair
$\rho_i\mathrm e^{\pm\mathrm j\theta_i}$, write
\[
r_i=\frac{\kappa_i^{\mathrm c}+\mathrm j\kappa_i^{\mathrm s}}{2}.
\]
The pair contributes
\[
\begin{aligned}
&r_i(\rho_i\mathrm e^{\mathrm j\theta_i})^\tau+
\overline{r_i}(\rho_i\mathrm e^{-\mathrm j\theta_i})^\tau\\
&\qquad=
\rho_i^\tau
\bigl(
\kappa_i^{\mathrm c}\cos(\theta_i\tau)
-\kappa_i^{\mathrm s}\sin(\theta_i\tau)
\bigr).
\end{aligned}
\]
Thus the impulse response can be written as
\[
CA^\tau B=
\sum_{j\in\mathcal R}r_j\lambda_j^\tau+
\sum_{i\in\mathcal C}\rho_i^\tau
\bigl(
\kappa_i^{\mathrm c}\cos(\theta_i\tau)
-\kappa_i^{\mathrm s}\sin(\theta_i\tau)
\bigr),
\]
where $\mathcal R$ indexes real poles and $\mathcal C$ indexes nonreal conjugate pairs.

The same conjugate-pair term has a real two-dimensional recurrence. An impulse injected along $e_1$ evolves as
\[
(\rho_iR(\theta_i))^\tau e_1
=
\rho_i^\tau
\bigl(\cos(\theta_i\tau),\sin(\theta_i\tau)\bigr)^\top.
\]
Reading this state with
$(\kappa_i^{\mathrm c},-\kappa_i^{\mathrm s})$
gives the expression above. A nonzero real pole is included by taking
$\rho=|\lambda|$ with $\theta=0$ for $\lambda>0$ and $\theta=\pi$ for $\lambda<0$. A zero pole contributes only at lag zero and can be handled separately.

Exact representation requires enough modes to cover the active poles and excludes nontrivial Jordan blocks, whose responses contain polynomial factors in $\tau$. The fitted bank uses $m$ modes per channel under these representation conditions.

\subsection[Oscillation of the spectral channels]{Oscillation of the spectral channels}
\label{app:theory-signs}

We first show that the $k$th Hankel filter has exactly $k-1$ sign changes, and then relate this oscillation count to representations by positive real decays.

For
\[
Z_{st}=\int_0^1(1-\lambda)^2\lambda^{s+t-2}\,\mathrm d\lambda,
\]
let $\phi_k$ denote the $k$th eigenvector when the eigenvalues are ordered in decreasing order. Then $\phi_k$ has exactly $k-1$ sign changes, where zero entries are omitted when counting.

Choose increasing row indices $a_1<\cdots<a_r$ and column indices $b_1<\cdots<b_r$. The corresponding minor satisfies
\[
\det[Z_{a_i b_j}]_{i,j=1}^r
=
\frac{1}{r!}\int_{(0,1)^r}
\det[\lambda_j^{a_i-1}]_{i,j}
\det[\lambda_j^{b_i-1}]_{i,j}
\prod_{j=1}^r(1-\lambda_j)^2
\,\mathrm d\lambda_1\cdots\mathrm d\lambda_r.
\]
On $0<\lambda_1<\cdots<\lambda_r<1$, both generalized Vandermonde determinants are positive. Their product is unchanged by a simultaneous permutation of the integration variables and is positive away from sets of measure zero. Hence every minor of $Z$ is positive, so $Z$ is strictly totally positive. The Gantmacher--Krein oscillation theorem~\citep{gantmacher2002oscillation} then gives simple positive eigenvalues and exactly $k-1$ sign changes in $\phi_k$.

Now consider distinct $\rho_1,\ldots,\rho_m\in(0,1)$ and a nonzero sum
\[
f(t)=\sum_{i=1}^{m}c_i\rho_i^t.
\]
Such a sum has at most $m-1$ distinct zeros on any real interval and therefore at most $m-1$ sign changes on any ordered sample grid. To see this, first combine repeated poles and remove zero coefficients. For $m>1$, divide by the positive function $\rho_m^t$. The derivative of the resulting function is a sum of at most $m-1$ exponentials with distinct rates. Induction and Rolle's theorem then bound the number of zeros of the original sum by $m-1$.

An exact representation of the Hankel filter $\phi_k$ by positive real decays therefore requires at least $k$ distinct positive poles. A damped rotation is not subject to this sign-change count, since one rotating mode can change sign repeatedly across the horizon. This argument concerns exact representation. Table~\ref{tab:init-real} measures the approximation errors of positive-real and rotating-mode fits at finite mode budgets.

\subsection[Hankel projection error]{Hankel projection error}
\label{app:theory-pca}

For $n=s+t-2$, direct integration gives
\begin{equation}
Z_{st}
=
\frac{1}{n+1}
-\frac{2}{n+2}
+\frac{1}{n+3}
=
\frac{2}{(s+t-1)(s+t)(s+t+1)}.
\label{eq:hankel_entries}
\end{equation}
By definition, for $\lambda\sim\mathrm{Unif}[0,1]$,
\[
Z
=
\mathbb E_\lambda
\bigl[\mu(\lambda)\mu(\lambda)^\top\bigr].
\]

Among rank-$K$ orthogonal projectors $\mathcal P$, the projector
\[
\mathcal P_K=\sum_{k=1}^K\phi_k\phi_k^\top
\]
minimizes
\[
\mathbb E_\lambda
\left\|
\mu(\lambda)-\mathcal P\mu(\lambda)
\right\|_2^2,
\]
with minimum error $\sum_{k>K}\sigma_k$. Indeed,
\[
\begin{aligned}
\mathbb E_\lambda\|(I-\mathcal P)\mu(\lambda)\|_2^2
&=\operatorname{tr}((I-\mathcal P)Z)\\
&=\operatorname{tr}(Z)
-\sum_j\sigma_j\,\phi_j^\top\mathcal P\phi_j.
\end{aligned}
\]
For a rank-$K$ orthogonal projector,
\[
0\leq\phi_j^\top\mathcal P\phi_j\leq1,
\qquad
\sum_j\phi_j^\top\mathcal P\phi_j
=
\operatorname{tr}(\mathcal P)=K.
\]
Since the eigenvalues are ordered as
$\sigma_1\geq\sigma_2\geq\cdots$,
\[
\sum_j\sigma_j\,\phi_j^\top\mathcal P\phi_j
\leq
\sum_{j=1}^K\sigma_j,
\]
with equality for $\mathcal P=\mathcal P_K$. The resulting projection error is therefore the eigenvalue tail $\sum_{k>K}\sigma_k$.

For a symmetric SSM with eigenvalues $0\leq\lambda_j<1$, write
\[
A=U\Lambda U^\top,
\qquad
\Gamma_j=(CU)_{:j}(U^\top B)_{j:}.
\]
Its temporal kernel is
\[
G(\tau)=\sum_j\lambda_j^\tau \Gamma_j.
\]
Over the length-$L$ horizon, projecting
\[
\nu(\lambda)
=
[1,\lambda,\ldots,\lambda^{L-1}]^\top
\]
onto the first $K$ Hankel filters gives
\[
M_k
=
\sum_j
\langle\nu(\lambda_j),\phi_k\rangle \Gamma_j,
\qquad
G(\tau)
\approx
\sum_{k=1}^K M_k\phi_k(\tau).
\]

Because $\mu(\lambda)=(1-\lambda)\nu(\lambda)$,
\[
(I-\mathcal P_K)\nu(\lambda)
=
\frac{1}{1-\lambda}
(I-\mathcal P_K)\mu(\lambda),
\qquad \lambda<1.
\]
The projection optimality is stated for the $(1-\lambda)$-weighted family $\mu$. The factor $1/(1-\lambda)$ in the unweighted response remains part of its error bound. The direct term $Du(t)$ is kept separately.

\subsection[Expanding the selective recurrence]{Expanding the selective recurrence}
\label{app:theory-normal}

For a time-varying SSM with zero initial state and readout $y(t)=C_t h(t)$,
\begin{equation}
h(t)=A_t h(t-1)+B_tu(t),\qquad
G(t,s)=C_t(A_tA_{t-1}\cdots A_{s+1})B_s.
\label{eq:ltv_kernel}
\end{equation}
The empty product at $s=t$ is the identity. We denote the mode transition product by
\[
\Pi_{k,i}(t,s)=A_{k,i}(t)A_{k,i}(t-1)\cdots A_{k,i}(s+1),
\qquad \Pi_{k,i}(t,t)=I_2.
\]
For each ESSH mode,
\[
A_{k,i}(t)
=
\rho_{k,i}\mathrm e^{-\Delta_{k,i}(t)}
R(\theta_{k,i}),
\]
so the scalar decay and fixed planar rotation commute across positions and give the structured transition product in Equation~\ref{eq:structured_transition}.

Fix the realized clock and gate sequences, and define
\[
q_{k,i}(t)=\sum_{\ell=1}^{t}\Delta_{k,i}(\ell).
\]
Unrolling the mode state from zero gives
\[
H_{k,i}(t)
=
\sum_{s=1}^{t}
\Pi_{k,i}(t,s)
\eta^{\mathrm w}_{k,i}(s)e_1v_k(s)^\top.
\]
For $\tau=t-s$,
\[
\begin{aligned}
\kappa_{k,i}^\top\Pi_{k,i}(t,s)e_1
&=
\rho_{k,i}^{\,\tau}
\mathrm e^{-[q_{k,i}(t)-q_{k,i}(s)]}
\kappa_{k,i}^\top R(\theta_{k,i}\tau)e_1\\
&=
\mathrm e^{-[q_{k,i}(t)-q_{k,i}(s)]}
\psi_{k,i}(\tau),
\end{aligned}
\]
where
\[
\psi_{k,i}(\tau)
=
\rho_{k,i}^{\,\tau}
\bigl(
\kappa^{\mathrm c}_{k,i}\cos(\theta_{k,i}\tau)
-
\kappa^{\mathrm s}_{k,i}\sin(\theta_{k,i}\tau)
\bigr).
\]
Therefore,
\begin{equation}
z_k^{\mathrm{core}}(t)
=
\sum_{s=1}^{t}\sum_i
\eta^{\mathrm r}_{k,i}(t)
\eta^{\mathrm w}_{k,i}(s)
\mathrm e^{-[q_{k,i}(t)-q_{k,i}(s)]}
\psi_{k,i}(t-s)v_k(s).
\label{eq:normal_kernel}
\end{equation}

For fixed realized controls, each source-to-target coefficient therefore factors into a source write gate, path-dependent retention, a lag-dependent damped-mode response, and a target read gate. The resulting map from channel values to outputs is linear under these fixed controls. The complete mixer remains nonlinear because the values, gates, and clocks depend on the mixer input.

\paragraph{Including the zero-lag correction.}
Equation~\ref{eq:zero_lag} changes only the coefficient of the current channel value. The implemented output is
\begin{equation}
\begin{aligned}
z_k(t)={}&
\sum_{s=1}^{t-1}\sum_{i=1}^{m}
\eta^{\mathrm r}_{k,i}(t)
\eta^{\mathrm w}_{k,i}(s)
\mathrm e^{-[q_{k,i}(t)-q_{k,i}(s)]}
\psi_{k,i}(t-s)v_k(s)\\
&+
\left[
\sum_{i=1}^{m}
(1-a_{k,i})
\eta^{\mathrm r}_{k,i}(t)
\eta^{\mathrm w}_{k,i}(t)
\kappa^{\mathrm c}_{k,i}
\right]v_k(t).
\end{aligned}
\label{eq:corrected_kernel_expansion}
\end{equation}
At $s=t$, $\Pi_{k,i}(t,t)=I_2$ and
$\kappa_{k,i}^\top e_1=\kappa^{\mathrm c}_{k,i}$, so the correction replaces only the current-input coefficient by its $(1-a_{k,i})$ multiple. In particular, $\Delta_{k,i}(t)$ acts on the state propagated into position $t$, not on the value injected at that position.

\paragraph{Effect of an intervening clock.}
For $s<\ell\leq t$, Equation~\ref{eq:structured_transition} gives
\begin{equation}
\frac{\partial\Pi_{k,i}(t,s)}
{\partial\Delta_{k,i}(\ell)}
=
-\Pi_{k,i}(t,s),
\qquad
\left.
\Pi_{k,i}(t,s)
\right|_{\Delta_{k,i}(\ell)\gets\Delta_{k,i}(\ell)+\delta}
=
\mathrm e^{-\delta}\Pi_{k,i}(t,s),
\quad \delta\geq0.
\label{eq:clock_path_effect}
\end{equation}
The derivative is zero for $\ell\notin(s,t]$. Thus increasing an intervening clock multiplicatively reduces the contribution propagated from position $s$ to position $t$ without changing its rotation. This derivative treats the realized values and other controls as fixed. It is not the full derivative with respect to the mixer input.

\paragraph{Selective transition versus tokenwise output gating.}
A three-position example isolates this distinction. Consider one scalar-valued channel with one active mode,
$\theta=0$, $\kappa^{\mathrm c}=1$, $\kappa^{\mathrm s}=0$, and both gates equal to one. Let
\[
v(1)\in\{-1,+1\},
\qquad
v(2)=v(3)=0.
\]
At position two, vary an input coordinate ignored by the value projection but used by the clock projection, producing
$\Delta(2)=\delta_+$ or $\delta_-$ with
$\delta_+\neq\delta_-$. Choose the two position-two inputs with equal norms so that normalization does not alter the construction, and keep the position-three mixer input fixed. Then
\[
z(3;v(1),\delta_\pm)
=
\rho^2
\mathrm e^{-\Delta(3)}
\mathrm e^{-\delta_\pm}
v(1).
\]
Because $v(3)=0$, the zero-lag correction contributes nothing at position three. The mixed finite difference between the earlier value and the intervening clock is therefore
\begin{equation}
\begin{aligned}
&z(3;+1,\delta_+)-z(3;-1,\delta_+)
-z(3;+1,\delta_-)+z(3;-1,\delta_-)\\
&\hspace{12mm}
=
2\rho^2\mathrm e^{-\Delta(3)}
\bigl(
\mathrm e^{-\delta_+}
-
\mathrm e^{-\delta_-}
\bigr)
\neq0.
\end{aligned}
\label{eq:three_position_selection}
\end{equation}

For the output-gated fixed-filter mixer in Equation~\ref{eq:prior_elastic}, the gate at position three is unchanged when its mixer input is fixed. With that gate fixed, the output is additive in the contributions from different source positions, so the same mixed finite difference is zero. The selective transition therefore permits an intervening token to multiplicatively change the later contribution of an earlier value. Task-level effects are evaluated in Table~\ref{tab:sel-mech}.

\subsection[State bounds and temporal retention]{State bounds and temporal retention}
\label{app:theory-state}

Fix the model parameters and a realized sequence of values and controls. If two mode trajectories differ only in their state at position $s$, then
\[
H_{k,i}(t)-H'_{k,i}(t)
=
\Pi_{k,i}(t,s)
\bigl(
H_{k,i}(s)-H'_{k,i}(s)
\bigr).
\]
Because planar rotation preserves the Frobenius norm,
\begin{equation}
\begin{aligned}
\|H_{k,i}(t)-H'_{k,i}(t)\|_F
={}&
\rho_{k,i}^{\,t-s}
\mathrm e^{-[q_{k,i}(t)-q_{k,i}(s)]}
\|H_{k,i}(s)-H'_{k,i}(s)\|_F\\
\leq{}&
\rho_{k,i}^{\,t-s}
\|H_{k,i}(s)-H'_{k,i}(s)\|_F.
\end{aligned}
\label{eq:state_difference_decay}
\end{equation}
Thus, for fixed realized controls, an existing mode-state difference is retained from $s$ to $t$ by the factor
\[
\rho_{k,i}^{\,t-s}
\mathrm e^{-[q_{k,i}(t)-q_{k,i}(s)]}.
\]
This factor is also the operator norm of the direct state-to-state map under the Frobenius norm. It does not include the dependence of the realized controls on the mixer input.

\paragraph{Bounded values and states.}
If $\|u(t)\|_2$ is bounded over the positions under consideration, Equation~\ref{eq:value_blend} gives
\[
\|v_k(t)\|_2
\leq
|1-\beta_k|
\|W_k^{\mathrm{in}}\|_2
\|u(t)\|_2
+
|\beta_k|\sqrt P.
\]
The absolute values are required because $\beta_k$ is unconstrained. Let $V_{k,\max}$ be any finite bound on the resulting channel values. Since
$\|A_{k,i}(t)\|_2\leq\rho_{k,i}$ and
$\eta^{\mathrm w}_{k,i}(t)\leq2$,
\[
\|H_{k,i}(t)\|_F
\leq
\rho_{k,i}
\|H_{k,i}(t-1)\|_F
+
2V_{k,\max}.
\]
Starting from $H_{k,i}(0)=0$ therefore gives
\begin{equation}
\|H_{k,i}(t)\|_F
\leq
2V_{k,\max}
\sum_{s=1}^{t}\rho_{k,i}^{\,t-s}
=
2V_{k,\max}
\frac{1-\rho_{k,i}^{\,t}}
{1-\rho_{k,i}}.
\label{eq:bounded_mode_state}
\end{equation}

\paragraph{The corrected output.}
Substituting the state update into Equation~\ref{eq:zero_lag} separates the propagated state from the corrected current injection:
\[
z_k(t)
=
\sum_i
\eta^{\mathrm r}_{k,i}(t)
\left[
\bigl(
A_{k,i}(t)H_{k,i}(t-1)
\bigr)^\top
\kappa_{k,i}
+
(1-a_{k,i})
\eta^{\mathrm w}_{k,i}(t)
\kappa^{\mathrm c}_{k,i}
v_k(t)
\right].
\]
Using the triangle inequality, the gate bounds, and Equation~\ref{eq:bounded_mode_state} at $t-1$ gives
\begin{equation}
\|z_k(t)\|_2
\leq
4V_{k,\max}
\sum_i
\left[
\|\kappa_{k,i}\|_2
\frac{
\rho_{k,i}
(1-\rho_{k,i}^{\,t-1})
}{
1-\rho_{k,i}
}
+
|1-a_{k,i}|
|\kappa^{\mathrm c}_{k,i}|
\right].
\label{eq:bounded_corrected_output}
\end{equation}
For fixed $V_{k,\max}$ and finite parameters, this bound is uniform in sequence length. Its constant depends on the modal decay and readout coefficients.

\subsection[Output change under channel truncation]{Output change under channel truncation}
\label{app:theory-trunc}

Fix the input and parameters of one mixer sublayer, and let
\[
y_K(t)
=
K^{-1/2}
\sum_{k=1}^{K}
W_k^{\mathrm{out}}z_k(t)
\]
denote its output when the first $K$ channels are retained, including the zero-lag correction. Define
\[
S_K(t)
=
\sum_{k=1}^{K}
W_k^{\mathrm{out}}z_k(t).
\]
The difference between the full mixer and the $K$-channel mixer is exactly
\begin{equation}
\begin{aligned}
y_{\bar K}(t)-y_K(t)
={}&
\frac{1}{\sqrt{\bar K}}
\sum_{k>K}
W_k^{\mathrm{out}}z_k(t)\\
&+
\bigl(
\bar K^{-1/2}-K^{-1/2}
\bigr)S_K(t).
\end{aligned}
\label{eq:tier_normalization}
\end{equation}
The first term contains the discarded channels, while the second accounts for the tier-dependent output normalization.

To bound the corrected channel output, define
\[
\widetilde G_k(\tau)
=
\begin{cases}
\displaystyle
\sum_i
|1-a_{k,i}|\,
|\kappa^{\mathrm c}_{k,i}|,
& \tau=0,\\[2mm]
\displaystyle
\sum_i
|\psi_{k,i}(\tau)|,
& \tau>0.
\end{cases}
\]
Equation~\ref{eq:corrected_kernel_expansion}, together with
$\eta^{\mathrm w}_{k,i},\eta^{\mathrm r}_{k,i}<2$ and
$\mathrm e^{-[q_{k,i}(t)-q_{k,i}(s)]}\leq1$, gives
\[
\|z_k(t)\|_2
\leq
4
\sum_{s=1}^{t}
\widetilde G_k(t-s)
\|v_k(s)\|_2.
\]
Applying the triangle inequality and
$\|Wz\|_2\leq\|W\|_2\|z\|_2$ to
Equation~\ref{eq:tier_normalization} yields
\begin{equation}
\begin{aligned}
\|y_{\bar K}(t)-y_K(t)\|_2
\leq{}&
\frac{4}{\sqrt{\bar K}}
\sum_{k>K}
\|W_k^{\mathrm{out}}\|_2
\sum_{s=1}^{t}
\widetilde G_k(t-s)
\|v_k(s)\|_2\\
&+
\left|
\bar K^{-1/2}-K^{-1/2}
\right|
\|S_K(t)\|_2.
\end{aligned}
\label{eq:tail_bound}
\end{equation}

Equation~\ref{eq:tail_bound} separates the two direct effects of channel truncation at a fixed mixer input: removing higher-indexed channel contributions and changing the tier normalization. Which retained configurations remain predictive is determined by the learned channel organization and capacity-mixed training.

\subsection[Mode-fit approximation and initialization]{Mode-fit approximation and initialization}
\label{app:theory-fit}

Define the relative mode-fit error
\[
\varepsilon_k^{(m)}
=
\frac{
\|\phi_k-\widehat\phi_k^{(m)}\|_2
}{
\|\phi_k\|_2
}.
\]
With $m=8$, the first eight channels retained by the smallest deployment tier have relative fit error at most $0.036$. Higher-index filters are more oscillatory, and their errors are weighted by smaller eigenvalues in Equation~\ref{eq:projection_plus_mode_fit}. At $L=2048$, $\sigma_9/\sigma_1=2.29\times10^{-6}$ and the tail beyond channel eight is $3.31\times10^{-6}\operatorname{tr}Z$. The projection-plus-fit error is below $4\times10^{-6}\operatorname{tr}Z$ at every deployment tier. These statistics concern the fitted weighted filter family. The modal parameters remain trainable.

\paragraph{Combining channel truncation with finite mode fits.}
The Hankel projection in Appendix~\ref{app:theory-pca} and the finite-mode approximation contribute separate terms to the reconstruction error of the weighted family $\mu$. Using the original Hankel projection coefficients with the fitted filters, define
\begin{equation}
\begin{aligned}
\mathcal E_K^{(m)}
&=
\mathbb E_{\lambda\sim\mathrm{Unif}[0,1]}
\left\|
\mu(\lambda)
-
\sum_{k=1}^{K}
\langle\mu(\lambda),\phi_k\rangle
\widehat\phi_k^{(m)}
\right\|_2^2\\
&=
\sum_{j>K}\sigma_j
+
\sum_{k=1}^{K}
\sigma_k
\bigl(
\varepsilon_k^{(m)}
\bigr)^2.
\end{aligned}
\label{eq:projection_plus_mode_fit}
\end{equation}
The first term is the Hankel projection tail from omitting channels $k>K$, while the second is the error from representing each retained filter with $m$ modes.

To verify the second line, expand $\mu(\lambda)$ in the orthonormal eigenbasis of $Z$. The residual is
\[
\sum_{j>K}
\langle\mu,\phi_j\rangle\phi_j
+
\sum_{k=1}^{K}
\langle\mu,\phi_k\rangle
\bigl(
\phi_k-\widehat\phi_k^{(m)}
\bigr).
\]
For distinct eigenvectors,
\[
\mathbb E_\lambda
\left[
\langle\mu,\phi_j\rangle
\langle\mu,\phi_k\rangle
\right]
=
\phi_j^\top Z\phi_k
=
\sigma_k\mathbf1[j=k],
\]
so the cross terms vanish after integration. Since
$\|\phi_k\|_2=1$, the remaining diagonal terms give
Equation~\ref{eq:projection_plus_mode_fit}. The fitted filters themselves need not be orthogonal.

\paragraph{Implemented initialization.}
The fitted response is the modal sum before input-dependent controls. The implemented initialization also applies the initial clock and zero-lag correction from Appendix~\ref{app:mixer-decay}. Its response is therefore evaluated using the complete recurrence in Equations~\ref{eq:th-rec} and~\ref{eq:zero_lag}.

For the unmodulated fitted response, the finite-horizon filter error also gives
\begin{equation}
\left\|
\sum_{\tau=0}^{t-1}
\bigl(
\phi_k(\tau)-\widehat\phi_k^{(m)}(\tau)
\bigr)
v_k(t-\tau)
\right\|_2
\leq
\varepsilon_k^{(m)}
\left(
\sum_{s=1}^{t}
\|v_k(s)\|_2^2
\right)^{1/2},
\qquad t\leq L.
\label{eq:fit_to_output_error}
\end{equation}
This follows from the triangle inequality and Cauchy--Schwarz over sequence positions.

These bounds characterize the initialization. During training, the modal parameters, clocks, and gates are learned, and the resulting dynamics are described by the selective recurrence in Equation~\ref{eq:normal_kernel}.

\section[Mixer Parameterization and Cost]{Mixer Parameterization and Cost}
\label{app:mixer_details}

This section specifies the spectral initialization, channel-value transformation, and zero-lag correction that complete the selective recurrence in Equation~\ref{eq:selective_mode}. It then gives the parameter, arithmetic, and recurrent-state costs of each capacity tier.

\subsection[Initialization from Hankel channels]{Initialization from Hankel channels}
\label{app:modal_initialization}

We compute the first $\bar K=32$ unit eigenvectors of the length-2048 Hankel matrix and fit each with a small bank of damped rotation modes. Filter-fit diagnostics use the first 24 eigenvectors resolved by the double-precision conditioning check. All 32 fitted modal parameter sets are trainable initializations. The weighted approximation error is quantified in Appendix~\ref{app:theory-fit}.

\subsubsection[The matrix-pencil fit]{The matrix-pencil fit}
\label{app:init-pencil}

For each length-2048 channel, the matrix-pencil procedure uses pencil parameter 512 and a rank-$m$ truncated SVD to estimate $m$ poles. Each pole is stored as one damped mode, so conjugate poles occupy separate modes. After clipping the pole magnitudes to stable decays, complex readout coefficients are obtained by least squares. The resulting
$(\rho,\theta,\kappa^{\mathrm c},\kappa^{\mathrm s})$
tables initialize every spectral layer.

With $m=8$, the first eight channels, which are retained by $T_1$, have maximum relative fitting error $0.036$. Fit error increases for higher-index, more oscillatory channels, while all modal parameters remain trainable after initialization. Appendices~\ref{app:init-real} and~\ref{app:init-r} evaluate fitting fidelity and trained-model behavior separately.

\subsection[Projections, gates, and clock]{Projections, gates, and clock}
\label{app:mixer-proj}

The spectral block receives the RMS-normalized mixer input $u(t)$~\citep{NEURIPS2019_1e8a1942}. Each channel first computes
$v_k^0(t)=W_k^{\mathrm{in}}u(t)$ and then applies the learned channel-local blend
\begin{equation}
v_k(t)
=
(1-\beta_k)v_k^0(t)
+
\beta_k
\frac{v_k^0(t)}
{\sqrt{P^{-1}\|v_k^0(t)\|_2^2+\epsilon_{\mathrm{norm}}}},
\label{eq:value_blend}
\end{equation}
where $\beta_k$ is a learned scalar initialized at zero and
$\epsilon_{\mathrm{norm}}>0$ stabilizes the denominator. The transformation therefore starts as the identity. Its normalization statistic uses only the $P$ coordinates of channel $k$, so removing other channels does not change a retained channel value for the same mixer input.

For channel $k$ and mode $i$, three affine projections of $u(t)$ produce the write gate, read gate, and clock increment:
\[
\begin{aligned}
\eta^{\mathrm w}_{k,i}(t)
&=
2\,\mathrm{sigmoid}
\bigl(
(w^{\mathrm w}_{k,i})^\top u(t)+b^{\mathrm w}_{k,i}
\bigr),\\
\eta^{\mathrm r}_{k,i}(t)
&=
2\,\mathrm{sigmoid}
\bigl(
(w^{\mathrm r}_{k,i})^\top u(t)+b^{\mathrm r}_{k,i}
\bigr),\\
\Delta_{k,i}(t)
&=
\mathrm{softplus}
\bigl(
(w^{\Delta}_{k,i})^\top u(t)+b^\Delta_{k,i}
\bigr).
\end{aligned}
\]
All three projection vectors lie in $\mathbb R^d$, and their biases are scalars. Gate weights and biases are initialized at zero, so both gates start at one. Clock weights are initialized at zero with bias $-3$, giving
$\Delta_0=\mathrm{softplus}(-3)\approx0.0486$. This corresponds to a clock-only e-folding scale of approximately $20.6$ positions at initialization. Because the clock parameters are learned, this additional attenuation can increase or decrease during training.

\subsection[Stable decay and zero-lag correction]{Stable decay and zero-lag correction}
\label{app:mixer-decay}

The implementation stores $\tilde\rho$ and parameterizes
\[
\gamma=\mathrm{softplus}(\tilde\rho),
\qquad
\rho=\exp(-\gamma).
\]
Hence
\[
\|A_{k,i}(t)\|_2
=
\exp\bigl(
-\gamma_{k,i}-\Delta_{k,i}(t)
\bigr)
<1.
\]
Each mode transition is therefore contractive, while the learned
$\tilde\rho_{k,i}$ allows its intrinsic timescale to increase or decrease during training.

Each mode also has a learned zero-lag correction coefficient $a_{k,i}$, initialized at $1/2$. The corrected channel output is
\begin{equation}
z_k(t)
=
z_k^{\mathrm{core}}(t)
-
\sum_i
a_{k,i}
\eta^{\mathrm r}_{k,i}(t)
\eta^{\mathrm w}_{k,i}(t)
\kappa^{\mathrm c}_{k,i}
v_k(t).
\label{eq:zero_lag}
\end{equation}
The correction changes only the lag-zero coefficient and requires no additional recurrent state.

To see the equivalent recurrence, suppress the mode index, let
$b_t=\eta^{\mathrm w}(t)e_1v(t)^\top$, and define
$J_t=H_t-a b_t$. With $b_0=0$,
\[
J_t
=
A_tJ_{t-1}
+
(1-a)b_t
+
aA_tb_{t-1},
\]
and
\[
\eta^{\mathrm r}(t)J_t^\top\kappa
=
\eta^{\mathrm r}(t)H_t^\top\kappa
-
a\eta^{\mathrm r}(t)b_t^\top\kappa.
\]
Thus the corrected readout can be computed from the existing mode state and current channel value. At the initialization $a=1/2$, the equivalent recurrence assigns equal coefficients to the current injection and the propagated previous injection.

At initialization the gates equal one and the clock is
$\Delta_0=\mathrm{softplus}(-3)$. Let
$\psi^0_{k,i}$ denote the fitted modal response from
Equation~\ref{eq:normal_kernel}. The implemented initial response from channel values to channel output is
\[
\widehat\phi_{k,\mathrm{init}}(\tau)
=
\mathrm e^{-\Delta_0\tau}
\sum_i\psi^0_{k,i}(\tau)
-
\mathbf1[\tau=0]
\sum_i
a^0_{k,i}
\kappa^{\mathrm c,0}_{k,i}.
\]
The first term incorporates the initial clock attenuation, while the second applies the learned form of the lag-zero correction. This is the implemented initialization analyzed in Appendix~\ref{app:theory-fit}.

\subsection[Channel aggregation and the complete computation]{Channel aggregation and the complete computation}
\label{app:mixer-complete}

At tier $T$, the mixer computes the first $K_T$ channel values and their channel-local blends, evaluates the corresponding $K_Tm$ gate and clock rows, updates the retained mode states, and applies Equation~\ref{eq:zero_lag}. The corrected channel outputs are concatenated and projected through the first $K_TP$ columns of $W^{\mathrm{out}}$, with scale $K_T^{-1/2}$. The resulting width-$d$ output is added to the residual stream.

No data-dependent normalization statistic is shared across spectral channels. For the same mixer input, truncating later channels therefore leaves the internal computation of every retained channel unchanged.

With the value blend disabled, the linear input--output map of channel $k$ factors as
\[
W_k^{\mathrm{out}}W_k^{\mathrm{in}},
\]
and has rank at most $P$. ESSH therefore uses a parameter-efficient factorized channel map in place of the dense $M_k$ in the reference spectral construction of Equation~\ref{eq:spectral_base}.

\subsection[Parameter and FLOP accounting]{Parameter and FLOP accounting}
\label{app:mixer-count}

Table~\ref{tab:mixer-count} separates the projection parameters from the smaller modal tables. At full capacity, the value and output projections contain $2d^2$ parameters, and the three selective projections contain $3\bar Kmd$ parameters before biases. A SwiGLU block contains $3d\,d_{\mathrm{ff}}$ projection parameters. At the 370M configuration, the feed-forward layer accounts for approximately $84\%$ of the parameters in a spectral block. This separation motivates reducing feed-forward width more aggressively than spectral-channel count in the two-rate capacity map.

We count one multiply-accumulate as two FLOPs. At tier $T$, the value/output and selective projections require
\[
2dK_TP+3dK_Tm
\]
MACs per token, while the feed-forward projections require
\[
3d\,n_{\mathrm{ff},T}
\]
MACs. The recurrent core requires $O(K_TmP)$ tokenwise work and
$2K_TmP$ recurrent-state entries per spectral layer. Attention remains shared across tiers and contributes its own projection cost and window KV cache. Table~\ref{tab:complexity} summarizes the corresponding full-capacity costs.

For Mamba-2, $d_{\mathrm{inner}}$ denotes its expanded mixer width and
$N_{\mathrm{state}}$ its state dimension per value coordinate. These are distinct from the generic SSM dimensions in Table~\ref{tab:notation-a}. The asymptotic expressions omit lower-order buffers and device-dependent effects such as memory traffic and kernel-launch overhead, which are measured directly in the systems evaluation.

\paragraph{Exact tier-dependent parameter count.}
For the stated bias-free value/output and SwiGLU projections, each retained spectral channel contributes
\[
2dP+3md+8m+1
\]
parameters: value/output weights, three selective-projection weights and biases, five learned modal scalars per mode, and one value-blend coefficient. With $N_{\mathrm{layers}}$ blocks and attention-layer set $\mathcal A$,
\begin{equation}
\begin{aligned}
\operatorname{Params}(T)
={}&
\operatorname{Params}_{\mathrm{fixed}}\\
&+
(N_{\mathrm{layers}}-|\mathcal A|)
K_T
(2dP+3md+8m+1)\\
&+
3N_{\mathrm{layers}}d\,n_{\mathrm{ff},T}.
\end{aligned}
\label{eq:exact_tier_parameter_count}
\end{equation}
The fixed term contains embeddings, attention projections, and normalization parameters. The recurrent state contains
\[
2(N_{\mathrm{layers}}-|\mathcal A|)
K_TmP
\]
entries and is independent of $n_{\mathrm{ff},T}$.

Thus the two elastic axes control different deployment costs: feed-forward width primarily changes parameter count and feed-forward computation, whereas spectral-channel count changes mixer computation and recurrent-state storage in addition to channel parameters. The shared window KV cache remains fixed across tiers.

\begin{table}[htbp]
\centering
\small
\setlength{\tabcolsep}{5pt}
\caption{Parameters per block for the two language-model configurations, with $\bar K=32$, $m=8$, and $P=d/\bar K$. Whole-model counts include tied embeddings.}
\label{tab:mixer-count}
\resizebox{\linewidth}{!}{%
\begin{tabular}{@{}llcc@{}}
\toprule
Component & Parameters & 1.5B ($d{=}2048$, $d_{\mathrm{ff}}{=}5632$) & 370M ($d{=}896$, $d_{\mathrm{ff}}{=}4608$) \\
\midrule
value in / out projections & $2d^{2}$ & 8,388,608 & 1,605,632 \\
write, read gates & $2(\bar Km\,d+\bar Km)$ & 1,049,088 & 459,264 \\
clock & $\bar Km\,d+\bar Km$ & 524,544 & 229,632 \\
spectral tables $(\rho,\theta,\kappa^{\mathrm c},\kappa^{\mathrm s})$ & $4\bar Km$ & 1,024 & 1,024 \\
zero-lag coefficient $a$, value blend $\beta$ & $\bar Km+\bar K$ & 288 & 288 \\
SwiGLU feed-forward~\citep{shazeer2020glu} & $3d\,d_{\mathrm{ff}}$ & 34,603,008 & 12,386,304 \\
sliding-window attention (per SWA block) & $4d^{2}$ & 16,777,216 & 3,211,264 \\
two RMSNorms & $2d$ & 4,096 & 1,792 \\
\midrule
whole model, full capacity & & 1,531,089,696 & 370,866,144 \\
whole model, exported $T_4$ & & 827,570,754 & 185,254,998 \\
\bottomrule
\end{tabular}}
\end{table}

\begin{table}[htbp]
\centering
\small
\setlength{\tabcolsep}{4pt}
\caption{Per-token arithmetic and per-sequence persistent state at full capacity. Compact-tier spectral and feed-forward costs follow by replacing $\bar K$ and $d_{\mathrm{ff}}$ with $K_T$ and $n_{\mathrm{ff},T}$, while attention remains shared. The Mamba-2 row assumes a shared $B/C$ group and $N_{\mathrm{state}}\le d_{\mathrm{inner}}$, as in the timed configuration~\citep{dao2024mamba2}. One multiply-accumulate counts as two FLOPs.}
\label{tab:complexity}
\setlength{\tabcolsep}{2.5pt}
\begin{tabular}{@{}lccc@{}}
\toprule
Block & projection MACs + scan cost & decode cost per token & persistent state \\
\midrule
Spectral mixer &
\shortstack{$2d^{2}+3\bar Kmd$\\$+\,O(Q\bar K(P+m)+\bar KmP)$} &
$O(d^{2}+\bar Kmd+md)$ &
$2\bar KmP=2md$ \\
Window attention &
$4d^{2}+O(wd)$ &
$O(d^{2}+wd)$ &
$2wd$ \\
SwiGLU feed-forward &
$3d\,d_{\mathrm{ff}}$ &
$O(d\,d_{\mathrm{ff}})$ &
$0$ \\
Mamba-2 block &
\shortstack{$O(d\,d_{\mathrm{inner}})$\\$+\,O((Q+N_{\mathrm{state}})d_{\mathrm{inner}})$} &
$O(d\,d_{\mathrm{inner}}+N_{\mathrm{state}}d_{\mathrm{inner}})$ &
$N_{\mathrm{state}}d_{\mathrm{inner}}+O(d_{\mathrm{inner}})$ \\
Full attention &
$4d^{2}+O(Ld)$ &
$O(d^{2}+Ld)$ &
$2Ld$ \\
\bottomrule
\end{tabular}
\end{table}

\section[Elastic Training]{Elastic Training}
\label{app:elastic_training}

This section specifies the nested tensor slices, capacity-mixed sampling, gradient allocation, full-tier distillation, and standalone export used to train the elastic family.

\begin{figure}[htbp]
\centering
\includegraphics[width=\linewidth]{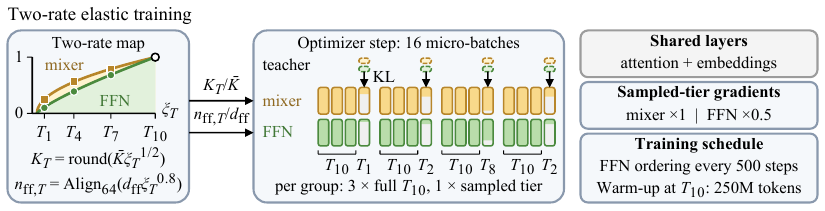}
\caption{Two-rate elastic training. The two-rate map reduces spectral-channel count and feed-forward width at different rates; $\operatorname{Align}_{64}(x)=64\max\{1,\lfloor x/64\rfloor\}$. The optimizer-step diagram shows joint training of full and sampled capacities with full-tier distillation. The right-hand boxes identify the shared attention and embeddings, reduced-capacity gradient scales, and the capacity warm-up and FFN-ordering schedule.}
\label{fig:elastic_training}
\end{figure}

Figure~\ref{fig:elastic_training} summarizes the two-rate capacity map and the capacity-mixed training rule.

\subsection[The two-rate capacity map]{The two-rate capacity map}
\label{app:elastic-map}

Equation~\ref{eq:two_rate} maps a budget variable to a retained spectral-channel count and feed-forward width. Attention, residual width, embeddings, and normalization dimensions remain fixed across tiers. The exponents $1/2$ and $0.8$ are empirical recipe choices evaluated in Appendix~\ref{app:el-nesting}.

At the 370M configuration, for example, $T_1$ retains eight of 32 spectral channels and 448 of 4608 feed-forward units. The map therefore preserves a larger fraction of recurrent channels while concentrating most parameter reduction in the feed-forward layers. Attention dimensions and window size remain shared across tiers.

Table~\ref{tab:el-map} lists the ten deployment tiers, whose budget fractions are
\[
(\xi_{T_1},\ldots,\xi_{T_{10}})
=
\frac{1}{32}(2,4,7,10,13,17,20,24,28,32).
\]
The pretraining sampler uses the separate ten training capacities
\[
\left\{
1/16,\,
3/32,\,
1/8,\,
5/32,\,
3/16,\,
1/4,\,
3/8,\,
1/2,\,
3/4,\,
1
\right\}.
\]
The ten training capacities place six budgets at $\xi\le1/4$. Deployment tiers evaluate slices of the same nested tensors. At 370M, $T_3$--$T_7$ and $T_9$ lie between the capacities directly sampled during training. Table~\ref{tab:main-370-tiers} evaluates these intermediate exports alongside the directly sampled configurations. Results under the default map use Table~\ref{tab:el-map}. Alternative-map ablations keep the budget variables and report their resulting sizes.

\begin{table}[htbp]
\centering
\small
\setlength{\tabcolsep}{5pt}
\caption{The ten deployment tiers and measured parameter counts. Channel count and FFN width follow the two-rate map. Fixed embedding parameters cause the exported parameter fractions to differ between the two model scales.}
\label{tab:el-map}
\begin{tabular}{@{}crrrrrr@{}}
\toprule
tier & $K_T$ & $n_{\mathrm{ff},T}$, 1.5B & $n_{\mathrm{ff},T}$, 370M & params, 1.5B (B) & params, 370M (M) & modes $K_Tm$ \\
\midrule
$T_1$ & 8 & 576 & 448 & 0.474 & 92.1 & 64 \\
$T_2$ & 11 & 1024 & 832 & 0.575 & 118.9 & 88 \\
$T_3$ & 15 & 1664 & 1344 & 0.716 & 154.7 & 120 \\
$T_4$ & 18 & 2176 & 1792 & 0.828 & 185.3 & 144 \\
$T_5$ & 20 & 2688 & 2240 & 0.931 & 214.5 & 160 \\
$T_6$ & 23 & 3392 & 2752 & 1.076 & 248.8 & 184 \\
$T_7$ & 25 & 3840 & 3136 & 1.168 & 274.3 & 200 \\
$T_8$ & 28 & 4416 & 3648 & 1.291 & 308.6 & 224 \\
$T_9$ & 30 & 5056 & 4096 & 1.416 & 337.9 & 240 \\
$T_{10}$ & 32 & 5632 & 4608 & 1.531 & 370.9 & 256 \\
\bottomrule
\end{tabular}
\end{table}

\subsection[Standalone export, tensor by tensor]{Standalone export, tensor by tensor}
\label{app:elastic-slice}

At tier $T$, retain the first $K_TP$ value-projection rows and matching output-projection columns, the first $K_Tm$ gate and clock rows and biases, and the first $K_T$ channel blocks of the modal tables, zero-lag coefficients, and channel-local blend coefficients. In each feed-forward layer, retain the first $n_{\mathrm{ff},T}$ rows of the gate and up projections and the matching columns of the down projection.

In exact arithmetic, these slices form a standalone model that computes the same function as the shared model evaluated at tier $T$. Every retained operation uses the same parameters and operands in both evaluations: channel-local normalization keeps width $P$ unchanged, retained mode states therefore match by induction over sequence positions, attention is shared, and the same FFN units are selected. The equality then propagates through the residual blocks to the final normalization and output head.

Different kernel reduction orders can introduce floating-point differences. Export checks therefore compare the implemented standalone and shared model outputs at the specified numerical precision.

\subsection[Importance ordering of the feed-forward units]{Importance ordering of the feed-forward units}
\label{app:elastic-sort}

Because reduced-capacity tiers retain the first $n_{\mathrm{ff},T}$ feed-forward units, their ordering determines which units remain in reduced-capacity models. A full-width SwiGLU block is invariant to a common permutation of the gate and up rows and the corresponding down-projection columns.

Every 500 steps through the first $80\%$ of training, the recipe orders units using a running squared-activation score weighted by the squared norm of the corresponding down-projection column. The same permutation is applied to the gate and up rows, down columns, optimizer moments, and running activation statistics. Distributed workers synchronize the scores before applying the permutation.

The permutation preserves the current full-width function in exact arithmetic while changing which units occupy the reduced-capacity models. Subsequent capacity-mixed training adapts the reduced-capacity models. Spectral channels are not reordered.

\subsection[Capacity-mixed micro-batches]{Capacity-mixed micro-batches}
\label{app:elastic-mix}

In the default recipe, all micro-batches use full capacity during the first $250$M training tokens. The adaptive sampling variants in Appendix~\ref{app:el-constants} use the activation schedules specified there. After this warm-up, each optimizer step reserves a fraction
$f=0.75$
of its micro-batches for the full-capacity tier. Each remaining micro-batch independently draws uniformly from the ten training capacities above, including full capacity. This sampling operation is budget dropout. The expected full-capacity share after warm-up is therefore
\[
0.75+\frac{0.25}{10}=0.775.
\]

For a reduced-capacity micro-batch, each retained feed-forward weight tensor is evaluated as
\[
\widehat W
=
\alpha_{\mathrm{ff}}W+
(1-\alpha_{\mathrm{ff}})
\operatorname{sg}(W),
\qquad
\alpha_{\mathrm{ff}}=0.5.
\]
Since $\widehat W=W$ in the forward pass, this transformation leaves activations unchanged. Under differentiation, the parameter gradient contributed by the reduced-capacity branch is multiplied by $\alpha_{\mathrm{ff}}$, while the gradient with respect to the layer input is unchanged. The scaling applies only to retained feed-forward weights. Mixer and attention parameters use their ordinary gradients. Compute savings come from evaluating the smaller retained tensor slices rather than from the gradient scaling itself. Reserved full-capacity micro-batches ensure that every parameter group participates in each optimizer step.

\subsection[Distillation from the full model]{Distillation from the full model}
\label{app:elastic-distill}

During pretraining, a reduced-capacity micro-batch uses the full-tier distillation objective of Equation~\ref{eq:cross_capacity_distillation}:
\begin{equation}
\mathcal L_T
=
\mathrm{CE}(p_T,\mathrm{id}_{t+1})
+
0.5\,
\mathrm{KL}
\bigl(
\operatorname{sg}[p_{T_{10}}]
\,\|\,p_T
\bigr).
\label{eq:el-distill}
\end{equation}
The teacher is the current full-capacity tier evaluated on the same tokens. Its prediction is stop-gradient, so the distillation term updates the reduced-capacity predictor without backpropagating through the teacher distribution. Each reduced-capacity micro-batch therefore evaluates both the full-capacity teacher and the retained student. Full-capacity micro-batches use the next-token cross-entropy term alone.

\subsection[Chunked loss evaluation]{Chunked loss evaluation}
\label{app:elastic-chunkce}

Cross-entropy and distillation are sums over token positions and can therefore be evaluated in contiguous position chunks while retaining the complete vocabulary within each chunk. Under bf16 autocast, each chunk computes logits in bf16 and log-softmax and loss accumulation in fp32. Activation checkpointing recomputes the chunk during backward. Weighting chunk losses by their valid-token counts recovers the same full-sequence objective and gradients while reducing peak logit storage. This optional path uses a fixed 4096-token loss chunk when enabled. The main 370M run and shared refinement use full-logit loss evaluation.

\subsection[One optimizer step]{One optimizer step}
\label{app:elastic-algo}

Algorithm~\ref{alg:elastic-step} summarizes one step of the default capacity-mixed pretraining recipe. Here $N_r$ is the valid-token count of micro-batch $r$, $N_{\mathrm{tok}}$ is the global valid-token count, $g_{\mathrm{acc}}$ is the accumulated parameter gradient, $N_{\mathrm{warm}}=250\mathrm{M}$, $j_{\mathrm{max}}$ is the final pretraining step, and $\bar a_{\ell,h}^{\,2}$ is the running squared activation of FFN unit $h$. $\operatorname{Slice}$ returns parameter views tied to $\Theta$, and $\operatorname{PermuteFFN}$ applies the same unit permutation to the FFN parameters, optimizer state, and running activation statistics.

\begin{algorithm}[H]
\small
\DontPrintSemicolon
\SetAlgoVlined
\caption{Capacity-mixed pretraining}
\label{alg:elastic-step}
\KwIn{$\Theta,\Omega$, micro-batches $\{(x_r,\mathrm{id}_r,N_r)\}_{r=1}^{M}$, reserved full slots $\mathcal F$, training capacities $\mathcal T$, tokens seen $n$, step $j$}
\KwOut{updated parameters $\Theta$, optimizer state $\Omega$, and sampled capacities $\{T_r\}_{r=1}^{M}$}

$g_{\mathrm{acc}}\gets0$\;
$N_{\mathrm{tok}}\gets\operatorname{AllReduceSum}(\sum_{r=1}^{M}N_r)$\;

\For{$r=1,\ldots,M$}{
  \eIf{$n<N_{\mathrm{warm}}$ or $r\in\mathcal F$}{
    $T_r\gets T_{10}$\;
  }{
    $T_r\sim\operatorname{Sample}(\mathcal T)$\;
  }

  \eIf{$T_r=T_{10}$}{
    $p_r\gets\operatorname{Model}(x_r;\Theta,T_{10})$\;
    $\ell_r\gets\mathrm{CE}(p_r,\mathrm{id}_r)$\;
  }{
    $p_{\mathrm{full}}\gets
    \operatorname{sg}[
    \operatorname{Model}(x_r;\Theta,T_{10})
    ]$\;

    $\Theta_r\gets
    \operatorname{Slice}
    (\Theta,K_{T_r},n_{\mathrm{ff},T_r})$\;

    $\widehat\Theta_r\gets\Theta_r$\;

    \ForEach{retained FFN weight $W$ in $\Theta_r$}{
      $\widehat\Theta_r[W]\gets
      \alpha_{\mathrm{ff}}W+
      (1-\alpha_{\mathrm{ff}})
      \operatorname{sg}(W)$\;
    }

    $p_r\gets
    \operatorname{Model}
    (x_r;\widehat\Theta_r,T_r)$\;

    $\ell_r\gets
    \mathrm{CE}(p_r,\mathrm{id}_r)
    +
    0.5\,
    \mathrm{KL}
    (p_{\mathrm{full}}\|p_r)$\;
  }

  $g_{\mathrm{acc}}\gets
  g_{\mathrm{acc}}
  +
  (N_r/N_{\mathrm{tok}})
  \nabla_\Theta\ell_r$\;
}

$g_{\mathrm{acc}}\gets
\operatorname{AllReduceSum}(g_{\mathrm{acc}})$\;

$(\Theta,\Omega)\gets
\operatorname{AdamW}
(\Theta,\Omega,\operatorname{Clip}(g_{\mathrm{acc}},1))$\;

\If{$j>0$, $j\bmod500=0$, and $j<0.8j_{\mathrm{max}}$}{
  \ForEach{FFN layer $\ell$}{
    $I_{\ell,h}\gets
    \operatorname{AllReduceMean}
    (
    \bar a_{\ell,h}^{\,2}
    \|W^{\mathrm{down}}_{\ell,:,h}\|_2^2
    )$\;

    $\pi_\ell\gets
    \operatorname{argsort}_{\downarrow}(I_\ell)$\;

    $(\Theta_\ell,\Omega_\ell,\bar a_\ell^{\,2})
    \gets
    \operatorname{PermuteFFN}
    (\Theta_\ell,\Omega_\ell,\bar a_\ell^{\,2};\pi_\ell)$\;
  }
}

\Return{$\Theta,\Omega,\{T_r\}_{r=1}^{M}$}\;
\end{algorithm}

\subsection[Standalone export]{Standalone export}
\label{app:elastic-export}

After capacity-mixed pretraining, shared refinement (SR) continues joint training of the full model and sampled reduced-capacity tiers. It combines their label losses with total-variation distillation from the full prediction:
\[
\mathcal L_{\mathrm{SR}}=\mathrm{CE}(p_{T_{10}},\mathrm{id}_{t+1})
+\mathrm{CE}(p_T,\mathrm{id}_{t+1})
+5\,\mathrm{TV}(\operatorname{sg}[p_{T_{10}}],p_T),
\quad \mathrm{TV}(p,q)=\tfrac12\sum_{v\in\mathcal V}|p(v)-q(v)|.
\]
Here $\mathcal V$ is the vocabulary. The full model receives gradients from its label loss, while the teacher distribution in the distillation term is held fixed. At 370M, one capacity is sampled uniformly per optimizer step from $\{1/16,1/8,1/4,1/2,3/4\}$ and is shared by that step's micro-batches. At 1.5B, each step shuffles the nine reduced deployment tiers $T_1$--$T_9$ and cycles through that order across its 16 micro-batches. Every micro-batch evaluates the full model and its sampled capacity. SR initializes a fresh AdamW optimizer with betas $(0.9,0.95)$, gradient clipping at 1.0 and weight decay 0.1 on tensors with at least two dimensions. This includes embeddings and modal parameter tables. The learning rate warms up for 5\% of the steps and follows cosine decay toward zero. The retained FFN gradient scale remains $\alpha_{\mathrm{ff}}=0.5$. Scale-specific settings are listed in Table~\ref{tab:repro-hparams}.

Every standalone export is obtained by slicing this final checkpoint as specified in Appendix~\ref{app:elastic-slice}. Each export stores only its retained tensors and configuration. The main ESSH results use the final model after the complete procedure in Appendix~\ref{app:repro-config}. The 370M ablation tables also report end-of-pretraining measurements where specified, with the same training stage used across arms within each column.

The export check compares each standalone model with its source shared model at the same tier on two validation batches of $4\times2048$ tokens. The comparison uses the same precision and kernel path for the shared model and standalone export. All ten final 1.5B exports match the shared model evaluated at the same tier bitwise on H200.

\section[Experimental Protocols]{Experimental Protocols}
\label{app:experimental_setup}

This section specifies the data, model configurations, training budgets, evaluation protocols, hardware, and baseline matching used in the experiments. Appendix~\ref{app:elastic-export} defines the final ESSH model and the training stages used in the 370M ablations.

\subsection[Data and tokenization]{Data and tokenization}
\label{app:setup-data}

The 370M experiments use FineWeb-Edu with the GPT-2 tokenizer and a 50,304-entry model vocabulary. Main runs process 7.4B tokens, while mechanism and hyperparameter ablations use 3.7B tokens. The 1.5B run processes 100B FineWeb-Edu pretraining tokens using the Llama-3.1 tokenizer and a 128,256-entry vocabulary. All training sequences are packed to length 2048~\citep{penedo2024fineweb,radford2019gpt2,grattafiori2024llama3}.

The token budgets count processed tokens. Data-preparation commands and stream checksums are provided with the code.

At 370M, the perplexity frontier uses the validation file prepared with the 30B-token stream, while retrieval filler, length sweeps and export checks use the file prepared with the 5B-token stream. Local model comparisons use fixed evaluation inputs within each protocol. The 1.5B experiments use the separately tokenized Llama-3.1 stream. Cross-domain comparisons use fixed token windows within each tokenizer group.

\subsection[Configurations]{Configurations}
\label{app:setup-config}

Table~\ref{tab:setup-config} lists the two language-model configurations. Both use the same selective spectral mixer parameterization and two-rate elastic recipe, while depth, residual width, feed-forward width, attention positions, tokenizer, learning rate, and token budget vary with scale. Complete run-specific launch configurations are included with the experimental artifacts. Distributed training preserves the stated global token batch across worker configurations. During pretraining, raw decay parameters, frequencies and modal readout coefficients are excluded from weight decay, as are embeddings, normalization parameters and biases. The two-dimensional zero-lag coefficient table uses weight decay 0.1. SR uses the optimizer groups defined in Appendix~\ref{app:elastic-export}.

\begin{table}[htbp]
\centering
\small
\setlength{\tabcolsep}{5pt}
\caption{Language-model configurations. Common algorithmic choices are listed alongside scale-specific depth, width, tokenizer, learning rate, and training budget.}
\label{tab:setup-config}
\begin{tabular}{@{}p{0.31\linewidth}p{0.37\linewidth}p{0.24\linewidth}@{}}
\toprule
& 370M & 1.5B \\
\midrule
blocks / width / feed-forward width & 22 / 896 / 4608 & 28 / 2048 / 5632 \\
channels $\bar K$ / modes $m$ / value width $P$ & 32 / 8 / 28 & 32 / 8 / 64 \\
sliding-window layers (zero-based; window 2048, RoPE~\citep{su2021roformer}, head dim. 64) & blocks 12, 17, 21 & blocks 15, 22, 27 \\
tokenizer / embedding & GPT-2, tied & Llama-3.1, tied \\
parameters at full capacity & 370.9M & 1.531B \\
pretraining tokens & 7.4B main / 3.7B ablations & 100B \\
tokens per step & 262,144 & 524,288 \\
peak learning rate, schedule & $1.2\times10^{-3}$, warm-up $1\%$, cosine to $10\%$ & $1.0\times10^{-3}$, same \\
optimizer & AdamW~\citep{loshchilov2019adamw} $(0.9,0.95)$, wd 0.1, clip 1.0, bf16 & same \\
full share $f$ / slice scale $\alpha_{\mathrm{ff}}$ / KL coefficient & 0.75 / 0.5 / 0.5 & same \\
training budget fractions & $1/16$, $3/32$, $1/8$, $5/32$, $3/16$, $1/4$, $3/8$, $1/2$, $3/4$, $1$ & same \\
validation / checkpoint & every 500 steps / 250 (main run) & every 1000 / 200 \\
\bottomrule
\end{tabular}
\end{table}

\subsection[Token budgets and training exposure]{Token budgets and training exposure}
\label{app:setup-budget}

The main 370M models and independently trained specialists use 7.4B pretraining tokens. The dense parents in the recovery controls use 3.7B tokens. Mechanism and hyperparameter ablations use 3.7B tokens with learning-rate schedules defined over that budget. The shared final stage in Table~\ref{tab:repro-hparams} brings total ESSH exposure to 7.67B tokens for the main runs and 3.97B for the ablations.

Post-hoc elastic baselines begin from their pretrained parents and use an additional 0.5B recovery tokens. Their corresponding total data exposures are therefore reported separately as 7.90B or 4.20B tokens. Training-token exposure and measured accelerator time are both reported when comparing the cost of producing multiple deployment sizes.

\subsection[Evaluation protocols]{Evaluation protocols}
\label{app:setup-eval}

\begin{figure}[htbp]
\centering
\includegraphics[width=\linewidth]{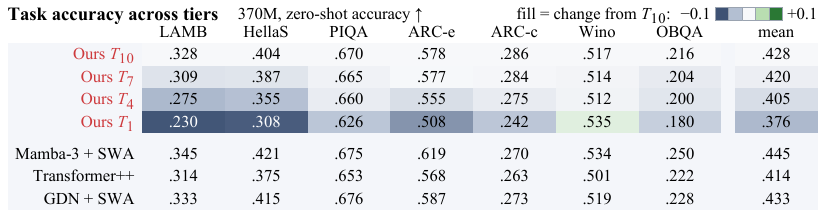}
\caption{Task-level zero-shot accuracy across ESSH capacities ($T_1$, $T_4$, $T_7$, $T_{10}$) and three independently trained full-size references. Cell values report absolute accuracy, and Mean is the unweighted average of the seven tasks. Shading for ESSH tiers represents the absolute score change relative to $T_{10}$ within each task; reference rows are unshaded. Table~\ref{tab:down-370} gives parameter counts and complete numerical results.}
\label{fig:downstream}
\end{figure}

Figure~\ref{fig:downstream} reports absolute task accuracy and its change relative to $T_{10}$. Table~\ref{tab:down-370} gives the corresponding scores and model sizes.

The seven-task mean is the unweighted average of LAMBADA accuracy, HellaSwag normalized accuracy, PIQA accuracy, ARC-Easy accuracy, ARC-Challenge normalized accuracy, WinoGrande accuracy, and OpenBookQA accuracy, evaluated zero-shot with lm-evaluation-harness v0.4.9.1~\citep{gao2024lmeval,paperno2016lambada,zellers2019hellaswag,bisk2020piqa,clark2018arc,sakaguchi2020winogrande,mihaylov2018openbookqa}. Additional evaluations use BoolQ, SIQA, SciQ, COPA, RACE, and LogiQA~\citep{clark2019boolq,sap2019socialiqa,welbl2017sciq,roemmele2011copa,lai2017race,liu2020logiqa}. Scores are reported as fractions unless a table explicitly uses percentages. The published 1.5B rows are transcribed from Table~1 of \citet{lahoti2026mamba3}, which labels OpenBookQA as accuracy. ESSH uses the same metric.

All 370M checkpoints in the main frontier comparison are evaluated on the same held-out shard with the same evaluator: 16 batches of eight length-2048 sequences under bf16 autocast. 

\paragraph{Needle retrieval.}
The NIAH evaluation inserts a six-digit key into a 2048-token filler context at five relative depths. A key pair contains a planted key and its counterfactual alternative. Margin win is the fraction of trials in which the planted key receives a higher conditional score. The NIAH tables use the counterfactual condition, which inserts the alternative key and treats it as the expected answer. The 1280 trials comprise 256 key pairs evaluated at each depth. The primary retrieval metric is teacher-forced full-key accuracy: every key token must be predicted correctly while conditioning on the correct preceding key tokens. First-token top-1 accuracy and candidate-margin win rate are reported separately where used.

Confidence intervals resample the 256 key pairs with replacement while keeping the five depths of each pair together. Reported $95\%$ intervals use 2000 bootstrap replicates.

\paragraph{Cross-domain language modeling.}
Cross-domain evaluation uses fixed text slices from PG-19, the Pile arXiv subset, GitHub code from CodeParrot, CC-News, Proof-Pile, and WikiText-103~\citep{rae2020pg19,gao2020pile,codeparrotgithub,ccnewsdataset,proofpiledataset,merity2017wikitext}. Documents are concatenated with newline separators and partitioned into non-overlapping windows of 2049 tokens. Each window supplies 2048 input tokens and 2048 next-token targets. Perplexity is the exponential of the mean target-token negative log-likelihood. All models within a tokenizer group use the same token windows.

\subsection[Hardware, precision, and the timing harness]{Hardware, precision, and the timing harness}
\label{app:setup-hw}

Training uses bf16 autocast with fp32 master weights and fp32 scan accumulation. Unless a table specifies another device, systems measurements use one A100-SXM4-80GB. ESSH uses bf16 weights and window KV cache with fp32 recurrent state. Baselines use the cache and state dtypes of their evaluated implementations. Memory totals are computed from allocated tensors.

Decode latency measures one autoregressive inference step with state and cache sized for a 2048-token context. \revr{All models use \texttt{torch.compile} on the whole decode step or prefill forward, CUDA graphs for decoding, the same merged gate/up FFN projection and cuDNN attention disabled. ESSH uses the trained SWA layout. The main parameter-matched configurations differ by at most 0.34\%.} Prefill processes the stated prompt at batch one.

The main 1.5B A100 fused-decode measurements use three independent sessions with 50 warm-up and 200 timed iterations, reporting the per-entry session median. \revr{The 370M family sweep uses at least three sessions at batch one, with 50 warm-up and 200 timed iterations per session.} The 370M sweep measures whole-model exported shapes with synthetic weights, while quality is evaluated from trained checkpoints. Each systems table identifies its device and execution settings.

\revr{Matched full-model decode measurements on A100, H200 and B300 use at least three sessions, and full-model prefill values are session medians.} Decode uses CUDA graphs with 50 warm-ups and 200 timed steps. Prefill uses three warm-ups and five timed calls. Baselines follow the same per-device protocol. Compact measurements with one session are marked in their table captions. Ratios use unrounded measurements.

\subsection[Baselines and parameter matching]{Baselines and parameter matching}
\label{app:setup-baselines}

The baselines address three comparison questions: full-capacity architecture quality, train-once compact families, and the quality obtainable from a dedicated training run at one deployment size. Parameter counts include embeddings and refer to the evaluated configuration. Published results are identified separately from locally trained or locally evaluated models. The 1.5B published quality rows come from the 100B-token study of \citet{lahoti2026mamba3}. Official release cards report 1.49B~\citep{mamba3siso2026card} and 1.50B~\citep{mamba3mimo2026card} parameters for Mamba-3 SISO and MIMO, respectively. The remaining published rows retain the source's nominal $\sim1.5$B size. The systems configurations are listed separately in Table~\ref{tab:sys-matched-config}. In the systems tables, Mamba-2/3 + FFN denote the official recurrent mixers alternating with residual feed-forward blocks. The systems Transformer++ uses the same 28-layer, width-2048 comparison setup. Our Transformer++ baseline uses causal full attention with RoPE, RMSNorm, and SwiGLU~\citep{su2021roformer,NEURIPS2019_1e8a1942,shazeer2020glu}.

\begin{table}[htbp]
\centering
\small
\setlength{\tabcolsep}{4pt}
\caption{Baseline roles and matching rules. Each method keeps its stated architecture and training budget. Provenance markers are defined in Appendix~\ref{app:setup-reading}.}
\label{tab:setup-baselines}
\begin{tabular}{@{}p{0.28\linewidth}p{0.25\linewidth}p{0.41\linewidth}@{}}
\toprule
{\raggedright Model\par} & {\raggedright Used in\par} & {\raggedright Matching rule\par} \\
\midrule
{\raggedright Transformer++, Gated DeltaNet, Mamba-2, Mamba-3 (SISO, MIMO), 1.5B, published\refpub\par} & {\raggedright Section~\ref{sec:exp-1p5b}, Appendix~\ref{app:complete_main_results}\par} & {\raggedright published source configurations and source-evaluation metrics; same corpus, tokenizer, context, and token budget where reported\par} \\

{\raggedright Transformer++, own, 370M\par} & {\raggedright Section~\ref{sec:exp-frontier}, Appendices~\ref{app:complete_main_results}, \ref{app:cross_domain}\par} & {\raggedright same depth, data, token budget, and schedule; parameter count within $2\%$\par} \\

{\raggedright Gated DeltaNet + SWA, own, 370M\par} & {\raggedright same\par} & {\raggedright parameter count within $2\%$; its own depth, width, and SWA configuration\par} \\

{\raggedright Mamba-3 + SWA\refhyb, own, 370M\par} & {\raggedright same\par} & {\raggedright dedicated full-capacity training; parameter count within $2\%$, with its own backbone configuration\par} \\

{\raggedright MatFormer, own, 370M\par} & {\raggedright Section~\ref{sec:exp-frontier}, Appendix~\ref{app:compression_baselines}\par} & {\raggedright Transformer++ backbone with nested FFN widths and four sampled training capacities\par} \\

{\raggedright MatMamba, own, 370M\par} & {\raggedright same\par} & {\raggedright Mamba-2 backbone with head-prefix nesting and four sampled capacities\par} \\

{\raggedright Post-hoc elastic, Mamba-3 + SWA parent, own, 370M\par} & {\raggedright Appendix~\ref{app:compression_baselines}\par} & {\raggedright post-hoc ranking and recovery from the pretrained reference hybrid, targeting the $T_4$ and $T_7$ deployment sizes\par} \\

{\raggedright Post-hoc elastic, dense-control parent, own, 370M\par} & {\raggedright same\par} & {\raggedright same recovery procedure applied to the ESSH dense-control architecture\par} \\

{\raggedright Dense control, own, 370M\par} & {\raggedright Section~\ref{sec:exp-ablations}, Appendices~\ref{app:elastic_ablations}, \ref{app:compression_baselines}\par} & {\raggedright ESSH trained only at full capacity, then truncated and recovered for 2060 steps as in Table~\ref{tab:comp}\par} \\

{\raggedright Label-only prune-and-recover, own, 370M\par} & {\raggedright Appendix~\ref{app:compression_baselines}\par} & {\raggedright fixed-shape recovery from the 3.7B dense control using label cross-entropy; no distillation or full-model label-loss term\par} \\

{\raggedright Dense specialists\refspec, own, 370M\par} & {\raggedright same\par} & {\raggedright exact $T_1$, $T_4$, and $T_7$ export architectures trained independently for 7.4B tokens\par} \\

{\raggedright Transformer and Mamba-3 at Zoology size~\citep{arora2023zoology}, own\par} & {\raggedright Appendix~\ref{app:synthetic}\par} & {\raggedright same generator, depth, residual width, training steps, and learning-rate grid; parameter counts reported separately\par} \\

{\raggedright CNN, HyenaDNA, Mamba, Caduceus, published\refpub~\citep{gresova2023genomicbenchmarks,schiff2024caduceus}\par} & {\raggedright Appendix~\ref{app:dna}\par} & {\raggedright published benchmark results at reported model sizes; bidirectional models identified separately\par} \\
{\raggedright Four independent Transformer++ sizes, 370M study\par} & {\raggedright Figure~\ref{fig:main_frontier}(b,c), Table~\ref{tab:external-four-sizes}\par} & {\raggedright 7.4B pretraining tokens per size; actual parameter counts near ESSH $T_1/T_4/T_7/T_{10}$\par} \\

{\raggedright Mamba-2/3 + FFN and Transformer++, 1.53B systems\par} & {\raggedright Figure~\ref{fig:exp-deploy}, Appendix~\ref{app:cross-device-final}\par} & {\raggedright matched depth, residual width and parameter counts; configurations in Table~\ref{tab:sys-matched-config}\par} \\

{\raggedright\leavevmode\revr{ES-SSM mixer, own, 370M}\par} & {\raggedright\leavevmode\revr{Section~\ref{sec:exp-ablations}, Appendix~\ref{app:rev-essm}}\par} & {\raggedright\leavevmode\revr{original LTI mixer from the released code in a 22-layer, width-512 hybrid, with matched total parameters and 3.7B-token dense pretraining}\par} \\

{\raggedright\leavevmode\revr{Same-recipe controls, own, 370M}\par} & {\raggedright\leavevmode\revr{Appendices~\ref{app:rev-init-s4d}, \ref{app:rev-same-recipe}, \ref{app:el-nesting}}\par} & {\raggedright\leavevmode\revr{LTI, real-pole, head-nested Mamba-3 + SWA, MatMamba and FFN-only arms trained with the ESSH recipe for 3.7B tokens, common refinement and evaluator}\par} \\

\bottomrule
\end{tabular}
\end{table}

\begin{table}[htbp]
\centering
\footnotesize
\setlength{\tabcolsep}{4pt}
\caption{Configurations used for the approximately 370M quality comparison. Attention positions are zero based. Pretraining KL denotes the full-tier distillation coefficient. \revr{MatFormer and MatMamba are elastic families (one run, several exports), while the other baselines are single dense models.}}
\label{tab:baseline-configs-v3}
\begin{tabular}{@{}lrrrrlc@{}}
\toprule
Model & \shortstack{Params\\(M)} & Layers & Width & FFN & \shortstack{Attention\\blocks} & \shortstack{Pretrain\\KL} \\
\midrule
\ours{} & 370.9 & 22 & 896 & 4608 & 12, 17, 21 & 0.5 \\
Mamba-3 + SWA & 377.1 & 24 & 1024 & 2304 & 10, 22 & -- \\
Gated DeltaNet + SWA & 378.1 & 24 & 1024 & 2432 & 10, 22 & -- \\
Transformer++ & 365.6 & 22 & 896 & 4224 & All 22 & -- \\
MatFormer & 365.6 & 22 & 896 & 4224 & All 22 & 0 \\
MatMamba & 369.9 & 48 & 1024 & -- & None & 0 \\
\bottomrule
\end{tabular}
\end{table}

\subsection[Post-hoc elastification protocol]{Post-hoc elastification protocol}
\label{app:setup-posthoc}

The post-hoc study uses two pretrained parents: Mamba-3 + SWA trained for 7.4B tokens and the ESSH dense control trained for 3.7B tokens. We adapt the ranking-and-distillation procedure of Nemotron Elastic~\citep{taghibakhshi2025nemotronelastic} to the fixed-depth comparison used here.

A 2M-token calibration subset supplies importance scores for feed-forward units in both parents. For Mamba-3 + SWA, it also scores recurrent heads and attention heads before constructing nested masks, following group-aware pruning principles~\citep{taghibakhshi2025minitronssm}. For the dense ESSH parent, spectral channels retain their original ordering and all attention heads are preserved. Depth and residual width remain fixed.

A frozen parent supervises the masked model for 0.5B recovery tokens using forward KL and label cross-entropy, both with coefficient one. Masks are sampled uniformly within each experiment. The dense-control recovery uses five masks: full capacity, approximately 75\% and 50\% of the parameters, and targets near 240.3M and 162.1M. The Mamba-3 comparisons use three masks: full capacity and the two targets near 185.3M and 274.3M. Recovery uses a $1\%$ warm-up followed by constant learning rate $10^{-4}$, with context length 2048. Compact comparisons target the $T_4$ and $T_7$ deployment sizes, with actual masked parameter counts reported for each export. Appendix~\ref{app:comp-attention-preservation} provides a matched control that preserves the parent attention layers during recovery.

\subsection[Seeds]{Seeds}
\label{app:setup-seeds}

Unless a table specifies another seed, the 370M language-model results use seed 197. The elastic-family comparison includes ESSH seeds \revr{197, 313 and 511}, MatFormer seeds 197, 313, and 511, and MatMamba seeds 197 and 313. The 3.7B ESSH control uses seeds 197, 313 and 511, and the dense controls use two seeds per pretraining budget. The initialization seeds are listed in Table~\ref{tab:new-init-seeds}.

\revr{Initialization seed sets are listed with each comparison, including the repetitions in Appendix~\ref{app:rev-init-s4d}. Seed-level values and ranges are descriptive and are not confidence intervals.}

\subsection[Baselines, row markers, and cost accounting]{Baselines, row markers, and cost accounting}
\label{app:setup-reading}

Where used, bold marks the best metric under comparable model sizes and evaluation settings, including ties at the displayed precision. Capacity sweeps are unranked. Recovery and ablation comparisons keep the target tier fixed. Unmeasured entries are marked with a dash.

Markers distinguish provenance: $\star$ identifies the locally trained Mamba-3 + SWA reference hybrid, $\dagger$ a published result, $\ddagger$ a public checkpoint evaluated with our harness, and $\diamond$ an independently trained specialist.

The primary quality frontier uses one final ESSH checkpoint for both full and compact tiers. Tables using raw checkpoints or other training stages identify them explicitly. Deployment tier and actual parameter count are reported together for compact models, and Table~\ref{tab:el-map} gives each tier's retained channel count.

\section[Complete Main Results]{Complete Main Results}
\label{app:complete_main_results}

This section reports the complete public ESSH capacity family together with the matched controls and reference models used in the main comparisons.

\subsection[The ten tiers at 370M]{The ten tiers at 370M}
\label{app:main-370-tiers}

The ten tiers trace a smooth PPL--size frontier, while task accuracy and retrieval degrade more slowly across intermediate capacities (Table~\ref{tab:main-370-tiers}). \revr{The table separates two kinds of reference. MatFormer and MatMamba are train-once families like ESSH. Mamba-3+SWA, Gated DeltaNet+SWA and Transformer++ are single models trained at full size and provide full-capacity quality references.} With $26\%$ fewer parameters, $T_7$ changes the seven-task mean from $0.428$ to $0.420$, while PPL increases from $14.71$ to $15.91$. At approximately half the full parameter count, $T_4$ retains a seven-task mean of $0.405$ versus $0.428$ and teacher-forced full-key accuracy of $0.370$ versus $0.389$.

PPL decreases monotonically as capacity increases. The seven-task mean and full-key accuracy are flatter over the intermediate tiers and need not vary monotonically. For example, $T_9$ reaches full-key accuracy $0.394$, compared with $0.389$ at full capacity. The metrics therefore have different capacity sensitivities, so the appropriate deployment tier depends on the target quality measure as well as model size.

\phantomsection\label{app:comp-specialists}
Table~\ref{tab:sharing-control} isolates the quality cost of sharing parameters across capacities within the ESSH architecture. Across all four matched shapes, the seven-task gap between the shared exports and independently trained models is \revr{less than $0.01$}, and the shared full-capacity model is slightly higher on this metric. Independent training nevertheless gives lower PPL at every shape, with the largest difference at $T_1$. Appendix~\ref{app:comp-amort} compares the corresponding training cost of producing the four model sizes.

\begin{table}[htbp]
\centering\scriptsize
\setlength{\tabcolsep}{3pt}
\caption{Complete 370M ESSH family and full-capacity local references. ESSH quality uses the final shared checkpoint after 7.4B-token pretraining and brief shared refinement; local full-model references use 7.4B training tokens. \revr{MatFormer and MatMamba are elastic families trained once, like ESSH, while the remaining references are separate models trained at one size.} Retained fraction $r$ and $\Delta$NLL are relative to ESSH $T_{10}$. Full-key is teacher-forced full-key accuracy. \revr{Latency is a separate B300 batch-one shape benchmark on the exact export architectures with synthetic weights, a 2048-token context, torch.compile and CUDA graphs, the same policy as Tables~\ref{tab:new-frontier-matformer} and~\ref{tab:new-frontier-matmamba}.}}
\label{tab:main-370-tiers}
\label{tab:exp-370}
\label{tab:new-frontier-essh}
\begin{tabular*}{\linewidth}{@{\extracolsep{\fill}}lrrrrrrr@{}}
\toprule
Model / tier & \shortstack{Params\\(M)} & $r$ & PPL $\downarrow$ & $\Delta$NLL $\downarrow$ & \shortstack{7-task\\$\uparrow$} & \shortstack{Full key\\$\uparrow$} & \shortstack{ms/token\\$\downarrow$} \\
\midrule
\ours{}, $T_1$ & 92.1 & 0.248 & 23.61 & 0.473 & 0.376 & 0.241 & \revr{0.71} \\
\ours{}, $T_2$ & 118.9 & 0.321 & 20.04 & 0.309 & 0.388 & 0.342 & \revr{0.63} \\
\ours{}, $T_3$ & 154.7 & 0.417 & 18.45 & 0.227 & 0.402 & 0.355 & \revr{0.79} \\
\ours{}, $T_4$ & 185.3 & 0.500 & 17.68 & 0.184 & 0.405 & 0.370 & \revr{0.80} \\
\ours{}, $T_5$ & 214.5 & 0.578 & 16.98 & 0.144 & 0.413 & 0.366 & \revr{0.80} \\
\ours{}, $T_6$ & 248.8 & 0.671 & 16.25 & 0.100 & 0.417 & 0.362 & \revr{0.86} \\
\ours{}, $T_7$ & 274.3 & 0.740 & 15.91 & 0.078 & 0.420 & 0.366 & \revr{0.86} \\
\ours{}, $T_8$ & 308.6 & 0.832 & 15.34 & 0.042 & 0.420 & 0.375 & \revr{0.82} \\
\ours{}, $T_9$ & 337.9 & 0.911 & 15.08 & 0.025 & 0.424 & 0.394 & \revr{0.84} \\
\midrule
\ours{}, $T_{10}$ (full) & 370.9 & 1.000 & 14.71 & 0.000 & 0.428 & 0.389 & \revr{0.83} \\
MatFormer, full & 365.6 & -- & 16.80 & -- & 0.411 & 0.266 & -- \\
MatMamba, full & 369.9 & -- & 16.98 & -- & 0.404 & 0.004 & -- \\
\midrule
Mamba-3 + SWA\refhyb & 377.1 & -- & 14.10 & -- & 0.445 & 0.230 & -- \\
Gated DeltaNet + SWA & 378.1 & -- & 14.31 & -- & 0.433 & 0.100 & -- \\
Transformer++ & 365.6 & -- & 15.88 & -- & 0.414 & 0.350 & -- \\
\bottomrule
\end{tabular*}
\end{table}

\begin{table}[htbp]
\centering\small
\setlength{\tabcolsep}{3pt}
\caption{Shared versus independently trained ESSH models at identical deployment shapes. The shared exports come from one 7.67B-token training run, while each independent model is trained separately for 7.4B tokens at the corresponding shape.}
\label{tab:sharing-control}
\begin{tabular*}{\linewidth}{@{\extracolsep{\fill}}crrrrr@{}}
\toprule
& & \multicolumn{2}{c}{PPL $\downarrow$} & \multicolumn{2}{c}{7-task $\uparrow$}\\
Tier & M params & Shared & Independent & Shared & Independent\\
\midrule
$T_1$ & 92.1 & 23.61 & \textbf{19.79} & 0.376 & \textbf{0.385}\\
$T_4$ & 185.3 & 17.68 & \textbf{16.25} & 0.405 & \textbf{0.412}\\
$T_7$ & 274.3 & 15.91 & \textbf{15.10} & 0.420 & \textbf{0.424}\\
$T_{10}$ & 370.9 & 14.71 & \textbf{14.44} & \textbf{0.428} & 0.425\\
\bottomrule
\end{tabular*}
\end{table}

\subsection[The 1.5B model]{The 1.5B model}
\label{app:main-1p5b}

Throughout the 1.5B study, ESSH denotes the final model produced by the complete training procedure in Appendix~\ref{app:elastic-export}. Every deployment tier uses that model.

The family extends the size--quality trade-off to the larger backbone and vocabulary. At $T_7$, 23.7\% fewer parameters retain a seven-task mean of 53.01\% compared with 55.03\% at full capacity. The 0.828B $T_4$ retains 50.72\% with 45.9\% fewer parameters, while $T_1$ reaches 43.83\% at 0.474B. These intermediate models provide substantial parameter savings before the larger accuracy reduction at the smallest tier.

The smallest export retains 31.0\% of the full model's parameters, compared with 24.8\% at 370M, because the larger vocabulary increases the fixed embedding cost. Tables~\ref{tab:scale-reference} and~\ref{tab:main-1p5b-tasks} give the selected tiers, and Table~\ref{tab:main-tiers} reports all ten. Published full-model comparisons remain in Table~\ref{tab:exp-1p5b}(a). \revr{Table~\ref{tab:exp-1p5b}(d) and Table~\ref{tab:compact-published-tasks} place $T_1$ and $T_4$ beside published models, each trained separately at about 440M or 880M.}

\begin{table}[!htbp]
\centering\footnotesize
\setlength{\tabcolsep}{3pt}
\caption{Final 1.5B ESSH at four deployment tiers. Accuracy and seven-task means are percentages. PPL uses the 1.5B validation stream. Published full-model references are in Table~\ref{tab:exp-1p5b}(a).}
\label{tab:scale-reference}
\begin{tabular*}{\linewidth}{@{\extracolsep{\fill}}lrrrrr@{}}
\toprule
Tier & Params (B) & FW-Edu PPL $\downarrow$ & LAMB PPL $\downarrow$ & LAMB acc. $\uparrow$ & 7-task $\uparrow$ \\
\midrule
\multicolumn{6}{l}{ESSH}\\
$T_{10}$ & 1.53 & 10.11 & 12.96 & 46.8 & 55.0 \\
$T_{7}$ & 1.17 & 10.97 & 15.39 & 44.9 & 53.0 \\
$T_{4}$ & 0.83 & 12.26 & 18.27 & 42.1 & 50.7 \\
$T_{1}$ & 0.47 & 16.49 & 36.87 & 34.0 & 43.8 \\
\bottomrule
\end{tabular*}
\end{table}
\begin{table}[htbp]
\centering\small
\setlength{\tabcolsep}{2pt}
\caption{\editblue{Compact ESSH exports and published models trained independently at nearby sizes, all with 100B pretraining tokens. ESSH values are from the original shared 1.5B run; $\dagger$ denotes published results from \citet{lahoti2026mamba3}. Task scores are percentages. PPL uses source-specific validation shards and is not ranked. Table~\ref{tab:exp-1p5b}(d) summarizes the seven-task means and gaps.}}
\label{tab:compact-published-tasks}
\begin{tabular*}{\linewidth}{@{\extracolsep{\fill}}lrrrrrrrrrr@{}}
\toprule
Model & Params (B) & PPL $\downarrow$ & LAMB & HellaS & PIQA & ARC-e & ARC-c & Wino & OBQA & Mean\\
\midrule
\rev{\textbf{ESSH} $T_4$} & \rev{0.83} & \rev{12.26} & \rev{42.1} & \rev{51.9} & \rev{69.4} & \rev{67.5} & \rev{35.1} & \rev{58.6} & \rev{\textbf{30.4}} & \rev{50.7} \\
\rev{Transformer++\refpub} & \rev{$\mathord{\sim}0.88$} & \rev{11.42} & \rev{44.7} & \rev{57.2} & \rev{72.6} & \rev{71.6} & \rev{39.2} & \rev{57.7} & \rev{26.8} & \rev{52.8} \\
\rev{Gated DeltaNet\refpub} & \rev{$\mathord{\sim}0.88$} & \rev{11.37} & \rev{47.6} & \rev{57.3} & \rev{73.3} & \rev{71.4} & \rev{38.7} & \rev{58.8} & \rev{28.6} & \rev{53.7} \\
\rev{Mamba-2\refpub} & \rev{$\mathord{\sim}0.88$} & \rev{11.35} & \rev{45.0} & \rev{58.1} & \rev{72.5} & \rev{72.3} & \rev{38.7} & \rev{56.8} & \rev{30.2} & \rev{53.4} \\
\rev{Mamba-3 SISO\refpub} & \rev{$\mathord{\sim}0.88$} & \rev{11.23} & \rev{47.2} & \rev{58.8} & \rev{73.6} & \rev{72.7} & \rev{40.2} & \rev{58.4} & \rev{30.0} & \rev{54.4} \\
\rev{Mamba-3 MIMO\refpub} & \rev{$\mathord{\sim}0.88$} & \rev{11.11} & \rev{\textbf{49.5}} & \rev{\textbf{59.2}} & \rev{\textbf{73.7}} & \rev{\textbf{74.7}} & \rev{\textbf{41.2}} & \rev{\textbf{59.9}} & \rev{28.6} & \rev{\textbf{55.3}} \\
\midrule
\rev{\textbf{ESSH} $T_1$} & \rev{0.47} & \rev{16.49} & \rev{34.0} & \rev{39.6} & \rev{66.5} & \rev{59.8} & \rev{29.5} & \rev{52.7} & \rev{24.6} & \rev{43.8} \\
\rev{Transformer++\refpub} & \rev{$\mathord{\sim}0.44$} & \rev{13.03} & \rev{41.7} & \rev{50.5} & \rev{69.9} & \rev{67.6} & \rev{34.6} & \rev{\textbf{56.7}} & \rev{26.0} & \rev{49.6} \\
\rev{Gated DeltaNet\refpub} & \rev{$\mathord{\sim}0.44$} & \rev{13.01} & \rev{41.9} & \rev{50.9} & \rev{70.0} & \rev{67.0} & \rev{34.6} & \rev{56.1} & \rev{27.6} & \rev{49.7} \\
\rev{Mamba-2\refpub} & \rev{$\mathord{\sim}0.44$} & \rev{13.00} & \rev{40.8} & \rev{51.7} & \rev{70.6} & \rev{68.8} & \rev{35.0} & \rev{54.1} & \rev{26.0} & \rev{49.6} \\
\rev{Mamba-3 SISO\refpub} & \rev{$\mathord{\sim}0.44$} & \rev{12.87} & \rev{40.2} & \rev{51.7} & \rev{\textbf{71.9}} & \rev{68.9} & \rev{34.4} & \rev{55.8} & \rev{26.0} & \rev{49.8} \\
\rev{Mamba-3 MIMO\refpub} & \rev{$\mathord{\sim}0.44$} & \rev{12.72} & \rev{\textbf{43.4}} & \rev{\textbf{52.8}} & \rev{70.8} & \rev{\textbf{69.6}} & \rev{\textbf{35.6}} & \rev{56.3} & \rev{\textbf{28.4}} & \rev{\textbf{51.0}} \\
\bottomrule
\end{tabular*}
\end{table}

Table~\ref{tab:main-1p5b-tasks} separates the task-level responses to capacity reduction. Table~\ref{tab:main-tiers} combines model quality with the measured deployment costs at the corresponding sizes.

\begin{table}[!htbp]
\centering\footnotesize
\setlength{\tabcolsep}{3pt}
\caption{Seven-task accuracy (\%) for the final 1.5B ESSH model. HellaSwag and ARC-Challenge use normalized accuracy; other columns use accuracy. Published full-model results are listed in Table~\ref{tab:exp-1p5b}(a).}
\label{tab:main-1p5b-tasks}
\begin{tabular*}{\linewidth}{@{\extracolsep{\fill}}lrrrrrrrr@{}}
\toprule
Model / tier & Params (B) & LAMB & HellaS & PIQA & ARC-e & ARC-c & Wino & OBQA \\
\midrule
ESSH, $T_{10}$ & 1.53 & 46.8 & 60.6 & 72.9 & 73.7 & 40.7 & 58.3 & 32.2 \\
ESSH, $T_{7}$ & 1.17 & 44.9 & 56.6 & 71.9 & 70.6 & 38.1 & 58.4 & 30.6 \\
ESSH, $T_{4}$ & 0.83 & 42.1 & 51.9 & 69.4 & 67.5 & 35.1 & 58.6 & 30.4 \\
ESSH, $T_{1}$ & 0.47 & 34.0 & 39.6 & 66.5 & 59.8 & 29.5 & 52.7 & 24.6 \\
\bottomrule
\end{tabular*}
\end{table}

\begin{table}[htbp]
\centering
\small
\setlength{\tabcolsep}{4pt}
\caption{1.5B deployment family. Quality uses one ESSH checkpoint across all ten tiers; runtime uses B300 shape benchmarks. Prefill is batch one at 2K; decode is batch one after 2K. \revr{Both compile the model with torch.compile, and decoding replays the step with CUDA graphs. Decode uses at least three sessions at every tier, and prefill uses at least three sessions at $T_{10}$ and one at $T_1$, $T_4$ and $T_7$.} Persistent memory is state+KV, excluding weights. Selected tiers are summarized in Table~\ref{tab:new-quality-latency-1p5b}(b).}
\label{tab:main-tiers}
\begin{tabular}{@{}lrcccccc@{}}
\toprule
tier & params (B) & PPL $\downarrow$ & 7-task $\uparrow$ & Full key $\uparrow$ & prefill ktok/s $\uparrow$ & decode ms/token $\downarrow$ & memory MB $\downarrow$ \\
\midrule
$T_1$ & 0.474 & 16.49 & 0.438 & 0.370 & \revr{127.6} & \revr{1.25} & 51.15 \\
$T_2$ & 0.575 & 14.08 & 0.462 & 0.384 & \revr{--} & \revr{1.13} & 51.46 \\
$T_3$ & 0.716 & 12.87 & 0.486 & 0.367 & \revr{--} & \revr{1.30} & 51.87 \\
$T_4$ & 0.828 & 12.26 & 0.507 & 0.379 & \revr{113.6} & \revr{1.34} & 52.17 \\
$T_5$ & 0.931 & 11.78 & 0.515 & 0.371 & \revr{--} & \revr{1.31} & 52.38 \\
$T_6$ & 1.076 & 11.23 & 0.526 & 0.323 & \revr{--} & \revr{1.38} & 52.69 \\
$T_7$ & 1.168 & 10.97 & 0.530 & 0.339 & \revr{107.4} & \revr{1.53} & 52.89 \\
$T_8$ & 1.291 & 10.63 & 0.538 & 0.321 & \revr{--} & \revr{1.48} & 53.20 \\
$T_9$ & 1.416 & 10.35 & 0.547 & 0.334 & \revr{--} & \revr{1.47} & 53.40 \\
$T_{10}$ & 1.531 & 10.11 & 0.550 & 0.330 & \revr{102.0} & \revr{1.35} & 53.61 \\
\bottomrule
\end{tabular}
\end{table}

\section[Downstream Language Understanding]{Downstream Language Understanding}
\label{app:downstream_results}

Per-task results complement the aggregate seven-task mean by showing how individual benchmarks respond to capacity reduction.

\subsection[370M per-task results]{370M per-task results}
\label{app:down-tasks}

Table~\ref{tab:down-370} compares individual task scores. Full ESSH exceeds Transformer++ on six of the seven tasks. Mamba-3 + SWA has the highest mean, while ESSH is higher on ARC-Challenge, 0.286 versus 0.270. Capacity reduction affects the tasks differently. At 214.5M parameters, $T_5$ reaches a seven-task mean of $0.413$, compared with $0.414$ for the 365.6M Transformer++, despite using approximately $41\%$ fewer parameters. Their nearly identical means arise from different task profiles: $T_5$ is higher on ARC-Easy and ARC-Challenge and equal on PIQA, while Transformer++ is higher on LAMBADA, HellaSwag, and OpenBookQA. WinoGrande is close to chance for both models. \revr{Seed ranges for these means are in Table~\ref{tab:rev-seed-ranges}: three 7.4B ESSH seeds span 1.28, 0.15, 0.36, and 0.21 points at $T_1$, $T_4$, $T_7$, and $T_{10}$. Two Mamba-3 + SWA seeds differ by 0.56 points.}

Within the ESSH family, LAMBADA and HellaSwag show the clearest upward trend with capacity. PIQA is comparatively flat over the upper tiers: $T_7$ scores $0.665$, only $0.005$ below the full-capacity score of $0.670$. ARC-Easy and ARC-Challenge vary less regularly across the intermediate tiers. Reducing from $T_{10}$ to $T_1$ changes LAMBADA and HellaSwag accuracy by 0.098 and 0.096, respectively, while each of the other five tasks changes by at most 0.070.

\begin{table}[htbp]
\centering
\scriptsize
\setlength{\tabcolsep}{3pt}
\caption{Zero-shot accuracy for the 370M comparison. ESSH rows use the final shared checkpoint and correspond to the ten deployment tiers; local reference models are trained independently\revr{, one model per row}. HellaSwag and ARC-Challenge use normalized accuracy. Scores are reported to three decimals.}
\label{tab:down-370}
\begin{tabular}{@{}lrcccccccc@{}}
\toprule
& & \multicolumn{8}{c}{accuracy $\uparrow$} \\
\cmidrule(lr){3-10}
Model & params (M) & LAMB acc & HellaS & PIQA & ARC-e & ARC-c & Wino & OBQA & mean \\
\midrule
\ours{}, $T_{1}$ & 92.1 & 0.230 & 0.308 & 0.626 & 0.508 & 0.242 & 0.535 & 0.180 & 0.376 \\
\ours{}, $T_{2}$ & 118.9 & 0.253 & 0.322 & 0.644 & 0.540 & 0.259 & 0.515 & 0.186 & 0.388 \\
\ours{}, $T_{3}$ & 154.7 & 0.269 & 0.344 & 0.656 & 0.561 & 0.275 & 0.516 & 0.190 & 0.402 \\
\ours{}, $T_{4}$ & 185.3 & 0.275 & 0.355 & 0.660 & 0.555 & 0.275 & 0.512 & 0.200 & 0.405 \\
\ours{}, $T_{5}$ & 214.5 & 0.293 & 0.365 & 0.653 & 0.574 & 0.282 & 0.513 & 0.210 & 0.413 \\
\ours{}, $T_{6}$ & 248.8 & 0.305 & 0.379 & 0.660 & 0.574 & 0.286 & 0.509 & 0.206 & 0.417 \\
\ours{}, $T_{7}$ & 274.3 & 0.309 & 0.387 & 0.665 & 0.577 & 0.284 & 0.514 & 0.204 & 0.420 \\
\ours{}, $T_{8}$ & 308.6 & 0.316 & 0.391 & 0.670 & 0.574 & 0.285 & 0.503 & 0.204 & 0.420 \\
\ours{}, $T_{9}$ & 337.9 & 0.329 & 0.399 & 0.671 & 0.574 & 0.277 & 0.519 & 0.202 & 0.424 \\
\ours{}, $T_{10}$ & 370.9 & 0.328 & 0.404 & 0.670 & 0.578 & 0.286 & 0.517 & 0.216 & 0.428 \\
\midrule
Mamba-3 + SWA\refhyb\ (reference hybrid) & 377.1 & 0.345 & 0.421 & 0.675 & 0.619 & 0.270 & 0.534 & 0.250 & 0.445 \\
Transformer++ & 365.6 & 0.314 & 0.375 & 0.653 & 0.568 & 0.263 & 0.501 & 0.222 & 0.414 \\
Gated DeltaNet + SWA & 378.1 & 0.333 & 0.415 & 0.676 & 0.587 & 0.273 & 0.519 & 0.228 & 0.433 \\
\bottomrule
\end{tabular}
\end{table}

\begin{table}[htbp]
\begin{revblockr}
\centering\footnotesize
\setlength{\tabcolsep}{4pt}
\revcaption{Seven-task mean accuracy (\%) across training seeds. The 7.4B-token block reports the main study: ESSH is a shared family, whereas Mamba-3 + SWA was trained independently at full size and has no trained compact tiers. The 3.7B-token block compares both shared families under the same ESSH recipe, refinement and evaluator; Mamba-3 head counts match the ESSH sizes (Table~\ref{tab:rev-aligned-m3}). Range is maximum minus minimum. Dashes indicate tiers not trained in that experiment.}{tab:rev-seed-ranges}
\begin{tabular*}{\linewidth}{@{\extracolsep{\fill}}llrrrr@{}}
\toprule
Model & Seed & $T_1$ & $T_4$ & $T_7$ & $T_{10}$ \\
\midrule
ESSH, 7.4B tokens & 197 & 37.56 & 40.45 & 42.00 & 42.84 \\
ESSH, 7.4B tokens & 313 & 36.50 & 40.30 & 41.89 & 43.01 \\
ESSH, 7.4B tokens & 511 & 36.28 & 40.35 & 41.64 & 42.80 \\
\multicolumn{2}{l}{range} & 1.28 & 0.15 & 0.36 & 0.21 \\
\midrule
Mamba-3 + SWA, 7.4B tokens & 197 & -- & -- & -- & 44.50 \\
Mamba-3 + SWA, 7.4B tokens & 313 & -- & -- & -- & 43.94 \\
\multicolumn{2}{l}{range} & & & & 0.56 \\
\midrule
ESSH, 3.7B tokens & 197 & 36.22 & 39.39 & 40.06 & 40.84 \\
ESSH, 3.7B tokens & 313 & 35.54 & 38.58 & 39.72 & 40.60 \\
ESSH, 3.7B tokens & 511 & 36.50 & 39.07 & 40.12 & 41.03 \\
\multicolumn{2}{l}{range} & 0.96 & 0.81 & 0.40 & 0.43 \\
\midrule
Mamba-3 + SWA, 3.7B tokens & 197 & 35.39 & 39.81 & 40.50 & 41.36 \\
Mamba-3 + SWA, 3.7B tokens & 511 & 35.64 & 38.82 & 40.28 & 41.47 \\
\multicolumn{2}{l}{range} & 0.25 & 0.99 & 0.22 & 0.11 \\
\bottomrule
\end{tabular*}
\end{revblockr}
\end{table}

\subsection[1.5B per-task results]{1.5B per-task results}
\label{app:down-tasks-1p5b}

The six additional tasks give a complementary view of capacity selection (Table~\ref{tab:down-1p5b}). At $T_7$, the mean is 0.547 compared with 0.549 at full capacity, and SciQ remains at 0.904 versus 0.911. The intermediate export therefore preserves a similar aggregate score with 23.7\% fewer parameters.

\begin{table}[!htbp]
\centering\footnotesize
\setlength{\tabcolsep}{3pt}
\caption{Additional zero-shot task accuracies (fractions, $\uparrow$) for the final 1.5B ESSH model. Mean averages the six task columns. The seven reference tasks are in Table~\ref{tab:main-1p5b-tasks}.}
\label{tab:down-1p5b}
\label{tab:best-two-stages}
\begin{tabular*}{\linewidth}{@{\extracolsep{\fill}}lrrrrrrrr@{}}
\toprule
Model / tier & Params (B) & BoolQ & SIQA & SciQ & COPA & RACE & LogiQA & Mean \\
\midrule
ESSH, $T_{10}$ & 1.531 & 0.613 & 0.419 & 0.911 & 0.790 & 0.344 & 0.214 & 0.549 \\
ESSH, $T_{7}$ & 1.168 & 0.594 & 0.422 & 0.904 & 0.810 & 0.344 & 0.209 & 0.547 \\
ESSH, $T_{4}$ & 0.828 & 0.596 & 0.400 & 0.890 & 0.740 & 0.344 & 0.198 & 0.528 \\
ESSH, $T_{1}$ & 0.474 & 0.617 & 0.395 & 0.870 & 0.710 & 0.297 & 0.220 & 0.518 \\
\bottomrule
\end{tabular*}
\end{table}

\section[Retrieval and Length]{Retrieval and Length}
\label{app:retrieval_results}

Retrieval is evaluated separately from language-model perplexity. The needle probe, open-book tasks, and synthetic suite test different behaviors and use different scoring rules.

\subsection[Needle recall by tier]{Needle recall by tier}
\label{app:ret-level}

The 2048-token NIAH test measures preference for a planted key over a counterfactual key, first-token prediction, and full-key accuracy (Table~\ref{tab:ret-level}). Full-key accuracy is reported in Table~\ref{tab:exp-1p5b}(b) as a percentage. At 1.5B, the selected ESSH tiers score 0.330--0.379. The 0.828B $T_4$ reaches 0.379 and the 0.474B $T_1$ reaches 0.370, preserving useful key retrieval after substantial parameter reduction.

At 370M, full-key accuracy is 0.370 for the half-size $T_4$ and 0.389 at full capacity, while the smallest $T_1$ reaches 0.241. Moderate reduction retains much of this capability, but more aggressive reduction incurs a larger retrieval loss even though the attention layout is fixed. Model scales use different tokenizers and are compared within scale.

The 370M $T_1$ ranks the planted key above its counterfactual in 0.845 of trials but predicts the full key correctly in 0.241. Relative preference and exact token prediction therefore respond differently to capacity reduction. The full-key metric directly evaluates the latter.

\revr{Table~\ref{tab:rev-ret-seeds} repeats the 2K full-key measurement for three training seeds of the 3.7B control after common refinement. Full-key accuracy varies by up to 0.25 across seeds at a fixed tier.}

\begin{table}[htbp]
\centering
\footnotesize
\setlength{\tabcolsep}{3pt}
\caption{NIAH retrieval at context 2048, using 256 key pairs at five depths. Full-key accuracy uses teacher forcing. First-token accuracy and candidate margin are separate fields. The final column is the half-width of the 95\% item-cluster bootstrap interval for full-key accuracy (2000 resamples of key pairs, keeping their five depths together). The 1.5B block uses the final ESSH model. Full key is also reported in Table~\ref{tab:exp-1p5b}(b), in percent.}
\label{tab:ret-level}
\begin{tabular}{@{}llrcccc@{}}
\toprule
Model & tier & params (M) & margin win $\uparrow$ & first token $\uparrow$ & full key $\uparrow$ & 95\% half-width \\
\midrule
ESSH & $T_{10}$ & 1531.1 & 0.826 & 0.332 & 0.330 & 0.039 \\
ESSH & $T_{7}$ & 1168.3 & 0.831 & 0.341 & 0.339 & 0.039 \\
ESSH & $T_{4}$ & 827.6 & 0.857 & 0.380 & 0.379 & 0.040 \\
ESSH & $T_{1}$ & 474.5 & 0.845 & 0.372 & 0.370 & 0.038 \\
\midrule
\ours{} 370M & $T_{10}$ & 370.9 & 0.826 & 0.389 & 0.389 & 0.036 \\
\ours{} 370M & $T_7$ & 274.3 & 0.824 & 0.366 & 0.366 & 0.034 \\
\ours{} 370M & $T_4$ & 185.3 & 0.828 & 0.370 & 0.370 & 0.035 \\
\ours{} 370M & $T_1$ & 92.1 & 0.845 & 0.242 & 0.241 & 0.025 \\
\midrule
Mamba-3 + SWA\refhyb & -- & 377.1 & 0.738 & 0.241 & 0.230 & 0.023 \\ 
Transformer++ & -- & 365.6 & 0.799 & 0.350 & 0.350 & 0.030 \\ 
Gated DeltaNet + SWA & -- & 378.1 & 0.771 & 0.100 & 0.100 & 0.018 \\ 
\bottomrule
\end{tabular}
\end{table}

\begin{table}[htbp]
\begin{revblockr}
\centering\footnotesize
\setlength{\tabcolsep}{5pt}
\revcaption{Retrieval across training seeds: teacher-forced full-key accuracy on the 2K NIAH test for three seeds of the 370M ESSH control trained for 3.7B tokens, all after the common shared refinement and scored with one evaluator. Mean and [minimum, maximum] over seeds.}{tab:rev-ret-seeds}
\begin{tabular}{@{}ccccc@{}}
\toprule
Tier & Seed 197 & Seed 313 & Seed 511 & Mean [min, max] \\
\midrule
$T_1$ & 0.29 & 0.09 & 0.06 & 0.15 [0.06, 0.29] \\
$T_4$ & 0.26 & 0.01 & 0.10 & 0.12 [0.01, 0.26] \\
$T_7$ & 0.26 & 0.06 & 0.16 & 0.16 [0.06, 0.26] \\
$T_{10}$ & 0.33 & 0.15 & 0.18 & 0.22 [0.15, 0.33] \\
\bottomrule
\end{tabular}\par
\end{revblockr}
\end{table}

\subsection[By depth and by context length]{By depth and by context length}
\label{app:ret-depth}

Table~\ref{tab:ret-depth} varies the key depth at 2048 tokens and the context length at the middle depth for the final 1.5B ESSH model. The depth dependence persists across capacities: accuracy at the most recent placement is 0.633--0.754, compared with 0.004--0.051 at the earliest placement.

{\color{revred}
All keys remain inside the attention window in this test. The depth effect therefore shows that window access alone does not ensure accurate key selection and prediction.
\par}

At the middle depth, shorter contexts give higher full-key accuracy. The following subsection evaluates next-token PPL over longer windows as a separate capability.

\begin{table}[!htbp]
\centering\footnotesize
\setlength{\tabcolsep}{3pt}
\caption{Teacher-forced full-key accuracy for the final 1.5B ESSH model. Left: five needle depths in a 2048-token sequence. Right: sequence lengths with the needle at depth 0.5. Each depth uses 256 trials; 2048 is the same depth-0.5 measurement in both panels.}
\label{tab:ret-depth}
\begin{tabular*}{\linewidth}{@{\extracolsep{\fill}}lrrrrrrrr@{}}
\toprule
& \multicolumn{5}{c}{Depth at 2048 tokens} & \multicolumn{3}{c}{Sequence length (tokens)}\\
\cmidrule(lr){2-6}\cmidrule(lr){7-9}
Tier & 0.1 & 0.3 & 0.5 & 0.7 & 0.9 & 512 & 1024 & 2048 \\
\midrule
\multicolumn{9}{l}{ESSH}\\
$T_{1}$ & 0.004 & 0.242 & 0.355 & 0.496 & 0.754 & 0.742 & 0.582 & 0.355 \\
$T_{4}$ & 0.051 & 0.270 & 0.352 & 0.500 & 0.723 & 0.711 & 0.582 & 0.352 \\
$T_{7}$ & 0.027 & 0.246 & 0.324 & 0.457 & 0.641 & 0.668 & 0.547 & 0.324 \\
$T_{10}$ & 0.035 & 0.227 & 0.309 & 0.445 & 0.633 & 0.664 & 0.512 & 0.309 \\
\bottomrule
\end{tabular*}
\end{table}

\subsection[Perplexity beyond the training length]{Perplexity beyond the training length}
\label{app:ret-length}

The length sweep tests whether capacity reduction introduces additional degradation when evaluation extends beyond the 2048-token training context. Each length from 2K to 16K is evaluated on its own set of 64 non-overlapping windows.

The ordering of the four ESSH tiers is unchanged at every tested length (Table~\ref{tab:ret-length}). PPL also remains in a similar range across the sweep. The 2K and 16K endpoint values are $14.91$ and $15.05$ for $T_{10}$, $16.03$ and $16.23$ for $T_7$, $17.76$ and $17.95$ for $T_4$, and $23.37$ and $23.70$ for $T_1$. Because each length uses a different set of evaluation windows, these endpoint differences include variation in the evaluated text as well as any effect of sequence length.

The compact-to-full gaps remain similar across the tested lengths. For example, the $T_4$--$T_{10}$ PPL gap is $2.85$ at 2K and $2.90$ at 16K, while the $T_7$--$T_{10}$ gap changes from $1.12$ to $1.18$. Capacity reduction therefore does not introduce a marked additional PPL gap over this length range.

The selective recurrence applies the same state update at every position, and the shared attention window remains fixed. The recurrent update and the 2048-token attention window keep the same local operations as context length increases. The Mamba-3 + SWA reference shows a similarly narrow range of PPL values across the same tested lengths. The 1.5B ESSH family likewise preserves its tier ordering from 2K to 16K. Full-model PPL changes from 10.28 to 10.99 and $T_4$ from 12.45 to 13.40. Their gap changes from 2.17 to 2.41. The largest tier remains the strongest language model, while the intermediate export retains a similar additional loss across this length range.

Longer-window PPL measures next-token prediction over longer inputs, separately from the key-retrieval measurements above.

\begin{table}[htbp]
\centering
\footnotesize
\setlength{\tabcolsep}{6pt}
\caption{Perplexity beyond the 2048-token training context. Each length is evaluated on its own set of 64 non-overlapping windows at each scale. Both ESSH scales use their final models.}
\label{tab:ret-length}
\begin{tabular}{@{}lrcccc@{}}
\toprule
& & \multicolumn{4}{c}{validation PPL $\downarrow$ by window length} \\
\cmidrule(lr){3-6}
Model & params (M) & 2048 & 4096 & 8192 & 16384 \\
\midrule
\ours{}, $T_{10}$ & 370.9 & 14.91 & 14.97 & 14.76 & 15.05 \\
\ours{}, $T_7$ & 274.3 & 16.03 & 16.13 & 15.92 & 16.23 \\
\ours{}, $T_4$ & 185.3 & 17.76 & 17.83 & 17.62 & 17.95 \\
\ours{}, $T_1$ & 92.1 & 23.37 & 23.42 & 23.22 & 23.70 \\
\midrule
Mamba-3 + SWA 370M\refhyb\ (reference hybrid) & 377.1 & 14.34 & 14.41 & 14.19 & 14.46 \\
\midrule
\multicolumn{6}{l}{1.5B ESSH}\\
$T_{10}$ & 1531.1 & 10.28 & 9.92 & 10.89 & 10.99 \\
$T_{7}$ & 1168.3 & 11.16 & 10.76 & 11.78 & 11.91 \\
$T_{4}$ & 827.6 & 12.45 & 12.03 & 13.22 & 13.40 \\
$T_{1}$ & 474.5 & 16.57 & 16.20 & 18.22 & 18.81 \\
\bottomrule
\end{tabular}
\end{table}

\subsection[Recall-intensive open-book tasks]{Recall-intensive open-book tasks}
\label{app:ret-based}

We use the six-task recall-intensive suite described by JRT~\citep{arora2024jrt}, extending the extraction and question-answering evaluations in Based~\citep{arora2024based}. All tasks use contains-match accuracy with a 2048-token context. The underlying datasets are SWDE, FDA, SQuAD, TriviaQA, Natural Questions, and DROP~\citep{hao2011swde,arora2023evaporate,rajpurkar2016squad,joshi2017triviaqa,kwiatkowski2019naturalquestions,dua2019drop}.

\begin{figure}[htbp]
\centering
\includegraphics[width=\linewidth]{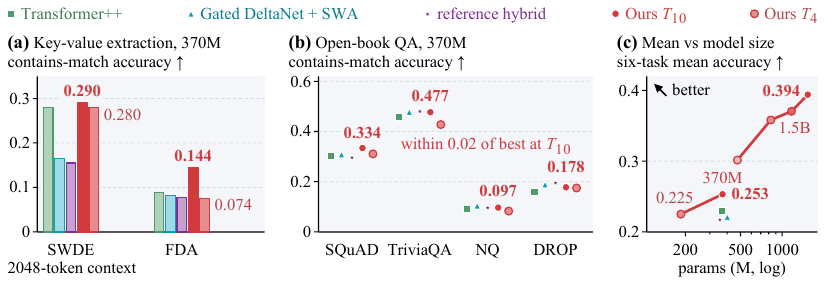}
\caption{Retrieval and question answering from a supplied 2048-token context. (a) SWDE and FDA test extraction of values. (b) four open-book QA tasks test answers recovered from the context. Scores use contains-match accuracy, with higher values better. Generation is greedy, stops at a newline and is capped at 48 tokens; inputs are left-truncated to at most 2000 tokens within the 2048-token evaluation budget. Panel (c) plots the mean of these six tasks against parameter count for the 370M and 1.5B ESSH families, so quality should be compared at similar sizes; the nearly coincident GDN + SWA and Mamba-3 + SWA points are offset slightly along the x-axis. The reference hybrid denotes Mamba-3 + SWA. Table~\ref{tab:ret-based} gives the complete 1.5B and 370M results.}
\label{fig:app-ret-based}
\end{figure}

At 370M, ESSH has the highest scores on SWDE, FDA and SQuAD and the highest six-task mean, 0.253 (Table~\ref{tab:ret-based}; Figure~\ref{fig:app-ret-based}). Its half-size $T_4$ retains a mean of 0.225, above the full Mamba-3 + SWA reference at 0.217 and Gated DeltaNet + SWA at 0.221. Thus the compact export preserves a useful advantage on these supplied-context tasks despite its reduced parameter count.

At 1.5B, the mean rises from 0.302 at $T_1$ to 0.394 at full capacity. SWDE, TriviaQA and NQ increase with capacity, while SQuAD and DROP reach their highest reported scores at $T_4$ (0.453 and 0.259). The intermediate tiers therefore preserve substantial extraction and question-answering performance without requiring every task to improve monotonically with model size.

\begin{table}[htbp]
\centering
\small
\setlength{\tabcolsep}{4pt}
\caption{Contains-match accuracy on six open-book tasks at context 2048. Locally trained 370M rows share data and pretraining tokens. The 1.5B group uses ESSH. Bold compares full-size 370M models.}
\label{tab:ret-based}
\label{tab:based-stage-t1}
\begin{tabular}{@{}lrccccccc@{}}
\toprule
 &  & \multicolumn{7}{c}{contains-match accuracy $\uparrow$} \\
\cmidrule(lr){3-9}
Model & Params (M) & SWDE & FDA & SQuAD & TriviaQA & NQ & DROP & mean \\
\midrule
\multicolumn{9}{l}{1.5B ESSH}\\
\ours{}, $T_{10}$ & 1531.1 & 0.491 & 0.358 & 0.446 & 0.629 & 0.189 & 0.251 & 0.394 \\
\ours{}, $T_{7}$ & 1168.3 & 0.452 & 0.282 & 0.446 & 0.618 & 0.173 & 0.251 & 0.371 \\
\ours{}, $T_{4}$ & 827.6 & 0.411 & 0.297 & 0.453 & 0.578 & 0.151 & 0.259 & 0.358 \\
\ours{}, $T_{1}$ & 474.5 & 0.365 & 0.216 & 0.394 & 0.515 & 0.111 & 0.208 & 0.302 \\
\midrule
\multicolumn{9}{l}{370M models}\\
\ours{}, $T_{10}$ & 370.9 & \textbf{0.290} & \textbf{0.144} & \textbf{0.334} & 0.477 & 0.097 & 0.178 & \textbf{0.253} \\
\ours{}, $T_4$ & 185.3 & 0.280 & 0.074 & 0.311 & 0.428 & 0.083 & 0.175 & 0.225 \\ 
Mamba-3 + SWA\refhyb & 377.1 & 0.155 & 0.077 & 0.297 & \textbf{0.482} & 0.098 & \textbf{0.196} & 0.217 \\ 
Gated DeltaNet + SWA & 378.1 & 0.166 & 0.083 & 0.308 & 0.477 & \textbf{0.104} & 0.189 & 0.221 \\ 
Transformer++ & 365.6 & 0.279 & 0.088 & 0.303 & 0.459 & 0.091 & 0.159 & 0.230 \\ 
\bottomrule
\end{tabular}
\end{table}

\section[Synthetic Recall Suite]{Synthetic Recall Suite}
\label{app:synthetic}

The synthetic suite isolates recall-intensive behaviors in small task-specific models and tests whether they remain available under the same spectral-channel and feed-forward truncation used by the language models. Each backbone applies the public capacity map to its own dimensions and uses the stated channel value width $P$. In the two-layer synthetic hybrid, the final block uses attention with a window covering the full input, shared across tiers.

\paragraph{Multi-query associative recall.}
MQAR uses two-layer, width-128 backbones with feed-forward width 512~\citep{arora2023zoology}. ESSH uses channel value width $P=24$. For ESSH, the learning rate for each sequence-length and pair-count setting is selected before the matched rerun and then held fixed while evaluating $T_1$, $T_4$, and $T_{10}$ from the same 20k-step checkpoint. Baseline rows report the best result from their learning-rate sweeps.

Across three seeds, the hybrid retains strong recall at 64 pairs: mean scaled accuracy remains between 0.971 and 1.000 across the three lengths and all reported tiers (Table~\ref{tab:syn-mqar-seeds}). At $L=1024$, reducing the model from $T_{10}$ to $T_1$ changes the mean from 0.981 to 0.971. The shared attention path remains present at every tier, so substantial channel and FFN reduction preserves this recall behavior in the tested hybrid. Mixer-only models remain near zero in these settings despite having more full-capacity parameters. \revr{The width/budget study in Appendix~\ref{app:rev-mqar-width} shows that the low Transformer scores depend strongly on the tested configuration.}

\revr{At fixed sequence length, increasing pair count also increases the number of supervised queries and changes their spacing, so these settings do not form a strict difficulty ordering.} The repeated runs also distinguish capacity effects from training variability. With 32 pairs at $L=512$, the three ESSH $T_{10}$ scores span 0.019--0.994, and the compact tiers track the full model within each run. Mamba-3 likewise varies substantially across seeds in several 32-pair settings. Thus the stable 64-pair results support preservation of the hybrid's recall under truncation, while the 32-pair averages show that changing capacity within one trained model has a different effect from changing its training seed. Table~\ref{tab:syn-mqar} reports the seed-0 settings for the displayed models. Table~\ref{tab:syn-mqar-seeds} reports the matched repeats.

\begin{table}[htbp]
\centering
\footnotesize
\setlength{\tabcolsep}{4pt}
\caption{MQAR scaled accuracy after 20k training steps, seed 0. Three-seed results for 32 and 64 pairs are in Table~\ref{tab:syn-mqar-seeds}. For each length/pair setting, ESSH uses one matched run with a preselected learning rate and evaluates $T_1$, $T_4$, and $T_{10}$ from the same checkpoint. Baseline rows report their best learning-rate sweep result. Columns vary sequence length and key--value pair count; parameter counts use the task-specific vocabulary.}
\label{tab:syn-mqar}
\label{tab:new-mqar-check}
\label{tab:exp-mqar}

\begin{tabular}{@{}p{.28\linewidth}p{.14\linewidth}rcccc@{}}
\toprule
\multicolumn{7}{l}{Sequence length $L=256$, scaled accuracy $\uparrow$}\\
Model & Capacity & Params (M) & 8 pairs & 16 pairs & 32 pairs & 64 pairs\\
\midrule
\ours{} & $T_{10}$ & 1.81 & 0.999 & 0.998 & 0.999 & 1.000 \\
\ours{} & $T_{4}$ & 1.43 & 0.998 & 0.998 & 0.998 & 1.000 \\
\ours{} & $T_{1}$ & 1.24 & 0.998 & 0.999 & 0.999 & 1.000 \\
\midrule
\ours{} (mixer only) & $T_{10}$ & 2.04 & 0.124 & 0.063 & 0.024 & 0.004 \\
\ours{} (mixer only) & $T_{4}$ & 1.53 & 0.128 & 0.062 & 0.023 & 0.003 \\
\ours{} (mixer only) & $T_{1}$ & 1.25 & 0.077 & 0.003 & 0.011 & 0.003 \\
\midrule
Transformer & -- & 1.57 & 1.000 & 1.000 & 0.022 & 0.008 \\
Mamba-3 & -- & 1.72 & 1.000 & 1.000 & 0.923 & 0.010 \\
\bottomrule
\end{tabular}

\par\smallskip

\begin{tabular}{@{}p{.28\linewidth}p{.14\linewidth}rcccc@{}}
\toprule
\multicolumn{7}{l}{Sequence length $L=512$, scaled accuracy $\uparrow$}\\
Model & Capacity & Params (M) & 8 pairs & 16 pairs & 32 pairs & 64 pairs\\
\midrule
\ours{} & $T_{10}$ & 1.81 & 0.998 & 0.874 & 0.990 & 0.992 \\
\ours{} & $T_{4}$ & 1.43 & 0.994 & 0.631 & 0.986 & 0.992 \\
\ours{} & $T_{1}$ & 1.24 & 0.997 & 0.650 & 0.978 & 0.989 \\
\midrule
\ours{} (mixer only) & $T_{10}$ & 2.04 & 0.125 & 0.059 & 0.003 & 0.000 \\
\ours{} (mixer only) & $T_{4}$ & 1.53 & 0.106 & 0.055 & 0.003 & 0.000 \\
\ours{} (mixer only) & $T_{1}$ & 1.25 & 0.053 & 0.017 & 0.002 & 0.000 \\
\midrule
Transformer & -- & 1.57 & 1.000 & 1.000 & 0.025 & 0.006 \\
Mamba-3 & -- & 1.72 & 0.991 & 1.000 & 0.825 & 0.008 \\
\bottomrule
\end{tabular}

\par\smallskip

\begin{tabular}{@{}p{.28\linewidth}p{.14\linewidth}rcccc@{}}
\toprule
\multicolumn{7}{l}{Sequence length $L=1024$, scaled accuracy $\uparrow$}\\
Model & Capacity & Params (M) & 8 pairs & 16 pairs & 32 pairs & 64 pairs\\
\midrule
\ours{} & $T_{10}$ & 1.81 & 0.995 & 0.992 & 0.953 & 0.978 \\
\ours{} & $T_{4}$ & 1.43 & 0.997 & 0.990 & 0.923 & 0.979 \\
\ours{} & $T_{1}$ & 1.24 & 0.993 & 0.983 & 0.906 & 0.977 \\
\midrule
\ours{} (mixer only) & $T_{10}$ & 2.04 & 0.051 & 0.047 & 0.002 & 0.000 \\
\ours{} (mixer only) & $T_{4}$ & 1.53 & 0.015 & 0.014 & 0.002 & 0.000 \\
\ours{} (mixer only) & $T_{1}$ & 1.25 & 0.016 & 0.001 & 0.000 & 0.000 \\
\midrule
Transformer & -- & 1.57 & 1.000 & 1.000 & 0.027 & 0.013 \\
Mamba-3 & -- & 1.72 & 1.000 & 1.000 & 0.024 & 0.119 \\
\bottomrule
\end{tabular}
\end{table}

\begin{table}[htbp]
\centering\footnotesize
\setlength{\tabcolsep}{4pt}
\caption{MQAR across training seeds 0, 1, and 2: mean [minimum, maximum] scaled accuracy after 20k steps. The learning rate is fixed across seeds within each model/length/pair setting. ESSH tiers share each seed's checkpoint.}
\label{tab:syn-mqar-seeds}
\begin{tabular*}{\linewidth}{@{\extracolsep{\fill}}lccc@{}}
\toprule
Model / tier & $L=256$ & $L=512$ & $L=1024$\\
\midrule
\multicolumn{4}{l}{32 key--value pairs}\\
ESSH $T_{10}$ & 0.937 [0.818, 0.999] & 0.668 [0.019, 0.994] & 0.917 [0.811, 0.988] \\
ESSH $T_4$ & 0.846 [0.549, 0.998] & 0.664 [0.014, 0.993] & 0.854 [0.659, 0.980] \\
ESSH $T_1$ & 0.792 [0.632, 0.999] & 0.661 [0.012, 0.991] & 0.867 [0.728, 0.966] \\
Mixer only $T_{10}$ & 0.024 [0.022, 0.025] & 0.005 [0.002, 0.009] & 0.005 [0.002, 0.010] \\
Mixer only $T_4$ & 0.022 [0.017, 0.024] & 0.002 [0.002, 0.003] & 0.003 [0.002, 0.004] \\
Mixer only $T_1$ & 0.010 [0.009, 0.011] & 0.002 [0.002, 0.002] & 0.002 [0.000, 0.004] \\
Transformer & 0.021 [0.021, 0.022] & 0.025 [0.025, 0.025] & 0.026 [0.025, 0.027] \\
Mamba-3 & 0.323 [0.022, 0.923] & 0.290 [0.022, 0.825] & 0.349 [0.024, 0.997] \\
\midrule
\multicolumn{4}{l}{64 key--value pairs}\\
ESSH $T_{10}$ & 1.000 [1.000, 1.000] & 0.991 [0.989, 0.992] & 0.981 [0.978, 0.989] \\
ESSH $T_4$ & 1.000 [1.000, 1.000] & 0.990 [0.987, 0.992] & 0.981 [0.974, 0.989] \\
ESSH $T_1$ & 1.000 [1.000, 1.000] & 0.985 [0.974, 0.992] & 0.971 [0.967, 0.977] \\
Mixer only $T_{10}$ & 0.008 [0.004, 0.011] & 0.002 [0.000, 0.005] & 0.000 [0.000, 0.001] \\
Mixer only $T_4$ & 0.006 [0.003, 0.008] & 0.002 [0.000, 0.004] & 0.000 [0.000, 0.001] \\
Mixer only $T_1$ & 0.004 [0.003, 0.005] & 0.001 [0.000, 0.002] & 0.000 [0.000, 0.001] \\
Transformer & 0.008 [0.007, 0.008] & 0.007 [0.006, 0.008] & 0.013 [0.013, 0.013] \\
Mamba-3 & 0.009 [0.009, 0.010] & 0.009 [0.008, 0.010] & 0.364 [0.004, 0.969] \\
\bottomrule
\end{tabular*}
\end{table}

\revsubsection{Width check for 32-pair MQAR at $L=256$}{app:rev-mqar-width}
\begin{revblockr}
On the 32-pair MQAR test at $L=256$, the width-128 Transformer scores 0.022 under the default budget (batch 32, 20k steps) and 0.053 with ten times the training tokens (batch 256, 25k steps), across a learning-rate sweep. At width 256, it reaches 0.998 under the default budget (Table~\ref{tab:rev-mqar-width}), while ESSH reaches 0.999 at width 128. Increasing the budget alone does not resolve the low score, whereas the wider configuration succeeds. The low baseline score is therefore specific to the tested narrow configuration.
\end{revblockr}

\begin{table}[htbp]
\begin{revblockr}
\centering\footnotesize
\setlength{\tabcolsep}{4pt}
\revcaption{MQAR scaled accuracy with 32 key--value pairs at $L=256$ at two widths $d$ and two training budgets (default budget: batch 32 for 20k steps, $10\times$: batch 256 for 25k steps). The Transformer is evaluated at multiple learning rates. The $d=128$ default-budget cells are the seed-0 values of Table~\ref{tab:syn-mqar} ($T_{10}$ for ESSH). Dashes: not run.}{tab:rev-mqar-width}
\begin{tabular}{@{}lcccc@{}}
\toprule
& \multicolumn{2}{c}{$d=128$} & \multicolumn{2}{c}{$d=256$} \\
\cmidrule(lr){2-3}\cmidrule(lr){4-5}
Model & default budget & $10\times$ budget & default budget & $10\times$ budget \\
\midrule
Transformer & 0.022 & 0.053 & 0.998 & 1.000 \\
Mamba-3 & 0.923 & 0.970 & -- & 1.000 \\
ESSH & 0.999 & 1.000 & -- & 1.000 \\
\bottomrule
\end{tabular}\par
\end{revblockr}
\end{table}

\subsection[Palindrome completion]{Palindrome completion}
\label{app:syn-palindrome}

Palindrome completion extends associative retrieval from individual key--value pairs to reproducing an entire input sequence in reverse order. At $L=1024$, the full-capacity hybrid scores $1.000$ and the $T_4$ export scores $0.997$. The full-capacity mixer-only model has more parameters, $0.996$M versus $0.765$M for the hybrid, but scores only $0.120$ (Table~\ref{tab:palindrome-new}). Its accuracy also falls sharply as sequence length increases, whereas the hybrid remains near perfect at $T_{10}$ and $T_4$.

For each sequence-length setting, the attention window covers the full input and remains fixed across tiers, so the hybrid retains a direct attention path from each copying position to the source sequence. The contrast with the mixer-only model is consistent with this path being important for sequence copying. Because attention is shared across ESSH tiers, channel and FFN truncation reduce recurrent and tokenwise capacity without removing this direct source-access path. Even at $T_1$, which retains only $0.198$M parameters, palindrome accuracy remains $0.974$ at $L=1024$.

\begin{table}[htbp]
\centering
\small
\caption{Palindrome token accuracy after 20k training steps, seed 0. Tier rows use the retained standalone model shape. Depth, residual width, and training steps are shared across the compared task-specific models; parameter counts differ.}
\label{tab:palindrome-new}
\begin{tabular}{@{}llrrrr@{}}
\toprule
Model & Capacity & Params (M) & $L=256$ & $L=512$ & $L=1024$ \\
\midrule
Hybrid & $T_{10}$ & 0.765 & 1.000 & 1.000 & 1.000 \\
Hybrid & $T_{4}$ & 0.389 & 1.000 & 1.000 & 0.997 \\
Hybrid & $T_{1}$ & 0.198 & 0.999 & 0.996 & 0.974 \\
\midrule
Mixer only & $T_{10}$ & 0.996 & 0.892 & 0.428 & 0.120 \\
Mixer only & $T_{4}$ & 0.491 & 0.088 & 0.111 & 0.040 \\
Mixer only & $T_{1}$ & 0.207 & 0.147 & 0.105 & 0.044 \\
\bottomrule
\end{tabular}
\end{table}

\section[Cross-Domain Language Modeling]{Cross-Domain Language Modeling}
\label{app:cross_domain}

Table~\ref{tab:domain} evaluates transfer to six text domains using perplexity within each tokenizer group. All 370M models share the same pretraining corpus and budget. ESSH also receives brief shared refinement. The 1.5B block uses the final ESSH model.

ESSH has lower PPL than Transformer++ on all six 370M domain slices and the lowest PPL on arXiv and Proof-Pile (Table~\ref{tab:domain}). The 274.3M $T_7$ also remains below the 377.1M Mamba-3 + SWA reference on arXiv, GitHub and Proof-Pile. Gated DeltaNet + SWA leads on PG-19, GitHub and WikiText, while Mamba-3 + SWA leads on CC-News, so the domain-level results complement the main-validation ranking.

Capacity sensitivity varies by domain. At 1.5B, reducing to $T_1$ raises GitHub PPL from 43.65 to 97.24, while CC-News changes from 13.89 to 22.79. The intermediate $T_7$ stays closer to full capacity at 50.63 and 14.93, respectively. These differences make the evaluated intermediate tiers useful for selecting a size under domain-specific quality requirements.

\begin{table}[!htbp]
\centering\footnotesize
\setlength{\tabcolsep}{3pt}
\caption{Cross-domain perplexity ($\downarrow$), compared within each tokenizer group. The 1.5B block uses ESSH; the 370M group uses the same training stages as the main comparison. Each group scores fixed non-overlapping 2048-target-token windows, up to 400 per domain; WikiText supplies 139 windows for GPT-2 and 141 for Llama-3.1. Bold compares full-size models in the 370M group.}
\label{tab:domain}
\begin{tabular*}{\linewidth}{@{\extracolsep{\fill}}lrrrrrrr@{}}
\toprule
Model & Params (M) & PG-19 & arXiv & GitHub & CC-News & Proof-Pile & WikiText \\
\midrule
\multicolumn{8}{l}{\textit{1.5B ESSH; Llama-3.1 tokenizer}}\\
\ours{}, $T_{10}$ & 1531.1 & 27.59 & 28.35 & 43.65 & 13.89 & 36.00 & 13.57 \\
\ours{}, $T_{7}$ & 1168.3 & 30.11 & 31.16 & 50.63 & 14.93 & 39.21 & 14.88 \\
\ours{}, $T_{4}$ & 827.6 & 33.30 & 34.36 & 58.86 & 16.70 & 43.92 & 16.71 \\
\ours{}, $T_{1}$ & 474.5 & 48.40 & 48.61 & 97.24 & 22.79 & 63.94 & 23.35 \\
\midrule
\multicolumn{8}{l}{\textit{370M models; GPT-2 tokenizer}}\\
\ours{}, $T_{10}$ & 370.9 & 31.34 & \textbf{23.75} & 14.28 & 20.62 & \textbf{28.91} & 22.01 \\
\ours{}, $T_{7}$ & 274.3 & 34.26 & 25.16 & 15.52 & 22.20 & 30.33 & 23.90 \\
\ours{}, $T_{4}$ & 185.3 & 37.22 & 27.25 & 16.73 & 24.62 & 33.04 & 26.72 \\
\ours{}, $T_{1}$ & 92.1 & 51.30 & 37.20 & 24.74 & 33.06 & 45.79 & 36.83 \\
Mamba-3 + SWA & 377.1 & 30.57 & 27.87 & 16.20 & \textbf{19.89} & 33.38 & 21.46 \\
Gated DeltaNet + SWA & 378.1 & \textbf{30.12} & 26.31 & \textbf{14.04} & 20.12 & 32.61 & \textbf{20.93} \\
Transformer++ & 365.6 & 34.50 & 29.42 & 16.44 & 22.07 & 34.97 & 23.99 \\
\bottomrule
\end{tabular*}
\end{table}

\FloatBarrier
\section[DNA Sequence Modeling]{DNA Sequence Modeling}
\label{app:dna}

The DNA study tests whether the same train-once, export-many capacity mechanism transfers beyond language modeling. We pretrain a four-layer, 0.534M-parameter ESSH backbone on GRCh38 for 3B nucleotide tokens, holding out the chr21 primary contig~\citep{schneider2017grch38}. The DNA backbone contains spectral blocks without attention and uses channel value width $P=8$, which determines its tier-specific parameter and recurrent-state counts.

Published DNA models at similar parameter scales are reported as external references in the downstream experiments. Their training recipes and directionality differ from ours, so controlled comparisons use the same ESSH backbone across capacities and pretraining conditions.

\subsection[Held-out next-nucleotide loss]{Held-out next-nucleotide loss}
\label{app:dna-loss}

Chromosome 21 evaluates the pretrained exports before task-specific adaptation. Across the four measured tiers, next-nucleotide loss ranges from $1.775$ bits per base at $T_1$ to $1.662$ at full capacity (Table~\ref{tab:dna-loss}), but the loss--capacity curve is not uniform.

The middle two tiers are nearly identical on this metric: $T_4$ reaches $1.732$ bits per base and $T_7$ reaches $1.729$, a difference of only $0.003$, although $T_4$ uses $34\%$ fewer parameters. By contrast, increasing from $T_7$ to $T_{10}$ reduces loss by $0.067$ bits per base. 

The next subsection evaluates the same pretrained exports after task-specific adaptation to genomic classification benchmarks.

\begin{table}[htbp]
\centering
\small
\setlength{\tabcolsep}{3pt}
\caption{Held-out chromosome-21 next-nucleotide loss. Every tier is evaluated on the same 4000 non-overlapping windows of 1024 bases. Parameter counts exclude task-specific classification heads.}
\label{tab:dna-loss}
\begin{tabular*}{\linewidth}{@{\extracolsep{\fill}}lrrrr@{}}
\toprule
& $T_1$ & $T_4$ & $T_7$ & $T_{10}$ \\
\midrule
Params (M) & 0.135 & 0.240 & 0.362 & 0.534 \\
Bits per base $\downarrow$ & 1.775 & 1.732 & 1.729 & 1.662 \\
\bottomrule
\end{tabular*}
\end{table}

\subsection[Genomic Benchmarks across capacities]{Genomic Benchmarks across capacities}
\label{app:dna-small}
\label{app:dna-gb}

Each export is adapted for ten epochs on eight of the nine Genomic Benchmarks tasks, with test accuracy taken at the best validation epoch and averaged over seeds 0, 197, and 511~\citep{gresova2023genomicbenchmarks}. All four exports are evaluated under this protocol.

The clearest control is the full architecture trained without pretraining. The pretrained full model improves on it in all eight tasks, and even the 0.135M $T_1$ exceeds this 0.534M control in every task. Shared pretraining therefore supplies useful sequence representations that survive substantial capacity reduction. This improvement remains available when roughly three quarters of the full backbone's parameters are removed.

Table~\ref{tab:dna-gb} shows task-specific responses to capacity reduction. $T_4$ and $T_7$ exceed the full model on mouse enhancers and regulatory sequences, whereas non-TATA promoter accuracy increases from 0.887 at $T_1$ to 0.942 at $T_{10}$. Capacity selection therefore depends on the downstream task: some tasks retain their strongest measured score in a compact export, while the promoter task benefits consistently from additional capacity. These outcomes complement the held-out next-nucleotide loss, which favors the full model overall. Among the published references, ESSH has the highest coding-versus-intergenomic score (0.923), and every tier exceeds the reported reference scores on human regulatory sequences. The 0.135M $T_1$ exceeds the 0.264M CNN on all eight tasks. Caduceus leads the other six tasks using a bidirectional, reverse-complement-aware architecture. Published values are taken from the comparison reported by \citet{schiff2024caduceus}, including its HyenaDNA and Mamba runs. The CNN originates from \citet{gresova2023genomicbenchmarks}.

\begin{table}[htbp]
\centering\scriptsize
\setlength{\tabcolsep}{3pt}
\caption{Genomic Benchmarks test accuracy ($\uparrow$). ESSH rows average three task-adaptation seeds (0, 197, 511); each export is adapted separately. No pretraining uses the full architecture. $\dagger$: results reported by \citet{schiff2024caduceus}, with the CNN from \citet{gresova2023genomicbenchmarks}; bidir.: bidirectional. Published recipes differ from ours. OCR denotes open chromatin regions.}
\label{tab:dna-gb}
\label{tab:new-dna-small}
\begin{tabular*}{\linewidth}{@{\extracolsep{\fill}}lrrrrrrrrr@{}}
\toprule
Model / tier & \shortstack{Params\\(M)} & \shortstack{Mouse\\enh.} & Coding & \shortstack{Human/\\worm} & Cohn & \shortstack{Ensembl\\enh.} & Regul. & OCR & \shortstack{Non-\\TATA} \\
\midrule
$T_{1}$ & 0.135 & 0.759 & 0.905 & 0.962 & 0.729 & 0.834 & 0.879 & 0.765 & 0.887 \\
$T_{4}$ & 0.240 & 0.774 & 0.906 & 0.960 & 0.721 & 0.836 & 0.885 & 0.761 & 0.913 \\
$T_{7}$ & 0.362 & 0.775 & 0.906 & 0.960 & 0.725 & 0.846 & 0.884 & 0.766 & 0.925 \\
$T_{10}$ & 0.534 & 0.760 & 0.923 & 0.961 & 0.726 & 0.869 & 0.879 & 0.773 & 0.942 \\
No pretraining & 0.534 & 0.716 & 0.871 & 0.949 & 0.682 & 0.773 & 0.745 & 0.683 & 0.808 \\
\midrule
CNN\refpub\ & 0.264 & 0.715 & 0.892 & 0.942 & 0.702 & 0.744 & 0.872 & 0.698 & 0.861 \\
HyenaDNA\refpub\ & 0.436 & 0.780 & 0.904 & 0.964 & 0.729 & 0.849 & 0.869 & 0.783 & 0.944 \\
Mamba\refpub\ & 0.468 & 0.743 & 0.904 & 0.967 & 0.732 & 0.862 & 0.814 & 0.815 & 0.933 \\
Caduceus-Ph\refpub\ (bidir.) & 0.470 & 0.754 & 0.915 & 0.973 & 0.747 & 0.893 & 0.872 & 0.828 & 0.946 \\
Caduceus-PS\refpub\ (bidir.) & 0.470 & 0.793 & 0.910 & 0.968 & 0.745 & 0.900 & 0.873 & 0.818 & 0.945 \\
\bottomrule
\end{tabular*}
\end{table}

\FloatBarrier
\section[Selective-Dynamics Ablations]{Selective-Dynamics Ablations}
\label{app:selective_ablations}
\label{app:ablation-summary}

This section examines the selective recurrence, its initialization, and the allocation of recurrent state under the 370M training recipe. Clock and gate controls test input-dependent state dynamics. The remaining experiments vary the initial temporal representation, mode count, or channel partition. Capacity-map and distillation controls are reported separately in Appendix~\ref{app:elastic_ablations}.

\subsection[Clock, gates, and initialization]{Clock, gates, and initialization}
\label{app:sel-mech}

The control is the 370M ESSH recipe trained for 3.7B tokens with seed 197. The input-independent-clock arm fixes the clock projection weights at zero while retaining trainable clock biases. The neutral-gate arm fixes both write/read gates at one. State dimensions, the capacity map, and the attention layout are unchanged. Full-capacity PPL uses the end-of-pretraining model for comparison with full-capacity-only training. Reduced-capacity PPL and retrieval use the final models. Comparisons between arms use the same stage within each column.

At full capacity, fixing the clock increases PPL by 0.34 and neutralizing the gates increases it by 0.63, compared with the 0.09 range across three control seeds (Table~\ref{tab:new-init-seeds}). Both interventions also increase PPL at the reduced capacities. The corresponding $T_4$ first-token accuracies fall from 0.259 to 0.139 and 0.102. Thus language modeling and retrieval both favor the input-dependent controls in these runs, with retrieval reported as a single-seed measurement.

\begin{table}[htbp]
\centering
\footnotesize
\setlength{\tabcolsep}{3.3pt}
\caption{Selective-dynamics controls at 370M after 3.7B pretraining tokens. Compact PPL and needle first-token accuracy use the final checkpoints; full-capacity PPL uses the corresponding raw checkpoint. Increasing $m$ also increases parameter count and recurrent-state size. \revr{Stage: $T_1$--$T_7$ PPL and needle accuracy after shared refinement, and $T_{10}$ PPL from the raw end-of-pretraining checkpoint. Evaluator: the common 16-batch PPL evaluator of Appendix~\ref{app:setup-eval}. Raw-checkpoint values of the control, random-mode and shuffled-slot arms are in Table~\ref{tab:new-init-seeds}.}}
\label{tab:sel-mech}
\label{tab:init-r}
\begin{tabular}{@{}lcccccccc@{}}
\toprule
& \multicolumn{4}{c}{validation PPL $\downarrow$}
& \multicolumn{2}{c}{needle first-token acc. $\uparrow$}
& 7-task $\uparrow$ & params \\
\cmidrule(lr){2-5}\cmidrule(lr){6-7}
Arm & $T_1$ & $T_4$ & $T_7$ & $T_{10}$ & $T_4$ & $T_{10}$ & $T_{10}$ & $T_{10}$ (M) \\
\midrule
\ours{} (control) & 25.32 & 19.15 & 17.38 & 16.46 & 0.259 & 0.336 & 0.408 & 370.9 \\
\ours{} (no selective clock) & 25.69 & 19.47 & 17.70 & 16.80 & 0.139 & 0.266 & 0.410 & 370.9 \\
\ours{} (neutral gates) & 26.04 & 19.79 & 17.99 & 17.09 & 0.102 & 0.140 & 0.406 & 370.9 \\
\ours{} (non-oscillatory init., no fit) & 26.18 & 19.92 & 18.01 & 17.00 & 0.140 & 0.180 & 0.405 & 370.9 \\
\ours{} (random damped modes) & 26.00 & 19.64 & 17.68 & 16.69 & 0.234 & 0.280 & 0.406 & 370.9 \\
\ours{} (shuffled slots) & 25.66 & 19.31 & 17.51 & 16.51 & 0.115 & 0.202 & 0.414 & 370.9 \\
\ours{} ($m=16$) & 26.17 & 19.39 & 17.51 & 16.61 & 0.263 & 0.347 & 0.409 & 384.0 \\
\bottomrule
\end{tabular}
\end{table}

\begin{figure}[htbp]
\centering
\includegraphics[width=\linewidth]{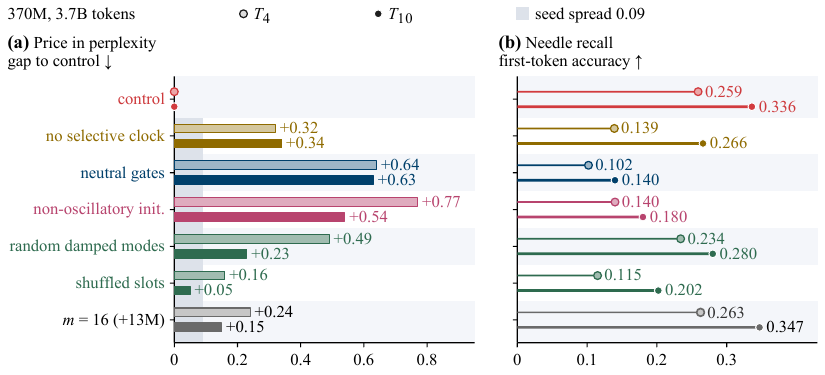}
\caption{Effects of selective-mixer controls on language modeling and retrieval. (a) PPL changes relative to the control; positive values are worse, and shading shows the three-seed range of the control's full-capacity PPL (0.09; Table~\ref{tab:new-init-seeds}). (b) Needle first-token accuracy at full capacity and $T_4$. The $m=16$ arm also increases parameter count and recurrent-state size. Table~\ref{tab:sel-mech} gives the numerical values and checkpoint stages.}
\label{fig:app-sel-mech}
\end{figure}

\subsection[Real poles against rotating modes]{Real poles against rotating modes}
\label{app:init-real}

We compare damped rotations with positive real poles when approximating the same Hankel filters. A rotation uses two real state coordinates, so eight rotations and sixteen real poles have equal state size (Table~\ref{tab:init-real}).

Eight rotations are more accurate on 14 of the 24 numerically resolved channels, including all eight channels retained by $T_1$. They fit eight channels within 5\% error, versus six for sixteen real poles. Here, damped modes include zero-frequency terms. At channel eight the relative errors are 0.036 and 0.058, respectively. Sixteen real poles perform better on channels 9--17 and 20 and have the lower all-channel median, so the rotation fit's strongest advantage is on the low-index channels shared by every deployment tier.

The trained initialization controls provide complementary evidence. Non-oscillatory initialization raises PPL from 25.32 to 26.18 at $T_1$ and from 16.46 to 17.00 at full capacity, while random damped rotations reach 26.00 and 16.69 (Table~\ref{tab:sel-mech}). The non-oscillatory arm starts with identical decays $\rho=\mathrm e^{-0.01}$, zero frequencies and zero readout coefficients, giving zero initial mixer output. The higher PPL under this control supports the fitted temporal initialization within the capacity-mixed training recipe.

\begin{table}[htbp]
\centering
\small
\setlength{\tabcolsep}{5pt}
\caption{Relative filter-fit error at $L=2048$ for positive real decays and damped rotations. Medians use the first 24 numerically resolved channels. Each damped mode uses two real state coordinates. Errors refer to the filter-fitting comparison on float64 Hankel eigenvectors.}
\label{tab:init-real}
\resizebox{\linewidth}{!}{%
\begin{tabular}{@{}lcccccc@{}}
\toprule
& \multicolumn{6}{c}{relative 2-norm error $\downarrow$} \\
\cmidrule(lr){2-7}
Fit & median ($k\le24$) & $k=1$ & $k=4$ & $k=8$ & $k=16$ & $k=24$ \\
\midrule
rotations, $m=8$ (initialization) & 0.59 & $1.4\times10^{-6}$ & $2.2\times10^{-4}$ & 0.04 & 0.75 & 0.81 \\
rotations, $m=16$ & $8.9\times10^{-6}$ & $2.4\times10^{-10}$ & $7.8\times10^{-8}$ & $1.8\times10^{-6}$ & $8.8\times10^{-7}$ & 0.56 \\
real poles, $m=8$ & 0.76 & $9.2\times10^{-3}$ & 0.06 & 0.70 & 0.95 & 1.00 \\
real poles, $m=16$ & 0.30 & $4.2\times10^{-4}$ & $4.1\times10^{-3}$ & 0.06 & 0.54 & 0.95 \\
real poles, $m=32$ & $5.6\times10^{-4}$ & $3.9\times10^{-7}$ & $1.2\times10^{-5}$ & $2.6\times10^{-4}$ & $8.1\times10^{-3}$ & 0.11 \\
\bottomrule
\end{tabular}}
\end{table}

\begin{figure}[htbp]
\centering
\includegraphics[width=\linewidth]{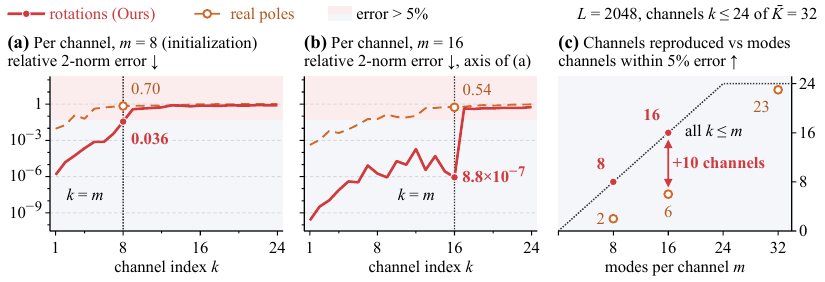}
\caption{Finite-mode approximation of the Hankel spectral filters. Panels (a,b) show relative $\ell_2$ error for eight and 16 modes over the first 24 numerically resolved channels; panel (c) counts channels below the stated error threshold as mode budget increases. A damped rotation uses two real state coordinates, while a positive real pole uses one. Panels (a,b) compare equal mode counts; eight rotations and sixteen real poles have equal state size. Table~\ref{tab:init-real} reports the numerical summaries.}
\label{fig:app-init-real}
\end{figure}

\subsection[Initialization controls and the channel importance profile]{Initialization controls and the channel importance profile}
\label{app:init-order}

The initialization controls vary either the temporal responses or the placement of complete channels. Random damped modes remove the fitted channel structure while preserving the initialization scale, whereas slot shuffling preserves each fitted channel but changes its position in the ordered bank. Uniform-decay and S4D-Lin~\citep{gu2022s4d} controls provide alternative starting timescales.

For the two seeds shared by the initialization controls, random damped modes give higher full and compact PPL than the Hankel initialization. Slot shuffling produces smaller PPL changes. The two S4D-Lin seeds give full-model PPL of 16.34--16.38, slightly below the Hankel control's 16.46--16.55 range. At reduced capacities, their $T_4$ PPL is approximately 19.61 and their $T_1$ range of 26.13--26.23 overlaps the Hankel control's 26.18--26.28 range. Thus both structured initializations support useful compact models under the same selective recurrence and capacity-mixed training procedure. Retrieval varies more strongly across seeds and is reported without ranking. \revr{Appendix~\ref{app:rev-init-s4d} compares final PPL, early optimization and truncation after dense pretraining for the two initializations.}

\begin{table}[!htbp]
\centering
\small
\setlength{\tabcolsep}{3.5pt}
\caption{\revr{Initialization controls after 3.7B pretraining tokens, before shared refinement, using the common 16-batch PPL evaluator of Appendix~\ref{app:setup-eval}. Rows report independent training seeds.} Full-key accuracy at $T_4$ is teacher forced.}
\label{tab:new-init-seeds}
\begin{tabular}{@{}lrrrrr@{}}
\toprule
Initialization & Seed & PPL $T_{10}$ & PPL $T_4$ & PPL $T_1$ & Full-key acc. $T_4$ \\
\midrule
Hankel control & 197 & 16.46 & 19.59 & 26.19 & 0.249 \\
Hankel control & 313 & 16.55 & 19.67 & 26.28 & 0.030 \\
Hankel control & 511 & 16.52 & 19.64 & 26.18 & 0.134 \\
Random damped modes & 197 & 16.68 & 20.11 & 26.91 & 0.253 \\
Random damped modes & 313 & 16.84 & 20.26 & 27.14 & 0.177 \\
Shuffled slots & 197 & 16.51 & 19.73 & 26.49 & 0.129 \\
Shuffled slots & 313 & 16.50 & 19.72 & 26.28 & 0.164 \\
S4D-Lin & 197 & 16.34 & 19.61 & 26.13 & 0.202 \\
S4D-Lin & 313 & 16.38 & 19.61 & 26.23 & 0.205 \\
\bottomrule
\end{tabular}
\end{table}

\begin{table}[htbp]
\centering
\footnotesize
\setlength{\tabcolsep}{3pt}
\caption{Initialization controls at 370M and 3.7B tokens. PPL differences use end-of-pretraining validation from the trainer (seed 197). Table~\ref{tab:sel-mech} reports reduced-capacity PPL after shared refinement, while Table~\ref{tab:new-init-seeds} uses the common PPL evaluator for independent seeds.}
\label{tab:init-controls}
\begin{tabular*}{\linewidth}{@{\extracolsep{\fill}}p{.30\linewidth}p{.43\linewidth}cc@{}}
\toprule
Arm & Change & $\Delta$PPL $T_{10}\downarrow$ & $\Delta$PPL $T_1\downarrow$ \\
\midrule
\ours{} (control) & -- & 0 & 0 \\
\ours{} (non-oscillatory init.) & identical real-decay modes, zero readout, no fit & $+0.54$ & $+0.90$ \\
\ours{} (random damped modes) & fitted channel structure removed & $+0.23$ & $+0.72$ \\
\ours{} (shuffled slots) & fitted channels placed in random slots & $+0.05$ & $+0.30$ \\
\ours{} (reversed slots) & fitted channels placed in reverse order & $+0.07$ & $+0.30$ \\
\ours{} (all decays $1-10^{-4}$) & uniform initial decay across modes & $+0.88$ & $+1.21$ \\
\ours{} (all decays $1-10^{-3}$) & uniform initial decay across modes & $+0.89$ & $+1.23$ \\
\ours{} (S4D-Lin init.) & S4D-Lin damped-mode initialization & $-0.11$ & $-0.05$ \\
\bottomrule
\end{tabular*}
\end{table}

Knock-out importance is the validation-PPL increase after zeroing a channel, while output energy is its mean squared activation magnitude. Budget dropout concentrates predictive importance in the retained channels: the first eight account for about 64\% of the knock-out effect in both control seeds, compared with 36\% in the dense control. Their knock-out ordering is 0.75--0.82 versus 0.29 for the dense control. Random-mode and shuffled-slot models also develop strong ordering under budget dropout. Output energy is more seed dependent, with first-eight shares of 0.46 and 0.74 in the two controls, so it is reported separately from predictive importance.

\begin{table}[htbp]
\centering
\small
\setlength{\tabcolsep}{5pt}
\caption{Channel importance at 370M after 3.7B tokens. Rows use seed 197 unless stated. Ordering is $-\operatorname{Spearman}(k,\mathrm{importance}_k)$, so larger values indicate a stronger decrease with channel index. Shares use the first eight channel indices. Output energy and knock-out importance are reported separately. The alternative map uses exponents $(1.0,0.7)$; mixer-only nesting keeps FFN width fixed.}
\label{tab:init-order}
\begin{tabular}{@{}lcccc@{}}
\toprule
& \multicolumn{2}{c}{Negative Spearman $\uparrow$}
& \multicolumn{2}{c}{first-8 share} \\
\cmidrule(lr){2-3}\cmidrule(lr){4-5}
Model & energy & knock-out & energy & knock-out \\
\midrule
initialization (Hankel, untrained) & 0.86 & -- & 0.92 & -- \\
\ours{} (control, seed 197) & 0.66 & 0.82 & 0.46 & 0.64 \\
\ours{} (control, seed 313) & 0.84 & 0.75 & 0.74 & 0.64 \\
\ours{} (alternative map) & 0.74 & 0.70 & 0.51 & 0.57 \\
\ours{} (mixer-only nesting) & 0.80 & 0.89 & 0.57 & 0.77 \\
\ours{} (dense control, nothing nested) & 0.29 & 0.29 & 0.46 & 0.36 \\
\ours{} (non-oscillatory init.) & $-0.05$ & 0.88 & 0.37 & 0.45 \\
\ours{} (random damped modes) & 0.15 & 0.87 & 0.85 & 0.75 \\
\ours{} (shuffled slots) & 0.65 & 0.72 & 0.12 & 0.55 \\
\ours{} (all decays $1-10^{-4}$) & 0.53 & 0.78 & 0.95 & 0.91 \\
\ours{} (all decays $1-10^{-3}$) & 0.51 & 0.78 & 0.78 & 0.86 \\
\bottomrule
\end{tabular}
\end{table}

\begin{figure}[!t]
\centering
\includegraphics[width=\linewidth]{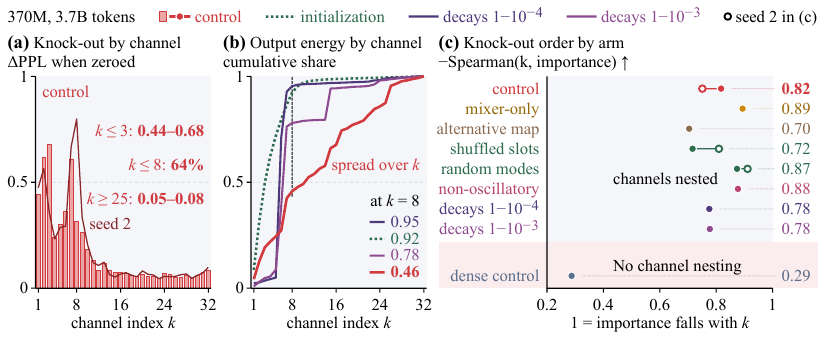}
\caption{Concentration of predictive importance toward low-index spectral channels. (a) Knock-out importance measures the PPL increase after zeroing one channel. (b) Cumulative output energy measures activation magnitude and is distinct from knock-out importance. (c) Larger $-\operatorname{Spearman}(k,\mathrm{importance}_k)$ indicates a stronger decrease of importance with channel index. Open circles in (c) and the line labeled ``seed 2'' in (a) denote seed 313; panel (b) uses seed 197. Table~\ref{tab:init-order} gives the corresponding statistics.}
\label{fig:app-init-order}
\end{figure}

\phantomsection
\label{app:init-drift}
\label{app:sel-drift}
The intrinsic horizon $1/\gamma$ describes a mode's fixed decay, while a clock-adjusted attenuation statistic uses $\rho\,\mathbb E[\exp(-\Delta)]$. These are per-mode summaries. Retrieval measurements in Appendix~\ref{app:retrieval_results} evaluate the complete hybrid, including its recurrent layers and shared attention.

\revsubsection{Hankel and S4D-Lin initializations: early training and dense truncation\revr{, and a joint LTI control}}{app:rev-init-s4d}
\begin{revblockr}
The initialization comparison holds the channel parameterization and capacity-mixed training recipe fixed. Hankel initialization assigns fitted temporal responses to the channel groups in spectral order. The S4D-Lin control instead uses a log-spaced step $\zeta_k\in[10^{-3},10^{-1}]$ for channel $k$, mode frequency $\pi i\zeta_k$ and initial decay $\exp(-\zeta_k/2)$. Its readout phases are random, with amplitude matched to the median fitted-coefficient magnitude. Both initialize the same trainable selective recurrence.

Hankel reaches the early compact-tier quality target with fewer training tokens. At 0.39B tokens, its $T_1$ PPL is 50.4, compared with 54.7 for S4D-Lin and 60.2 for sixteen real poles (Table~\ref{tab:rev-init-early}). It reaches $T_1$ PPL 50 after about 0.40B tokens, compared with 0.44--0.51B for the alternatives (Figure~\ref{fig:rev-init-curves}). This is an early optimization benefit under the standard capacity-mixed recipe.

After 3.7B-token pretraining and the common 2060-step shared refinement, the two initializations reach similar final PPL: 25.32--25.38 at $T_1$ and 16.22--16.32 at $T_{10}$ for Hankel, and 25.29--25.35 and 16.12--16.16 for S4D-Lin (three seeds each; Table~\ref{tab:rev-lti}).

Dense pretraining followed by the same shared refinement tests the initialization choice when reduced capacities are not trained during pretraining (Table~\ref{tab:rev-init-dense}). Across two seeds, Hankel reaches $T_1$ PPL 45.41--45.65 and $T_4$ PPL 23.75--24.06, compared with 63.62--63.91 and 29.40--29.65 for S4D-Lin. Full-capacity PPL remains close (16.05--16.07 versus 15.96--15.98). Thus Hankel also improves compact-model recovery from the densely trained parents. Table~\ref{tab:init-order} separately shows that random-mode and shuffled-slot controls learn useful importance ordering under budget dropout. Initialization supplies the starting temporal responses, while capacity-mixed training optimizes the retained predictors.

The sixteen-real-pole control tests the role of oscillatory modes at matched active state size. It replaces eight damped rotations, with FFN width adjusted to match the full-model parameter count within 0.5\%. After the same training and refinement, its PPL is higher than Hankel's at every tier in both seeds, by 0.3--0.5 at full capacity and 0.8--0.9 at $T_1$ (Table~\ref{tab:rev-lti}). This supports the oscillatory mode bank in this parameterization.

The joint LTI control removes the clock's input dependence and fixes both gates, while its constant clock bias remains learnable. The architecture, capacity map, training budget and evaluator are unchanged. It raises PPL by 1.1--1.8 at every tier in both seeds. These results support selective dynamics under the same training recipe. Retrieval ranges overlap at $T_4$ and $T_7$, so the repeated evidence is clearest in next-token prediction.
\end{revblockr}

\begin{figure}[htbp]
\begin{revblockr}
\centering
\includegraphics[width=\linewidth]{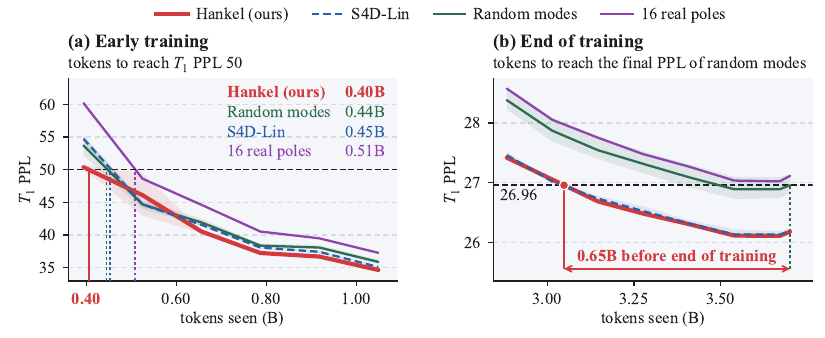}
\revfigcaptionr{\revr{$T_1$ PPL of the raw model during capacity-mixed training at 370M (trainer validation every 500 steps). Lines average validation NLL across seeds before exponentiation. Bands span the seeds (Hankel two seeds, S4D-Lin and random modes three, sixteen real poles one). (a) The Hankel models reach $T_1$ PPL 50 after 0.40B tokens, against 0.44 to 0.51B for the other initializations. (b) They reach the final $T_1$ PPL of the random-mode models (26.96) after 3.05B tokens, 0.65B tokens before the end of training. Threshold-crossing times use log-linear interpolation between evaluations. S4D-Lin and Hankel reach similar final $T_1$ PPL.}}{fig:rev-init-curves}
\end{revblockr}
\end{figure}

\begin{table}[htbp]
\begin{revblockr}
\centering\footnotesize
\setlength{\tabcolsep}{4pt}
\revcaptionr{Joint LTI, real-pole and S4D-Lin controls against the Hankel control (370M, 3.7B tokens, common 2060-step shared refinement, common evaluator). PPL at the four tiers, the seven-task mean (HellaSwag and ARC-Challenge normalized accuracy, the others accuracy), and teacher-forced full-key accuracy on the 2K NIAH test. Hankel control: range over seeds 197, 313 and 511. Controls: range over seeds 197 and 511. S4D-Lin initialization: range over seeds 197, 313 and 511, PPL only. The real-pole control has 372.6M parameters at full capacity.}{tab:rev-lti}
\begin{tabular*}{\linewidth}{@{\extracolsep{\fill}}lcccc@{}}
\toprule
& $T_1$ & $T_4$ & $T_7$ & $T_{10}$ \\
\midrule
\multicolumn{5}{l}{PPL} \\
Hankel control (\ours{}) & 25.32--25.38 & 19.14--19.23 & 17.38--17.46 & 16.22--16.32 \\
S4D-Lin initialization & 25.29--25.35 & 19.16--19.17 & 17.30--17.35 & 16.12--16.16 \\
16 real poles, no oscillation & 26.18--26.27 & 19.79--19.93 & 17.76--17.89 & 16.64--16.73 \\
LTI, input-independent dynamics & 27.01--27.09 & 20.52--20.57 & 18.61--18.67 & 17.43--17.46 \\
\midrule
\multicolumn{5}{l}{Seven-task mean} \\
Hankel control (\ours{}) & 35.5--36.5 & 38.6--39.4 & 39.7--40.1 & 40.6--41.0 \\
16 real poles, no oscillation & 35.4--36.0 & 38.3--38.8 & 39.9--40.1 & 40.2--40.7 \\
LTI, input-independent dynamics & 35.2--35.7 & 37.5--38.1 & 39.2 & 39.7--40.3 \\
\midrule
\multicolumn{5}{l}{Full-key accuracy, 2K NIAH} \\
Hankel control (\ours{}) & 0.06--0.29 & 0.01--0.26 & 0.06--0.26 & 0.15--0.33 \\
16 real poles, no oscillation & 0.05--0.08 & 0.10--0.24 & 0.18--0.30 & 0.25--0.36 \\
LTI, input-independent dynamics & 0.01--0.04 & 0.05--0.10 & 0.07--0.08 & 0.10--0.15 \\
\bottomrule
\end{tabular*}
\end{revblockr}
\end{table}

\begin{table}[htbp]
\begin{revblockr}
\centering\footnotesize
\setlength{\tabcolsep}{3.5pt}
\revcaption{Compact-tier PPL during capacity-mixed training at 370M (3.7B tokens). Trainer validation PPL of the raw model, recorded every 500 steps. Columns give tokens seen. The S4D-Lin and random-mode seed-511 runs and the sixteen-real-pole run (the seed-197 control of Table~\ref{tab:rev-lti}) record $T_1$, $T_4$, $T_7$ and $T_{10}$ only, so they have no $T_2$ row. }{tab:rev-init-early}
\begin{tabular*}{\linewidth}{@{\extracolsep{\fill}}lrrrrrrr@{}}
\toprule
& \multicolumn{7}{c}{tokens seen (B)} \\
\cmidrule(lr){2-8}
Initialization / seed & 0.26 & 0.39 & 0.79 & 1.05 & 1.57 & 2.62 & 3.70 \\
\midrule
\multicolumn{8}{l}{$T_1$}\\
Hankel, 197 & 101.1 & 50.7 & 37.3 & 34.9 & 31.6 & 28.2 & 26.19 \\
Hankel, 511 & 88.9 & 50.1 & 37.2 & 34.4 & 31.4 & 28.2 & 26.18 \\
S4D-Lin, 197 & 180.4 & 55.3 & 38.2 & 35.2 & 31.9 & 28.2 & 26.13 \\
S4D-Lin, 313 & 170.2 & 54.7 & 37.9 & 35.1 & 31.8 & 28.4 & 26.23 \\
S4D-Lin, 511 & 195.5 & 54.2 & 38.1 & 35.0 & 31.8 & 28.2 & 26.21 \\
Random modes, 197 & 96.5 & 53.9 & 38.4 & 35.9 & 32.5 & 29.1 & 26.91 \\
Random modes, 313 & 103.4 & 55.1 & 38.7 & 36.2 & 32.9 & 29.3 & 27.14 \\
Random modes, 511 & 82.6 & 52.1 & 38.1 & 35.6 & 32.4 & 29.0 & 26.83 \\
16 real poles, 197 & 128.6 & 60.2 & 40.5 & 37.3 & 33.4 & 29.3 & 27.12 \\
\midrule
\multicolumn{8}{l}{$T_2$}\\
Hankel, 197 & 63.9 & 41.5 & 31.4 & 29.3 & 26.6 & 23.6 & 22.21 \\
Hankel, 511 & 61.0 & 41.1 & 31.3 & 29.1 & 26.5 & 23.6 & 22.20 \\
S4D-Lin, 197 & 115.1 & 44.9 & 32.1 & 29.6 & 26.7 & 23.6 & 22.23 \\
S4D-Lin, 313 & 111.6 & 44.7 & 32.0 & 29.6 & 26.7 & 23.7 & 22.20 \\
Random modes, 197 & 63.0 & 43.7 & 32.4 & 30.1 & 27.4 & 24.2 & 22.80 \\
Random modes, 313 & 67.7 & 44.9 & 32.7 & 30.5 & 27.6 & 24.5 & 22.92 \\
\bottomrule
\end{tabular*}
\end{revblockr}
\end{table}

\begin{table}[htbp]
\begin{revblockr}
\centering\footnotesize
\setlength{\tabcolsep}{4pt}
\revcaption{PPL after 3.7B-token dense pretraining and the same 2060-step shared refinement at 370M. The two-rate map defines each export, with channel order unchanged. Both initializations use seeds 197 and 511 and the common evaluator. Bold compares the two seed means within each tier.}{tab:rev-init-dense}
\begin{tabular*}{\linewidth}{@{\extracolsep{\fill}}llrrrr@{}}
\toprule
Initialization & Seed & $T_1$ & $T_4$ & $T_7$ & $T_{10}$ \\
\midrule
Hankel & 197 & 45.65 & 23.75 & 18.63 & 16.05 \\
Hankel & 511 & 45.41 & 24.06 & 18.71 & 16.07 \\
Hankel & mean & \textbf{45.53} & \textbf{23.91} & \textbf{18.67} & 16.06 \\
\midrule
S4D-Lin & 197 & 63.62 & 29.40 & 19.94 & 15.96 \\
S4D-Lin & 511 & 63.91 & 29.65 & 19.81 & 15.98 \\
S4D-Lin & mean & 63.77 & 29.52 & 19.87 & \textbf{15.97} \\
\bottomrule
\end{tabular*}\par
\end{revblockr}
\end{table}

\subsection[Mode budget]{Mode budget}
\label{app:init-r}

Increasing the mode count improves the initial filter approximation but does not improve next-token loss in this sweep. With $m=16$, PPL is higher at every reported tier than with $m=8$, while needle first-token accuracy increases only slightly: from $0.259$ to $0.263$ at $T_4$ and from $0.336$ to $0.347$ at full capacity.

Each damped-rotation mode contributes two recurrent-state coordinates for every channel value. Doubling $m$ from 8 to 16 therefore doubles recurrent-state storage at fixed channel shape and increases the full model from 370.9M to 384.0M parameters. We use $m=8$ in the main recipe because it gives lower PPL at half the mode-state storage. The $m=16$ model provides only the small retrieval gain shown above.

\subsection[Channel budget]{Channel budget}
\label{app:sel-kbar}

This sweep compares channel-budget configurations with $\bar K P$ fixed. The full recurrent-state size therefore remains
\[
2\bar K mP=2md.
\]
Increasing $\bar K$ creates more independently controlled channel groups with fewer value coordinates per group, while also changing the number of selective-projection parameters. The sampler retains its fixed integer budget indices, so their fractions $\xi=c/\bar K$ change with $\bar K$. Shared refinement uses teacher index $c=32$: this is full capacity for $\bar K\leq32$, but $\xi=1/2$ for $\bar K=64$, retaining 45 spectral channels under the two-rate map. The comparison therefore includes the corresponding training-capacity schedules.

The default $\bar K=32$ gives the lowest full-capacity PPL among the elastic runs and is within 0.20 PPL of the best intermediate-tier values. Increasing to $\bar K=64$ improves $T_1$ PPL to 23.46 but raises $T_7$ and full-capacity PPL to 17.78 and 16.60 and adds 13.1M parameters. We use $\bar K=32$ to balance the full and reduced capacities within the 370M parameter budget. Retrieval values are single-seed measurements and are reported alongside the language-model results.

\begin{table}[htbp]
\centering
\footnotesize
\setlength{\tabcolsep}{4pt}
\caption{Channel-count controls at 370M after 3.7B pretraining tokens. Channel value width $P$ is adjusted so that the full recurrent-state size is constant. Compact elastic PPL uses final checkpoints; full PPL is measured from raw checkpoints. The retrieval column reports teacher-forced full-key accuracy. For both $\bar K=64$ rows, retrieval uses $\xi=1/2$ (45 retained spectral channels); the other retrieval values use full capacity. Dashes denote configurations without a reported evaluation; dense PPL entries use their full-capacity models.}
\label{tab:sel-kbar}
\begin{tabular}{@{}lccccccc@{}}
\toprule
& & \multicolumn{4}{c}{validation PPL $\downarrow$}
& full-key acc. $\uparrow$ & params, full \\
\cmidrule(lr){3-6}
Arm & $P$ & $T_1$ & $T_4$ & $T_7$ & $T_{10}$ & NIAH & (M) \\
\midrule
\ours{} ($\bar K=32$) & 28 & 25.32 & 19.15 & 17.38 & 16.46 & 0.334 & 370.9 \\
\ours{} ($\bar K=16$) & 56 & -- & 18.95 & 17.35 & 16.49 & 0.316 & 364.3 \\
\ours{} ($\bar K=64$) & 14 & 23.46 & 19.16 & 17.78 & 16.60 & 0.312 & 384.0 \\
\ours{} ($\bar K=16$, dense control) & 56 & -- & -- & -- & 16.09 & 0.304 & 364.3 \\
\ours{} ($\bar K=32$, dense control) & 28 & -- & -- & -- & 15.98 & 0.181 & 370.9 \\
\ours{} ($\bar K=64$, dense control) & 14 & -- & -- & -- & 15.97 & 0.040 & 384.0 \\
\bottomrule
\end{tabular}
\end{table}

\section[Elasticity Ablations]{Elasticity Ablations}
\label{app:elastic_ablations}

These experiments isolate the choices that turn the nested tensor structure into a trained elastic family: the two-rate capacity map, FFN importance ordering, capacity sampling, gradient allocation, distillation, and the separation between recurrent and feed-forward capacity.

\subsection[Nesting rule and ordering]{Nesting rule and ordering}
\label{app:el-nesting}

The nesting sweep first asks which resources must shrink to obtain a materially smaller model (Table~\ref{tab:el-nesting}). Truncating spectral channels alone leaves the FFNs unchanged, so the smallest mixer-only export still contains 338.2M parameters, or $91.2\%$ of the full model. Substantial parameter reduction therefore requires reducing FFN width as well as spectral-channel count. The two-rate map retains a larger fraction of spectral channels than FFN units while concentrating most parameter reduction in the FFNs.

The exponent sweeps show the resulting quality--size trade-off. Flatter channel or FFN reduction generally improves the smallest-tier PPL but also produces a larger export. For example, the flatter-FFN map retains $27.9\%$ of the full parameters at $T_1$, compared with $24.8\%$ for the control. No tested map lowers $T_1$ PPL at an equal or smaller export. The control also has the lowest full-capacity PPL among maps that reduce both axes, tied with $(1.0,0.7)$. These results support the default exponents as a balance across capacities at the intended parameter budget.

At the same 92.1M shape, the recovered dense control has PPL 45.66 versus 25.32 for ESSH, supporting direct optimization of reduced capacities during training. FFN importance ordering provides a smaller, consistent improvement: removing it raises PPL at every measured tier by 0.07--0.37, whereas ordering every 1000 steps instead of every 500 changes PPL by at most 0.07. Training at reduced capacities accounts for the larger observed difference, while ordering improves the selection of retained FFN units.

\begin{revblockr}
FFN-only nesting keeps all 32 spectral channels and shrinks FFN width under the same 3.7B-token recipe and shared refinement (Table~\ref{tab:rev-ffnonly}). Its smallest trained export is 124.9M. Even removing all FFNs leaves 98.4M parameters, above ESSH's 92.1M $T_1$. At the small end, ESSH reaches PPL 21.65--21.70 at 118.9M, versus 22.25 for FFN-only at 124.9M. Against the ESSH seed means, FFN-only is up to 0.49 PPL better at intermediate sizes, with similar full-model PPL. Its seven-task means are 39.51\% and 40.33\% at 185.4M and 276.2M, compared with ESSH's 38.6--39.4\% and 39.7--40.1\%, consistent with the PPL comparison.

The spectral-channel axis also reduces recurrent state and decoding cost. FFN-only retains 1.09\,MB of recurrent state at every size, whereas ESSH reaches 0.27\,MB at $T_1$. At 118.9M versus 124.9M, ESSH's compiled batch-one decode takes 0.97 versus 1.14\,ms on A100 and 0.63 versus 0.75\,ms on B300 (Table~\ref{tab:rev-ffnonly}). Thus the two-rate map extends the attainable size range and reduces execution costs, with its quality advantage concentrated at the compact end.
\end{revblockr}

\begin{table}[htbp]
\begin{revblockr}
\centering\footnotesize
\setlength{\tabcolsep}{4pt}
\revcaption{FFN-only nesting against ESSH at nearby retained parameter counts (370M, 3.7B tokens, same capacity-mixed recipe, common evaluator, after the common 2060-step shared refinement). FFN-only keeps all 32 spectral channels and shrinks the FFN width by the same map, and its smallest trained export is 124.9M. Each row pairs the nearest evaluated capacities. Where they differ by more than 1\%, the ESSH export is the smaller one. ESSH: seeds 197/313/511; FFN-only: seed 197. Seven-task means (\%) use the same evaluator and checkpoint as PPL at the selected deployment sizes. Dashes denote unmeasured task scores. Decode is the compiled batch-one step with a 2K cache for the 118.9M ESSH and 124.9M FFN-only exports, measured as in Table~\ref{tab:sys-small-devices}.}{tab:rev-ffnonly}
\begin{tabular*}{\linewidth}{@{\extracolsep{\fill}}lrrrrrr@{}}
\toprule
& \multicolumn{3}{c}{ESSH} & \multicolumn{3}{c}{FFN-only} \\
\cmidrule(lr){2-4}\cmidrule(lr){5-7}
Tier & M & PPL $\downarrow$ & 7-task $\uparrow$ & M & PPL $\downarrow$ & 7-task $\uparrow$ \\
\midrule
$T_1$ & 92.1 & 25.32--25.38 & 35.5--36.5 & \multicolumn{3}{c}{below the 98.4M parameter floor} \\
 & 106.2 & 23.07--23.09 & -- & \multicolumn{3}{c}{below the smallest trained export} \\
 & 118.9 & 21.65--21.70 & -- & 124.9 & 22.25 & -- \\
 & 133.0 & 20.88--20.90 & -- & 136.2 & 21.00 & -- \\
 & 154.7 & 20.00--20.02 & -- & 158.9 & 19.58 & -- \\
$T_4$ & 185.3 & 19.14--19.23 & 38.6--39.4 & 185.4 & 18.70 & 39.51 \\
$T_7$ & 274.3 & 17.38--17.46 & 39.7--40.1 & 276.2 & 17.16 & 40.33 \\
$T_{10}$ & 370.9 & 16.22--16.32 & 40.6--41.0 & 370.9 & 16.29 & 41.18 \\
\midrule
\multicolumn{2}{@{}l}{Recurrent state (MB)} & 0.27--1.09 & & & 1.09 & \\
\multicolumn{2}{@{}l}{Decode, A100 / B300 (ms)} & 0.97 / 0.63 & & & 1.14 / 0.75 & \\
\bottomrule
\end{tabular*}
\end{revblockr}
\end{table}

\begin{table}[htbp]
\centering
\footnotesize
\setlength{\tabcolsep}{4pt}
\caption{Capacity-map and ordering controls. In the upper block, compact PPL uses final checkpoints and full PPL uses raw checkpoints. Different maps produce different parameter counts at the same nominal tier, so rows are not parameter matched. The lower diagnostic block reports the recorded $T_4$ needle first-token accuracy and raw $T_1$ PPL. For the dense control, the raw $T_1$ value is before compact recovery; the upper block reports the recovered compact checkpoint.}
\label{tab:el-nesting}
\begin{tabular}{@{}lcccccc@{}}
\toprule
& & & \multicolumn{4}{c}{validation PPL $\downarrow$} \\
\cmidrule(lr){4-7}
Arm & \shortstack{exponents\\(channel, FFN)}
& $r$ at $T_1$
& $T_1$ & $T_4$ & $T_7$ & $T_{10}$ \\
\midrule
\ours{} (control) & $(0.5,\,0.8)$ & 0.248 & 25.32 & 19.15 & 17.38 & 16.46 \\
\ours{} (alternative map) & $(1.0,\,0.7)$ & 0.257 & 25.54 & 19.50 & 17.50 & 16.46 \\
\ours{} (steeper channels) & $(0.65,\,0.8)$ & 0.237 & 26.34 & 19.40 & 17.50 & 16.50 \\
\ours{} (flatter channels) & $(0.35,\,0.8)$ & 0.263 & 24.41 & 19.02 & 17.35 & 16.53 \\
\ours{} (flattest channels) & $(0.25,\,0.8)$ & 0.278 & 23.74 & 18.76 & 17.26 & 16.47 \\
\ours{} (flatter feed-forward) & $(0.5,\,0.7)$ & 0.279 & 23.89 & 18.81 & 17.25 & 16.52 \\
\ours{} (mixer-only nesting) & FFN fixed & 0.912 & 16.82 & 16.22 & 16.10 & 16.12 \\
\ours{} (no FFN importance ordering) & $(0.5,\,0.8)$ & 0.248 & 25.56 & 19.52 & 17.64 & 16.53 \\
\ours{} (ordering every 1000 steps) & $(0.5,\,0.8)$ & 0.248 & 25.39 & 19.18 & 17.41 & 16.50 \\
\ours{} (dense control, reduced) & -- & 0.248 & 45.66 & 23.75 & 18.63 & 15.98 \\
\bottomrule
\end{tabular}

\vspace{4pt}

\begin{tabular}{@{}lccc@{}}
\toprule
Arm &
\shortstack{needle first-token\\acc. $T_4\uparrow$} &
\shortstack{raw PPL\\$T_1\downarrow$} &
\shortstack{params (M)\\at $T_1$} \\
\midrule
\ours{} (control) & 0.259 & 26.19 & 92.1 \\
\ours{} (alternative map) & 0.152 & 26.42 & 95.3 \\
\ours{} (flatter channels) & 0.331 & 25.26 & 97.6 \\
\ours{} (flattest channels) & 0.220 & 24.54 & 103.0 \\
\ours{} (mixer-only nesting) & 0.134 & 17.25 & 338.2 \\
\ours{} (no FFN importance ordering) & 0.281 & 26.41 & 92.1 \\
\ours{} (dense control, reduced) & 0.134 & 53162.13 & 92.1 \\
\bottomrule
\end{tabular}
\end{table}

\subsection[Capacity sampling and optimization controls]{Capacity sampling and optimization controls}
\label{app:el-constants}

The reserved full-capacity share $f$ directly trades optimization between the full model and the reduced-capacity models. Reducing $f$ from $0.75$ to $0.5$ improves $T_1$ PPL from $25.32$ to $23.50$ but increases full-capacity PPL from $16.46$ to $17.01$. Increasing $f$ to $0.875$ moves the endpoints in the opposite direction: full PPL improves to $16.26$, while $T_1$ worsens to $27.56$. Capacity-mixed training therefore allocates a finite update budget across predictors that share parameters but use different retained configurations.

Full-tier distillation primarily benefits the reduced-capacity models in this sweep. Removing it increases $T_1$ PPL from $25.32$ to $26.42$, while full PPL changes only from $16.46$ to $16.48$. The $T_4$ needle first-token accuracy also decreases from $0.259$ to $0.193$. With coefficient 0.25, $T_1$ PPL is 25.92 and $T_4$ first-token accuracy is 0.290. Increasing the coefficient to 1.0 lowers $T_1$ PPL to 24.69 while the measured $T_4$ first-token accuracy is 0.055. We retain 0.5 to balance compact-model PPL and retrieval in this sweep.

The main recipe scales reduced-capacity FFN parameter gradients by $\alpha_{\mathrm{ff}}=0.5$. Here, $\alpha_{\mathrm{mix}}$ denotes the analogous gradient scale applied to retained mixer parameters on reduced-capacity micro-batches. The arm with $\alpha_{\mathrm{ff}}=\alpha_{\mathrm{mix}}=0.5$ applies the same scaling to both retained FFN and mixer parameters. Both alternatives raise PPL at all four measured tiers: by 0.06--0.93 for $\alpha_{\mathrm{ff}}=0.25$, and by 0.04--0.40 when mixer gradients are also scaled. These results support scaling only the retained FFN gradients, at 0.5.

A weighted-label control tests whether increasing the CE coefficient reproduces the distillation result (Table~\ref{tab:distillation-weight-control}). At $T_1$, increasing CE to $1.5\,\mathrm{CE}$ lowers end-of-pretraining PPL from 27.61 to 26.76, while $\mathrm{CE}+0.5\,\mathrm{KL}$ reaches 26.19. Thus label weighting recovers part of the improvement, while the distillation objective gives lower PPL at all four evaluated capacities in these runs.

\begin{table}[htbp]
\centering\footnotesize
\setlength{\tabcolsep}{4pt}
\caption{Reduced-capacity objective controls at 370M, 3.7B tokens and seed 197. All four PPL values use the end-of-pretraining model and the common 16-batch evaluator. Full-capacity micro-batches always use CE. All three runs use 128 sequences per optimizer update.}
\label{tab:distillation-weight-control}
\begin{tabular*}{\linewidth}{@{\extracolsep{\fill}}lrrrr@{}}
\toprule
Reduced-capacity objective & PPL $T_1\downarrow$ & PPL $T_4\downarrow$ & PPL $T_7\downarrow$ & PPL $T_{10}\downarrow$ \\
\midrule
$\mathrm{CE}$ & 27.61 & 20.08 & 18.03 & 16.48 \\
$1.5\,\mathrm{CE}$ & 26.76 & 19.90 & 18.01 & 16.58 \\
$\mathrm{CE}+0.5\,\mathrm{KL}$ & \textbf{26.19} & \textbf{19.59} & \textbf{17.83} & \textbf{16.46} \\
\bottomrule
\end{tabular*}
\end{table}

The learning-rate rows test sensitivity at 3.7B tokens. The $1.5\times10^{-3}$ setting stays within 0.1 PPL of the control at every tier, while $2.0\times10^{-3}$ gives lower PPL across the four tiers. The reported 1.5B run retains the configuration fixed before this study (Appendix~\ref{app:repro-config}).

\begin{table}[htbp]
\centering
\footnotesize
\setlength{\tabcolsep}{4pt}
\caption{One-setting controls at 370M and 3.7B pretraining tokens. Compact PPL uses final checkpoints; full PPL and the seven-task mean use the corresponding full-model evaluations. The retrieval column reports NIAH first-token accuracy at the exact $T_4$ shape.}
\label{tab:el-constants}
\begin{tabular}{@{}lcccccc@{}}
\toprule
& \multicolumn{4}{c}{validation PPL $\downarrow$}
& \shortstack{needle first-token\\acc. $\uparrow$}
& 7-task $\uparrow$ \\
\cmidrule(lr){2-5}
Arm & $T_1$ & $T_4$ & $T_7$ & $T_{10}$ & $T_4$ & $T_{10}$ \\
\midrule
\ours{} (control) & 25.32 & 19.15 & 17.38 & 16.46 & 0.259 & 0.408 \\
\ours{} ($f=0.5$) & 23.50 & 18.67 & 17.44 & 17.01 & 0.216 & 0.407 \\
\ours{} ($f=0.875$) & 27.56 & 19.92 & 17.53 & 16.26 & 0.199 & 0.410 \\
\ours{} ($\alpha_{\mathrm{ff}}=0.25$) & 26.25 & 19.59 & 17.59 & 16.52 & 0.135 & 0.410 \\
\ours{} ($\alpha_{\mathrm{ff}}=\alpha_{\mathrm{mix}}=0.5$) & 25.72 & 19.31 & 17.48 & 16.50 & 0.162 & 0.413 \\
\ours{} (no full-tier distillation) & 26.42 & 19.53 & 17.51 & 16.48 & 0.193 & 0.412 \\
\ours{} (KL coefficient 0.25) & 25.92 & 19.35 & 17.45 & 16.49 & 0.290 & 0.406 \\
\ours{} (KL coefficient 1.0) & 24.69 & 18.96 & 17.35 & 16.53 & 0.055 & 0.413 \\
\ours{} (lr $1.5\times10^{-3}$) & 25.23 & 19.08 & 17.32 & 16.47 & 0.150 & 0.406 \\
\ours{} (lr $2.0\times10^{-3}$) & 24.99 & 18.90 & 17.15 & 16.34 & 0.309 & 0.409 \\
\bottomrule
\end{tabular}
\end{table}

The sampling schedule determines how often the reduced-capacity models are optimized directly. Sampling only five deployment tiers,
$T_1,T_3,T_5,T_7,T_{10}$,
leaves the omitted $T_2$ with the largest measured penalty, $+2.59$ PPL, despite sampling both family endpoints. Uniformly sampling all ten deployment tiers instead improves some larger tiers but worsens $T_1$ and $T_2$ by $1.38$ and $1.37$ PPL relative to the control menu. These results show that endpoint coverage alone is insufficient and that the set of training capacities and deployment grid serve different roles.

The progressive schedule initially samples training capacities $\{16,24\}/32$, then adds $\{8,12\}/32$, $\{4,5,6\}/32$, and $\{2,3\}/32$ at 15\%, 30\%, and 45\% of training, respectively. The share ramp increases the reserved full-capacity fraction linearly from 0.75 to 1.0 over the second half. Gap-proportional sampling updates every 500 steps using each reduced capacity's nonnegative validation cross-entropy gap to full capacity, with linear interpolation in $\log_2 K$ for unevaluated capacities. It uses full capacity until the initial validation at step 0, then samples reduced capacities using the resulting weights.

The adaptive-share variant also updates every 500 steps and activates its weights after the second validation, at step 500, once loss changes are available. In both weighted-sampling variants, activation bypasses the remainder of the default 250M-token full-capacity warm-up. The adaptive-share variant assigns larger exponential weights to capacities whose validation loss has decreased less over the preceding interval, using a temperature equal to half the mean absolute loss decrease. The resulting full-capacity share is clipped to $[0.5,0.9]$ and changes by at most 0.05 per subsequent update. Reduced-capacity weights are interpolated in $\log_2 K$ and mixed equally with uniform sampling. Compared with the fixed-share control, the ramp improves full PPL by 0.14 but worsens $T_1$ by 2.34. Gap-proportional sampling reverses that emphasis, improving $T_1$ by 1.66 while worsening full PPL by 0.21. Adaptation therefore redistributes quality across the family without improving all measured tiers together.

\begin{table}[htbp]
\centering
\small
\setlength{\tabcolsep}{4pt}
\caption{Raw PPL differences relative to the fixed-share control for alternative 370M sampling schedules after 3.7B tokens. Negative values indicate lower PPL than the control. Columns evaluate the exact deployment shapes.}
\label{tab:el-schedule}
\resizebox{\linewidth}{!}{%
\begin{tabular}{@{}lccccc@{}}
\toprule
Schedule & $T_1$ & $T_2$ & $T_4$ & $T_7$ & $T_{10}$ \\
\midrule
\ours{} (fixed share $f=0.75$, control) & 0 & 0 & 0 & 0 & 0 \\
\ours{} (fixed share $f=0.5$) & $-2.16$ & $-1.30$ & $-0.58$ & $-0.02$ & $+0.55$ \\
\ours{} (adaptive share, 0.5 to 0.9) & $-2.59$ & $-0.96$ & $-0.25$ & $+0.17$ & $+0.52$ \\
\ours{} (sampled capacities = the ten tiers) & $+1.38$ & $+1.37$ & $+0.01$ & $-0.28$ & $-0.08$ \\
\ours{} (sampled capacities = five tiers $T_1,T_3,T_5,T_7,T_{10}$) & $+0.10$ & $+2.59$ & $+0.23$ & $-0.21$ & $-0.03$ \\
\ours{} (progressive capacity reduction, four stages) & $+0.92$ & $+0.11$ & $-0.10$ & $-0.05$ & $+0.07$ \\
\ours{} (share ramp 0.75 to 1.0 over the second half) & $+2.34$ & $+1.47$ & $+0.79$ & $+0.22$ & $-0.14$ \\
\ours{} (gap-proportional sampling) & $-1.66$ & $-0.48$ & $+0.14$ & $+0.33$ & $+0.21$ \\
\bottomrule
\end{tabular}}
\end{table}

\begin{figure}[htbp]
\centering
\includegraphics[width=\linewidth]{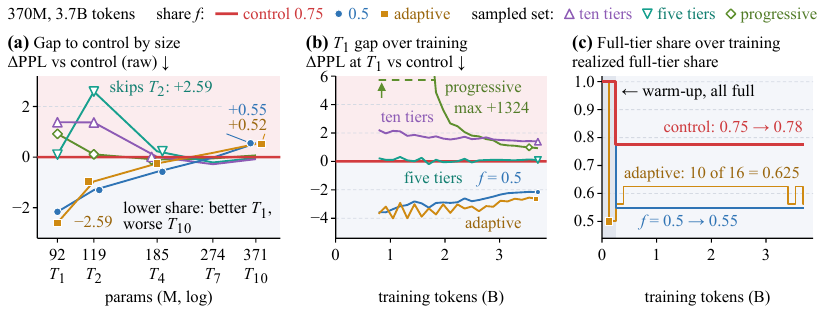}
\caption{Selected capacity-sampling schedules. Panel (a) shows PPL differences at the five evaluated tiers, panel (b) tracks $T_1$ during training, and panel (c) shows the realized full-capacity share. The progressive schedule's early maximum of +1324 PPL is clipped by the plotting range. The complete schedule comparison, including the share ramp and gap-proportional variant, is in Table~\ref{tab:el-schedule}.}
\label{fig:app-el-schedule}
\end{figure}

\subsection[Full-capacity cost of shared training]{Full-capacity cost of shared training}
\label{app:el-cost}

The dense ESSH control isolates the full-capacity PPL cost of optimizing multiple capacities with shared parameters. At matched architecture and pretraining budget, the raw full-capacity PPL gap between elastic and dense training is $+0.48$ after 3.7B tokens and $+0.41$ after 7.4B tokens (Table~\ref{tab:el-cost}). These gaps are larger than the observed seed ranges for either training condition. The independently trained Mamba-3 + SWA reference gives PPL 14.10 and 13.97 for seeds 197 and 313 under the same frontier protocol, a range of 0.13. 

This cost is consistent with the allocation trade-off in the full-share sweep: increasing full-capacity participation improves the full endpoint while weakening the smallest export. The resulting reduced-capacity models benefit from those shared updates: at equal total training exposure, $T_4$ PPL is 19.15 for ESSH versus 23.75 for the truncated-and-recovered dense control (Table~\ref{tab:el-nesting}). Shared refinement subsequently improves the delivered elastic family and is reported separately with its own token budget and checkpoint stage.

\begin{table}[htbp]
\centering
\small
\setlength{\tabcolsep}{5pt}
\caption{Full-capacity PPL at the end of pretraining. ESSH comparisons use the same architecture and token budget within each block. Mamba-3 + SWA is a separate independently trained reference. PPL reports seed 197; range is the maximum minus the minimum across three seeds for 3.7B elastic ESSH and two for each other training condition.}
\label{tab:el-cost}
\begin{tabular}{@{}lccc@{}}
\toprule
Model (370M scale) & tokens (B) & PPL $T_{10}$ $\downarrow$ & seed range \\
\midrule
\ours{} & 3.7 & 16.46 & 0.09 \\
\ours{} (dense control, no capacity mixing) & 3.7 & 15.98 & 0.06 \\
PPL gap (elastic $-$ dense) & 3.7 & $+0.48$ & -- \\
\midrule
\ours{} & 7.4 & 14.85 & 0.05 \\
\ours{} (dense control, no capacity mixing) & 7.4 & 14.44 & 0.05 \\
PPL gap (elastic $-$ dense) & 7.4 & $+0.41$ & -- \\
\midrule
\multicolumn{4}{l}{Independent full-model reference}\\
Mamba-3 + SWA (377.1M) & 7.4 & 14.10 & 0.13 \\
\bottomrule
\end{tabular}
\end{table}

\subsection[Mixer and feed-forward budgets]{Mixer and feed-forward budgets}
\label{app:ratio}

Spectral channels, modes, and FFN width allocate different resources. Channels partition the recurrent representation, modes determine the temporal-state budget, and the FFN provides tokenwise nonlinear capacity. Table~\ref{tab:ratio-dff} varies FFN width while keeping depth, residual width, and the spectral-state configuration fixed.

Increasing FFN width improves PPL at every reported tier and also increases $T_4$ needle first-token accuracy. The marginal full-capacity PPL gain nevertheless decreases across the sweep. Increasing $d_{\mathrm{ff}}$ from 3072 to 4608 adds 90.83M full-model parameters and improves full PPL by $0.66$. Increasing it from 4608 to 6144 adds the same 90.83M parameters but improves full PPL by $0.37$.

Keeping this tokenwise expansion outside the recurrent mixer makes the two budgets independently adjustable: increasing FFN width changes parameters and tokenwise computation without multiplying recurrent-state storage. The two-rate map uses the same separation in the opposite direction, reducing the parameter-heavy FFN faster than the spectral-channel count. We use $d_{\mathrm{ff}}=4608$ to meet the 370M parameter budget used for the baseline comparison (Appendix~\ref{app:setup-baselines}).

\begin{table}[htbp]
\centering
\small
\setlength{\tabcolsep}{5pt}
\caption{FFN-width sweep at fixed depth, residual width, and spectral mixer after 3.7B pretraining tokens. Exported parameter counts vary with FFN width. Compact PPL uses final checkpoints; full-capacity PPL uses the corresponding raw checkpoints.}
\label{tab:ratio-dff}
\begin{tabular}{@{}lcccccc@{}}
\toprule
$d_{\mathrm{ff}}$
& params $T_1/T_{10}$ (M)
& PPL $T_1$ $\downarrow$
& PPL $T_4$ $\downarrow$
& PPL $T_7$ $\downarrow$
& PPL $T_{10}$ $\downarrow$
& \shortstack{needle first-token\\acc. $T_4\uparrow$} \\
\midrule
3072 & 84.6 / 280.0 & 26.44 & 20.26 & 18.29 & 17.12 & 0.191 \\
4608 (\ours{}) & 92.1 / 370.9 & 25.32 & 19.15 & 17.38 & 16.46 & 0.259 \\
6144 & 103.5 / 461.7 & 24.18 & 18.59 & 16.91 & 16.09 & 0.319 \\
\bottomrule
\end{tabular}
\end{table}

\section[Compression and Train-Once Baselines]{Compression and Train-Once Baselines}
\label{app:compression_baselines}

We compare three ways to obtain multiple deployment sizes: joint training of nested predictors, post-hoc recovery from a pretrained parent, and independent training at each target shape. Because these procedures use different training stages and token budgets, we report training exposure, recovery steps, and actual exported parameter counts separately.

\subsection[Elastic-family comparison curves]{Elastic-family comparison curves}
\label{app:comp-family-curves}

The family comparison evaluates ESSH, MatFormer and MatMamba using actual exported parameter counts (Figure~\ref{fig:main_frontier}(a)). MatFormer varies feed-forward width, while MatMamba varies mixer inner width and the corresponding head count~\citep{devvrit2024matformer,shukla2024matmamba}. The local baselines use four training capacities: $1/8$, $1/4$, $1/2$, and full width. The local MatMamba implementation reserves one quarter of micro-batches for full width and samples the other three widths uniformly in the remaining micro-batches, forming a stochastic version of the nested-width recipe.

The baseline rows at the fractions assigned to $T_2$ and $T_{10}$ coincide with trained capacities, while $T_3$--$T_9$ are interpolated and $T_1$ is below the trained range. ESSH also evaluates interpolated capacities: its 370M $T_3$--$T_7$ and $T_9$ are not sampled directly during training. The comparisons therefore assess both directly trained and intermediate exports, with the below-range baseline results separated in the tables.

ESSH uses its final shared model, while MatFormer and MatMamba use their 7.4B-token models without an additional shared stage. Before that stage, ESSH already has PPL 17.94 at $T_4$, 16.22 at $T_7$, and 14.85 at full capacity. These are below the comparable 188.4M MatMamba result of 18.47, the 270.9M MatFormer result of 16.98, and the two full-baseline results of 16.80--16.98. Final ESSH further reaches 16.98 at the 214.5M $T_5$.

The baselines' additional directly trained widths provide useful references: MatFormer has PPL 17.93 at 176.3M and 17.03 at 240.7M, and MatMamba has 19.21 at 140.6M and 17.50 at 217.1M. These values support the intermediate-size comparison without relying solely on interpolated baseline configurations. The $\Delta$NLL columns equal $\ln(\mathrm{PPL}_T/\mathrm{PPL}_{T_{10}})$ within each family. Absolute PPL and parameter count determine cross-family quality comparisons.

\begin{table}[!htbp]
\centering\small
\setlength{\tabcolsep}{3.5pt}
\caption{MatFormer at the ten ESSH deployment budget fractions, using its end-of-pretraining 7.4B-token model. Architecture-specific retained dimensions and parameter counts differ from ESSH. $r$ is the retained parameter fraction; $\Delta$NLL is relative to the full model. $T_1$ lies below the smallest capacity used during MatFormer training. Seed 197. \revr{Latency is a batch-one B300 measurement after a 2K context, with torch.compile and CUDA graphs for every family}; see Appendix~\ref{app:setup-hw}.}
\label{tab:new-frontier-matformer}
\begin{tabular}{@{}lrrrrrr@{}}
\toprule
Tier & Params (M) & $r$ & PPL $\downarrow$ & $\Delta$NLL $\downarrow$ & 7-task $\uparrow$ & ms/token $\downarrow$\\
\midrule
$T_{2}$ & 146.0 & 0.400 & 19.37 & 0.142 & 0.390 & \revr{0.93} \\
$T_{3}$ & 168.7 & 0.462 & 18.48 & 0.095 & 0.396 & \revr{0.93} \\
$T_{4}$ & 191.5 & 0.524 & 17.82 & 0.058 & 0.401 & \revr{0.95} \\
$T_{5}$ & 214.2 & 0.586 & 17.50 & 0.041 & 0.404 & \revr{0.97} \\
$T_{6}$ & 248.2 & 0.679 & 17.02 & 0.012 & 0.408 & \revr{0.96} \\
$T_{7}$ & 270.9 & 0.741 & 16.98 & 0.011 & 0.410 & \revr{0.97} \\
$T_{8}$ & 301.2 & 0.824 & 16.93 & 0.007 & 0.410 & \revr{0.99} \\
$T_{9}$ & 331.5 & 0.907 & 16.87 & 0.004 & 0.411 & \revr{1.00} \\
$T_{10}$ & 365.6 & 1.000 & 16.80 & 0.000 & 0.411 & \revr{1.01} \\
\midrule
\multicolumn{7}{l}{Below the smallest trained capacity}\\
$T_{1}$ & 130.9 & 0.358 & 5844.15 & 5.852 & 0.282 & \revr{0.90} \\
\bottomrule
\end{tabular}
\end{table}

\begin{table}[!htbp]
\centering\small
\setlength{\tabcolsep}{3.5pt}
\caption{MatMamba at the ten ESSH deployment budget fractions, using its end-of-pretraining 7.4B-token model. Architecture-specific retained dimensions and parameter counts differ from ESSH. $r$ is the retained parameter fraction; $\Delta$NLL is relative to the full model. $T_1$ lies below the smallest capacity used during MatMamba training. Seed 197. \revr{Latency is a batch-one B300 measurement after a 2K context, with torch.compile and CUDA graphs for every family}; see Appendix~\ref{app:setup-hw}.}
\label{tab:new-frontier-matmamba}
\begin{tabular}{@{}lrrrrrr@{}}
\toprule
Tier & Params (M) & $r$ & PPL $\downarrow$ & $\Delta$NLL $\downarrow$ & 7-task $\uparrow$ & ms/token $\downarrow$\\
\midrule
$T_{2}$ & 102.4 & 0.277 & 22.06 & 0.262 & 0.371 & \revr{1.04} \\
$T_{3}$ & 131.1 & 0.354 & 20.53 & 0.190 & 0.379 & \revr{1.11} \\
$T_{4}$ & 159.7 & 0.432 & 19.10 & 0.118 & 0.386 & \revr{1.15} \\
$T_{5}$ & 188.4 & 0.509 & 18.47 & 0.084 & 0.390 & \revr{1.18} \\
$T_{6}$ & 226.6 & 0.613 & 17.47 & 0.028 & 0.398 & \revr{1.31} \\
$T_{7}$ & 255.3 & 0.690 & 17.38 & 0.023 & 0.400 & \revr{1.34} \\
$T_{8}$ & 293.5 & 0.793 & 17.24 & 0.015 & 0.399 & \revr{1.33} \\
$T_{9}$ & 331.7 & 0.897 & 17.11 & 0.008 & 0.403 & \revr{1.24} \\
$T_{10}$ & 369.9 & 1.000 & 16.98 & 0.000 & 0.404 & \revr{1.42} \\
\midrule
\multicolumn{7}{l}{Below the smallest trained capacity}\\
$T_{1}$ & 83.3 & 0.225 & 3029.37 & 5.184 & 0.275 & \revr{0.92} \\
\bottomrule
\end{tabular}
\end{table}

The four independent Transformer++ models test quality at similar deployment sizes (Table~\ref{tab:external-four-sizes}). ESSH has lower PPL and higher seven-task mean at $T_4$, $T_7$, and $T_{10}$. The independently trained smallest model is stronger on both metrics at $T_1$. For example, the near-274M comparison gives PPL 15.91 versus 16.69 and task mean 42.00\% versus 40.79\%. Shared training therefore retains both next-token and task quality at the intermediate sizes, with the largest sharing cost at the smallest export.

\begin{table}[htbp]
\centering\small
\setlength{\tabcolsep}{4pt}
\caption{PPL and seven-task mean accuracy (\%) at four deployment sizes. ESSH follows the complete shared-training procedure; Transformer++ uses four independent runs. Both use 7.4B pretraining tokens with the same data, tokenizer and sequence length. ESSH's additional shared stage is specified in Appendix~\ref{app:setup-budget}.}
\label{tab:external-four-sizes}
\begin{tabular*}{\linewidth}{@{\extracolsep{\fill}}lrrrrrr@{}}
\toprule
 & \multicolumn{2}{c}{Params (M)} & \multicolumn{2}{c}{PPL $\downarrow$} & \multicolumn{2}{c}{7-task $\uparrow$}\\
\cmidrule(lr){2-3}\cmidrule(lr){4-5}\cmidrule(lr){6-7}
Target & ESSH & TF++ & ESSH & TF++ & ESSH & TF++\\
\midrule
$T_{1}$ & 92.1 & 92.7 & 23.61 & 20.36 & 37.56 & 38.41 \\
$T_{4}$ & 185.3 & 186.3 & 17.68 & 17.76 & 40.45 & 39.84 \\
$T_{7}$ & 274.3 & 273.6 & 15.91 & 16.69 & 42.00 & 40.79 \\
$T_{10}$ & 370.9 & 365.6 & 14.71 & 15.88 & 42.84 & 41.38 \\
\bottomrule
\end{tabular*}
\end{table}

\FloatBarrier
\revsubsection{Elastic baselines under the ESSH recipe}{app:rev-same-recipe}
\begin{revblockr}
We apply the ESSH capacity-sampling and distillation settings to MatMamba and to a head-nested Mamba-3 + SWA hybrid: ten sampled capacities, a 75\% reserved full-capacity share, KL weight 0.5, 3.7B pretraining tokens and the common 2060-step shared refinement. MatMamba retains its native inner-width and head nesting. The Mamba-3 hybrid uses the two-rate map for heads and FFN width. The hybrid uses the ESSH depth, residual width and attention layout (22 layers, $d=896$, window attention at layers 12, 17 and 21). Its retained head count grows with the square root of capacity. This matches the spectral-channel reduction rule, but produces different parameter counts because a Mamba-3 head and an ESSH channel have different sizes. Table~\ref{tab:rev-same-recipe} reports the resulting models, and Table~\ref{tab:rev-same-recipe-matched} pairs nearby exported sizes using a denser evaluation grid.

With this shared reduction rule, the head-nested hybrid's smallest export is 105.6M. At a similar size, ESSH reaches PPL 23.07--23.09 at 106.2M, compared with 23.68--23.84 for the hybrid. ESSH remains ahead near 119M. The models are level near 132M, and the hybrid is 0.1--0.4 PPL better above that. Table~\ref{tab:rev-aligned-m3} below tests a head-count map that also reaches ESSH's 92M tier. These comparisons place the spectral mixer's PPL advantage at the compact end.

ESSH also uses less recurrent state. ESSH stores 0.27--1.09 MB per sequence, versus 4.4--17.7 MB for the hybrid under the shared reduction rule (Table~\ref{tab:rev-same-recipe-aux}). Each ESSH channel uses eight damped rotations, while each Mamba-3 head carries a larger expanded state. Both models also retain the same 22.02 MB window KV cache. The fixed attention cache makes the proportional reduction in total memory smaller than the reduction in recurrent state.

The MatMamba comparison holds the pretraining token budget fixed while changing the recipe (Table~\ref{tab:rev-matmamba-own}). With the same shared refinement, the ESSH recipe reduces MatMamba's mean smallest/full PPL from 60.08/19.03 to 33.87/16.71. ESSH itself reaches 25.34/16.27 at its smallest/full tiers and has lower PPL over the measured common size range. Thus the recipe improves both endpoints of another nested architecture, while the ESSH parameterization provides a further advantage at compact sizes. The smallest exports are 83.3M for MatMamba and 92.1M for ESSH.

{\color{revred}
At matched retained sizes, ESSH reaches PPL 25.32--25.38 at 92.1M versus 29.47--29.51 for MatMamba at 92.9M (Table~\ref{tab:rev-same-recipe-matched}). The advantage persists when capacity sampling, distillation and refinement are matched and the exported sizes are nearly equal.
\par}

\end{revblockr}

\begin{table}[htbp]
\begin{revblockr}
\centering\footnotesize
\setlength{\tabcolsep}{4pt}
\revcaptionr{MatMamba under its own recipe and under the ESSH recipe at the same 3.7B tokens (370M, common evaluator). Own recipe is the local four-capacity baseline, with seeds 197 and 313 shown before and after the common 2060-step shared refinement. Its 83.3M export lies below the native pretraining range. ESSH recipe: seeds 197 and 313 after the same refinement. Seven-task mean as in Table~\ref{tab:rev-same-recipe-aux}.}{tab:rev-matmamba-own}
\begin{tabular*}{\linewidth}{@{\extracolsep{\fill}}lcccc@{}}
\toprule
Retained parameters (M) & 83.3 & 159.7 & 255.3 & 369.9 \\
\midrule
Own recipe, PPL before refinement & 2084.5--5107.2 & 20.98--21.01 & 19.43 & 19.15 \\
Own recipe, PPL after refinement & 57.72--62.44 & 20.79--20.83 & 19.23--19.24 & 19.03 \\
ESSH recipe, PPL after refinement & 33.80--33.93 & 22.47--22.62 & 19.14--19.24 & 16.69--16.74 \\
\midrule
Own recipe, seven-task mean & 30.6--31.1 & 37.2--37.4 & 38.3 & 38.3--38.4 \\
ESSH recipe, seven-task mean & 32.9--33.3 & 36.3--36.7 & 38.4--38.7 & 40.3--40.6 \\
\bottomrule
\end{tabular*}
\end{revblockr}
\end{table}

\begin{revblockr}
To compare at the four ESSH deployment sizes, we also choose Mamba-3 head counts to match the exported parameter counts, including the 92M tier (3, 13, 21 and 28 of 28 heads, Table~\ref{tab:rev-aligned-m3}). The training recipe and evaluation protocol remain the same. Across the repeated runs, ESSH is 2.0--2.2 PPL better at $T_1$ and 0.2--0.3 better at $T_4$, while Mamba-3 + SWA is 0.1--0.3 better at the larger tiers. ESSH's quality advantage is concentrated at compact capacities, where it also uses less recurrent state: 0.27 MB versus 1.90 MB at $T_1$. At full capacity, the corresponding state sizes are 1.09 and 17.70 MB. ESSH uses eight rotations, or 16 real state values, per residual dimension. The Mamba-3 configuration expands to twice the residual width with 128 state values per inner dimension. Both retain the same 22.02 MB attention cache. We also train this matched-size hybrid with the MatMamba recipe of Table~\ref{tab:rev-matmamba-own}, which samples the 1/8, 1/4, 1/2 and full capacities in equal shares, keeping the architecture, token budget, shared refinement and evaluator unchanged (seed 197). 
\end{revblockr}

\begin{table}[htbp]
\begin{revblockr}
\centering\footnotesize
\setlength{\tabcolsep}{3.5pt}
\revcaptionr{Head-nested Mamba-3 + SWA with head counts chosen so that every tier matches the ESSH parameter count (370M, 3.7B tokens, the ESSH recipe unless stated, common shared refinement and common evaluator). ESSH: seeds 197, 313 and 511. Head-nested hybrid: seeds 197 and 511. The ESSH-recipe PPL row gives seed means and the other rows give ranges. The MatMamba-recipe row trains the same hybrid with the four-capacity recipe of Table~\ref{tab:rev-matmamba-own} (seed 197). Seven-task mean as in Table~\ref{tab:rev-same-recipe-aux}, full-key accuracy on the 2K NIAH test (teacher forced) and recurrent state per sequence in MB.}{tab:rev-aligned-m3}
\resizebox{\linewidth}{!}{%
\begin{tabular}{@{}l cccc cccc@{}}
\toprule
& \multicolumn{4}{c}{\ours{}} & \multicolumn{4}{c}{Head-nested Mamba-3 + SWA, matched sizes} \\
\cmidrule(lr){2-5}\cmidrule(lr){6-9}
& $T_1$ & $T_4$ & $T_7$ & $T_{10}$ & $T_1$ & $T_4$ & $T_7$ & $T_{10}$ \\
\midrule
Parameters (M) & 92.1 & 185.3 & 274.3 & 370.9 & 92.3 & 186.1 & 277.1 & 372.3 \\
Heads kept & -- & -- & -- & -- & 3 & 13 & 21 & 28 \\
PPL, ESSH recipe & 25.34 & 19.19 & 17.43 & 16.27 & 27.43 & 19.44 & 17.31 & 16.06 \\
PPL, MatMamba recipe & -- & -- & -- & -- & 38.25 & 18.57 & 17.71 & 17.64 \\
Seven-task mean & 35.5--36.5 & 38.6--39.4 & 39.7--40.1 & 40.6--41.0 & 35.4--35.6 & 38.8--39.8 & 40.3--40.5 & 41.4--41.5 \\
Full-key accuracy & 0.06--0.29 & 0.01--0.26 & 0.06--0.26 & 0.15--0.33 & 0.00--0.02 & 0.14--0.22 & 0.28--0.29 & 0.37--0.41 \\
Recurrent state (MB) & 0.27 & 0.61 & 0.85 & 1.09 & 1.90 & 8.22 & 13.28 & 17.70 \\
\bottomrule
\end{tabular}}
\end{revblockr}
\end{table}

\begin{table}[htbp]
\color{revred}\centering\footnotesize
\setlength{\tabcolsep}{3.5pt}
\revcaption{Elastic families under the ESSH recipe at 370M and 3.7B pretraining tokens, after the common 2060-step shared refinement and with the common evaluator. Retained parameters (M) and PPL ranges are shown. Actual sizes differ across architectures at the same capacity index. Mamba-3+SWA uses the shared fractional head-reduction rule; matched-size comparisons are in Tables~\ref{tab:rev-same-recipe-matched} and~\ref{tab:rev-aligned-m3}. MatMamba's native-recipe control at the same budget is in Table~\ref{tab:rev-matmamba-own}.}{tab:rev-same-recipe}
\begin{tabular*}{\linewidth}{@{\extracolsep{\fill}}lrrrrrr@{}}
\toprule
& \multicolumn{2}{c}{\shortstack{ESSH\\seeds 197/313/511}} & \multicolumn{2}{c}{\shortstack{Mamba-3+SWA\\seeds 197/511}} & \multicolumn{2}{c}{\shortstack{MatMamba\\seeds 197/313}} \\
\cmidrule(lr){2-3}\cmidrule(lr){4-5}\cmidrule(lr){6-7}
Capacity & M & PPL $\downarrow$ & M & PPL $\downarrow$ & M & PPL $\downarrow$ \\
\midrule
$T_1$ & 92.1 & 25.32--25.38 & 105.6 & 23.68--23.84 & 83.3 & 33.80--33.93 \\
$T_4$ & 185.3 & 19.14--19.23 & 196.1 & 18.66--18.72 & 159.7 & 22.47--22.62 \\
$T_7$ & 274.3 & 17.38--17.46 & 280.4 & 17.08--17.12 & 255.3 & 19.14--19.24 \\
$T_{10}$ & 370.9 & 16.22--16.32 & 372.3 & 16.04--16.06 & 369.9 & 16.69--16.74 \\
\bottomrule
\end{tabular*}
\end{table}

\begin{table}[htbp]
\color{revred}\centering\footnotesize
\setlength{\tabcolsep}{4pt}
\revcaption{PPL at matched retained parameters with the ESSH capacity-sampling and distillation settings (370M, 3.7B tokens, common evaluator, after the common 2060-step shared refinement). Paired exports differ by at most 3.3\% of the ESSH parameter count. Mamba-3 uses the shared fractional head-reduction rule; the parameter-aligned variant that reaches 92M is reported separately in Table~\ref{tab:rev-aligned-m3}. The grid includes intermediate capacities. ESSH: seeds 197/313/511; Mamba-3 + SWA: seeds 197/511; MatMamba: seeds 197/313.}{tab:rev-same-recipe-matched}
\begin{tabular*}{\linewidth}{@{\extracolsep{\fill}}lrrrr@{}}
\toprule
& \multicolumn{2}{c}{\textbf{ESSH (ours)}} & \multicolumn{2}{c}{\textbf{Baseline, same recipe}} \\
\cmidrule(lr){2-3}\cmidrule(lr){4-5}
Baseline & M & PPL $\downarrow$ & M & PPL $\downarrow$ \\
\midrule
Head-nested Mamba-3 + SWA & 92.1 & 25.32--25.38 & \multicolumn{2}{c}{minimum under this map: 105.6M} \\
& 106.2 & 23.07--23.09 & 105.6 & 23.68--23.84 \\
& 118.9 & 21.65--21.70 & 119.9 & 21.87--22.01 \\
& 133.0 & 20.88--20.90 & 130.8 & 20.84--21.01 \\
& 185.3 & 19.14--19.23 & 189.0 & 18.85--18.89 \\
& 194.2 & 18.91--18.99 & 196.1 & 18.66--18.72 \\
& 274.3 & 17.38--17.46 & 276.6 & 17.12--17.17 \\
& 283.2 & 17.24--17.32 & 280.4 & 17.08--17.12 \\
& 370.9 & 16.22--16.32 & 372.3 & 16.04--16.06 \\
\midrule
MatMamba & 92.1 & 25.32--25.38 & 92.9 & 29.47--29.51 \\
& 154.7 & 20.00--20.02 & 159.7 & 22.47--22.62 \\
& 185.3 & 19.14--19.23 & 188.4 & 21.08--21.14 \\
& 257.8 & 17.61--17.69 & 255.3 & 19.14--19.24 \\
& 274.3 & 17.38--17.46 & 274.4 & 18.61--18.64 \\
& 370.9 & 16.22--16.32 & 369.9 & 16.69--16.74 \\
\bottomrule
\end{tabular*}
\end{table}

\begin{table}[htbp]
\color{revred}\centering\footnotesize
\setlength{\tabcolsep}{3.5pt}
\revcaption{Seven-task mean (\%), teacher-forced full-key accuracy on 2K NIAH, and recurrent-state memory (MB per sequence) for the 3.7B same-recipe families after common shared refinement. HellaSwag and ARC-Challenge use normalized accuracy; other tasks use accuracy. ESSH gives the three-seed range. Recurrent memory excludes the 22.02 MB attention cache of each hybrid. M3+SWA uses the shared fractional head-reduction rule, seed 197; its export sizes are in Table~\ref{tab:rev-same-recipe}. The parameter-aligned variant is in Table~\ref{tab:rev-aligned-m3}. MatMamba: range over seeds 197 and 313.}{tab:rev-same-recipe-aux}
\resizebox{\linewidth}{!}{%
\begin{tabular}{@{}l ccc ccc ccc@{}}
\toprule
& \multicolumn{3}{c}{Seven-task mean (\%) $\uparrow$} & \multicolumn{3}{c}{Full-key accuracy $\uparrow$} & \multicolumn{3}{c}{Recurrent state (MB) $\downarrow$} \\
\cmidrule(lr){2-4}\cmidrule(lr){5-7}\cmidrule(lr){8-10}
Capacity & ESSH & M3+SWA & MatMamba & ESSH & M3+SWA & MatMamba & ESSH & M3+SWA & MatMamba \\
\midrule
$T_1$ & 35.5--36.5 & 37.2 & 32.9--33.3 & 0.06--0.29 & 0.15 & 0.00 & 0.27 & 4.43 & 1.72 \\
$T_4$ & 38.6--39.4 & 39.5 & 36.3--36.7 & 0.01--0.26 & 0.22 & 0.00--0.01 & 0.61 & 10.12 & 8.21 \\
$T_7$ & 39.7--40.1 & 41.1 & 38.4--38.7 & 0.06--0.26 & 0.24 & 0.00--0.04 & 0.85 & 13.91 & 16.32 \\
$T_{10}$ & 40.6--41.0 & 41.7 & 40.3--40.6 & 0.15--0.33 & 0.25 & 0.02--0.09 & 1.09 & 17.70 & 26.05 \\
\bottomrule
\end{tabular}}

\end{table}

\subsection[Independent training seeds]{Independent training seeds}
\label{app:comp-family-seeds}

The seed comparison asks whether the intermediate-size PPL differences exceed the observed variation across independent runs of each family. Near 274M parameters, \revr{the highest of the three ESSH values is $15.96$}, still below the lowest MatFormer value of $16.98$. Full-capacity ESSH likewise remains below the MatFormer and MatMamba runs in this recipe-level comparison.

The baseline $T_1$ results are much more variable because this tier lies below the capacities sampled during baseline training.

\begin{table}[!htbp]
\centering\small
\setlength{\tabcolsep}{3.5pt}
\caption{PPL across independent training seeds at four deployment shapes. \revr{These are the 7.4B-token main-study recipes. All three are elastic families, and each seed is one shared run exported at the four shapes. The common-recipe 3.7B-token controls are in Appendix~\ref{app:rev-same-recipe}.} ESSH uses final shared checkpoints; MatFormer and MatMamba use end-of-pretraining checkpoints. Baseline $T_1$ lies below the smallest capacity sampled during their training and is not used for the common-range comparison.}
\label{tab:new-frontier-seeds}
\begin{tabular*}{\linewidth}{@{\extracolsep{\fill}}llrrrr@{}}
\toprule
Family & Seed & PPL $T_1\downarrow$ & PPL $T_4\downarrow$ & PPL $T_7\downarrow$ & PPL $T_{10}\downarrow$ \\
\midrule
ESSH & 197 & 23.61 & 17.68 & 15.91 & 14.71 \\
\revr{ESSH} & \revr{313} & \revr{23.73} & \revr{17.69} & \revr{15.90} & \revr{14.69} \\
ESSH & 511 & 23.77 & 17.76 & 15.96 & 14.73 \\
\midrule
MatFormer & 197 & 5844.15 & 17.82 & 16.98 & 16.80 \\
MatFormer & 313 & 106.56 & 18.08 & 17.30 & 17.14 \\
MatFormer & 511 & 112.01 & 18.05 & 17.25 & 17.10 \\
\midrule
MatMamba & 197 & 3029.37 & 19.10 & 17.38 & 16.98 \\
MatMamba & 313 & 13651.22 & 19.12 & 17.41 & 17.01 \\
\bottomrule
\end{tabular*}
\end{table}

\subsection[Shape-matched recovery and specialist references]{Shape-matched recovery and specialist references}
\label{app:comp-protocol}

Table~\ref{tab:comp} compares ways to obtain the exact $T_4$ and $T_7$ deployment shapes. The first block starts from 3.7B pretraining tokens and contrasts capacity-mixed training with truncating and recovering a full-capacity-only model. The second block compares the 7.4B shared family with specialists trained independently at each target size.

Within the 3.7B block, ESSH reaches full-key accuracy 0.258 at $T_4$ and 0.256 at $T_7$, compared with at most 0.146 and 0.116 for the recovered dense controls. It also has lower $T_4$ PPL, while 4120-step label-only recovery reaches similar $T_7$ PPL with twice the recovery steps. In the 7.4B block, the specialists have lower PPL and slightly higher task means, while ESSH retains higher full-key accuracy and supplies both sizes from shared training.

\begin{table}[htbp]
\centering\footnotesize
\setlength{\tabcolsep}{3pt}
\caption{Models at the exact ESSH $T_4/T_7$ shapes, grouped by pretraining budget. Additional tokens and steps specify the subsequent shared-training or recovery stage. Full-key accuracy uses the common 2K NIAH protocol.}
\label{tab:comp}
\resizebox{\linewidth}{!}{%
\begin{tabular}{@{}lrrrrrrrr@{}}
\toprule
 & \multicolumn{2}{c}{Additional training} & \multicolumn{3}{c}{$T_4$, 185.3M} & \multicolumn{3}{c}{$T_7$, 274.3M}\\
\cmidrule(lr){2-3}\cmidrule(lr){4-6}\cmidrule(lr){7-9}
Method & Tokens (B) & Steps & PPL $\downarrow$ & 7-task $\uparrow$ & Full key $\uparrow$ & PPL $\downarrow$ & 7-task $\uparrow$ & Full key $\uparrow$\\
\midrule
\multicolumn{9}{l}{3.7B pretraining tokens}\\
ESSH & 0.27 & 2060 & 19.15 & 0.394 & 0.258 & 17.38 & 0.401 & 0.256 \\
Dense control, truncated and recovered & 0.27 & 2060 & 23.75 & 0.376 & 0.131 & 18.63 & 0.397 & 0.116 \\
Dense control, label-only recovery & 0.54 & 4120 & 20.07 & 0.386 & 0.146 & 17.22 & 0.407 & 0.114 \\
\midrule
\multicolumn{9}{l}{7.4B pretraining tokens}\\
ESSH & 0.27 & 2060 & 17.68 & 0.405 & 0.370 & 15.91 & 0.420 & 0.366 \\
Independent specialist\refspec & 0 & -- & 16.25 & 0.412 & 0.326 & 15.10 & 0.424 & 0.306 \\
\bottomrule
\end{tabular}}
\end{table}

\subsection[Post-hoc elastification of the reference hybrid]{Post-hoc elastification of the reference hybrid}
\label{app:comp-posthoc-hybrid}

The post-hoc comparison recovers the Mamba-3 + SWA parent into models near the ESSH $T_4$ and $T_7$ sizes. {\color{revred}We compare PPL, seven-task accuracy and full-key retrieval under the common evaluation protocols.}

At the larger compact target, the two models have similar seven-task means: $0.425$ for the post-hoc model and $0.420$ for ESSH. Their teacher-forced full-key accuracies differ more substantially, $0.147$ and $0.366$, respectively. At the smaller target, ESSH is higher on both common metrics, with seven-task mean $0.405$ versus $0.397$ and full-key accuracy $0.370$ versus $0.034$. The compact models can therefore have similar aggregate task accuracy while differing markedly on planted-key retrieval.

{\color{revred}
PPL is also close at $T_7$, 15.91 for ESSH versus 15.92 after post-hoc recovery. At the more compressed $T_4$, ESSH retains lower PPL, 17.68 versus 19.64 (Table~\ref{tab:comp-posthoc-hybrid}).
\par}

\begin{table}[htbp]
\centering
\footnotesize
\setlength{\tabcolsep}{3pt}
\caption{Post-hoc masks targeting $T_4/T_7$ sizes. Recovered Mamba-3 models use 7.90B total tokens; ESSH uses 7.67B. Seven-task and full-key metrics use the same evaluation protocols. Actual parameter counts are shown. \revr{PPL uses one evaluator (the common 16-batch evaluator) for every compact row, and the parent rows have no entry.}}
\label{tab:comp-posthoc-hybrid}
\begin{tabular}{@{}p{0.36\linewidth}ccccc@{}}
\toprule
Configuration & \shortstack{tokens\\\mbox{(B)}} & \shortstack{params\\\mbox{(M)}} & \revr{PPL $\downarrow$} & \shortstack{7-task\\mean $\uparrow$} & \shortstack{Full-key\\acc. $\uparrow$} \\
\midrule
parent, dense Mamba-3 + SWA\refhyb & 7.4 & 377.1 & & 0.445 & 0.230 \\
$100\%$ mask, after elastification & 7.9 & 377.1 & & 0.438 & 0.248 \\
\midrule
mask targeting $T_4$ & 7.9 & 185.0 & \revr{19.64} & 0.397 & 0.034 \\
\ours{}, $T_4$ & 7.67 & 185.3 & \revr{17.68} & 0.405 & 0.370 \\
\midrule
mask targeting $T_7$ & 7.9 & 273.8 & \revr{15.92} & 0.425 & 0.147 \\
\ours{}, $T_7$ & 7.67 & 274.3 & \revr{15.91} & 0.420 & 0.366 \\
\bottomrule
\end{tabular}
\end{table}

\subsection[Preserving attention in post-hoc masks]{Preserving attention in post-hoc masks}
\label{app:comp-attention-preservation}

The attention-preserving control uses the same parent and 0.5B-token recovery budget while preserving all heads in both attention layers. The feed-forward units within those blocks remain subject to pruning. At the $T_7$ target, this changes full-key accuracy only from $0.147$ to $0.148$, compared with $0.366$ for the near-size ESSH export. The $T_4$ attention-preserving mask likewise remains far below ESSH.

Retaining the attention components alone is therefore insufficient to recover the retrieval behavior under this post-hoc recipe. The result is consistent with the broader comparison above: jointly training the reduced-capacity predictor produces different behavior from first training a dense model and recovering its compact masks afterwards.

\begin{table}[htbp]
\centering\small
\caption{Original and attention-preserving post-hoc masks near the $T_4/T_7$ sizes. Both recovery strategies use the same parent and 0.5B recovery tokens. Task and retrieval scores use the common protocols. \revr{PPL uses one evaluator (the common 16-batch evaluator) for every compact row, and the parent row has no entry.}}
\label{tab:new-attention-preservation}
\begin{tabular}{@{}lrrrrr@{}}
\toprule
Model / target & Params (M) & \revr{PPL $\downarrow$} & First token $\uparrow$ & Full key $\uparrow$ & 7-task $\uparrow$ \\
\midrule
Mamba-3 + SWA parent & 377.1 & & 0.241 & 0.230 & 0.445 \\
\midrule
Original mask, $T_4$ target & 185.0 & \revr{19.64} & 0.036 & 0.034 & 0.397 \\
Preserve attention, $T_4$ target & 185.3 & \revr{19.96} & 0.025 & 0.023 & 0.395 \\
\ours{}, $T_4$ & 185.3 & \revr{17.68} & 0.370 & 0.370 & 0.405 \\
\midrule
Original mask, $T_7$ target & 273.8 & \revr{15.92} & 0.152 & 0.147 & 0.425 \\
Preserve attention, $T_7$ target & 276.5 & \revr{15.91} & 0.154 & 0.148 & 0.426 \\
\ours{}, $T_7$ & 274.3 & \revr{15.91} & 0.366 & 0.366 & 0.420 \\
\bottomrule
\end{tabular}
\end{table}

\subsection[Post-hoc elastification of the dense control]{Post-hoc elastification of the dense control}
\label{app:comp-posthoc-twin}

This comparison fixes the ESSH architecture, the 3.7B-token parent pretraining budget, and the exported $T_4/T_7$ shapes, while changing how the compact models are trained. Post-hoc elastification adds 0.5B recovery tokens for 4.20B total tokens. Capacity-mixed training is followed by 0.27B shared refinement for 3.97B total tokens.

At $T_7$, post-hoc recovery and ESSH have nearly identical PPL, $17.35$ and $17.38$, but full-key accuracy is $0.107$ and $0.256$, respectively. At $T_4$, both have seven-task mean $0.394$, while full-key accuracy is $0.154$ after post-hoc recovery and $0.258$ for ESSH. Thus, under the same architecture and deployment shape, similar aggregate PPL or task accuracy can coexist with substantially different retrieval behavior.

These controls motivate training the retained configurations throughout pretraining rather than relying only on post-hoc recovery. Under the measured recipes, capacity-mixed training preserves more of the planted-key behavior despite using fewer total training tokens.

\begin{table}[htbp]
\centering
\footnotesize
\setlength{\tabcolsep}{3pt}
\caption{Same-architecture compact-model comparison. Post-hoc elastification uses 4.20B total tokens; capacity-mixed ESSH and the directly reduced dense control use 3.97B. Compact rows use the exact deployment shapes. Full-key is teacher-forced full-key accuracy.}
\label{tab:comp-posthoc-twin}
\begin{tabular}{@{}p{0.34\linewidth}ccccc@{}}
\toprule
Configuration & \shortstack{tokens\\\mbox{(B)}} & \shortstack{params\\\mbox{(M)}} & PPL $\downarrow$ & \shortstack{7-task\\mean $\uparrow$} & \shortstack{Full-key\\acc. $\uparrow$} \\
\midrule
parent: \ours{} (dense control, 3.7B) & 3.7 & 370.9 & 15.98 & 0.414 & 0.181 \\
$100\%$ mask, after elastification & 4.2 & 370.9 & 16.15 & 0.417 & 0.113 \\
\midrule
Post-hoc $T_4$ & 4.2 & 185.3 & 19.64 & 0.394 & 0.154 \\
\ours{} (3.7B), $T_4$ & 3.97 & 185.3 & 19.15 & 0.394 & 0.258 \\
\ours{} (dense control, reduced), $T_4$ & 3.97 & 185.3 & 23.75 & 0.376 & 0.131 \\
\midrule
Post-hoc $T_7$ & 4.2 & 274.3 & 17.35 & 0.404 & 0.107 \\
\ours{} (3.7B), $T_7$ & 3.97 & 274.3 & 17.38 & 0.401 & 0.256 \\
\ours{} (dense control, reduced), $T_7$ & 3.97 & 274.3 & 18.63 & 0.397 & 0.116 \\
\bottomrule
\end{tabular}
\end{table}

\subsection[Training cost within the ESSH architecture]{Training cost within the ESSH architecture}
\label{app:comp-amort}

The internal cost comparison asks how much accelerator time is required to obtain the same four ESSH deployment shapes, $T_1/T_4/T_7/T_{10}$, either from one shared training run or from four independent runs. Pretraining A100-hours are estimated by dividing 7.4B tokens by each configuration's rate over steps 20--35 of a single-A100 probe. The shared run includes capacity mixing and teacher forwards. Shared refinement uses its measured 2060-step elapsed time. These estimates exclude restarts, evaluation, and data-loading stalls.

The shared procedure requires an estimated 53.9 A100-hours under this accounting, compared with 50.6 hours for one independently trained full model and 147.4 hours for four independent models. Thus producing the four-size family reduces the four-model total from 147.4 to 53.9 A100-hours, a $63.4\%$ reduction, while requiring only about $6.5\%$ more than the single full-model run.

The shared family obtains this saving while staying \revr{within 0.01 of the independent models on seven-task mean at all four shapes}. Independent training gives lower compact-model PPL, with the largest difference at $T_1$ (Table~\ref{tab:sharing-control}). The cost comparison accounts for obtaining all four sizes.

\begin{table}[htbp]
\centering\small
\caption{Internal ESSH cost comparison for obtaining $T_1/T_4/T_7/T_{10}$ on one A100-80GB. Pretraining A100-hours are computed from 7.4B tokens and measured steady-state throughput; shared refinement uses the measured 2060-step wall-clock time. The four deployment sizes are 92.1, 185.3, 274.3, and 370.9M parameters.}
\label{tab:comp-amort}
\begin{tabular*}{\linewidth}{@{\extracolsep{\fill}}lrrr@{}}
\toprule
Stage & Tokens (B) & ktok/s & A100-h $\downarrow$ \\
\midrule
Shared pretraining & 7.4 & 41.9 & 49.1 \\
Shared refinement & 0.27 & -- & 4.81 \\
Independent $T_1$ & 7.4 & 90.2 & 22.8 \\
Independent $T_4$ & 7.4 & 62.1 & 33.1 \\
Independent $T_7$ & 7.4 & 50.2 & 40.9 \\
Independent $T_{10}$ & 7.4 & 40.66 & 50.6 \\
\midrule
\multicolumn{3}{l}{Shared training, total} & 53.9 \\
\multicolumn{3}{l}{Four independent models, total} & 147.4 \\
\bottomrule
\end{tabular*}
\end{table}

\subsection[Training cost of the external four-model comparison]{Training cost of the external four-model comparison}
\label{app:comp-external-cost}

The four independently trained Transformer++ models require an estimated 101.4 A100-hours to obtain the four measured deployment sizes. Shared ESSH requires 53.9 hours, including its shared refinement stage, reducing the total by 46.9\% (Table~\ref{tab:external-training-cost}). The quality comparison uses these same four sizes. Their task accuracies are listed in Table~\ref{tab:external-four-sizes}.

The independent total credits the different training rates of the four models: the smallest requires 14.55 hours, compared with 35.52 for the full-size reference. The saving therefore persists after accounting for the lower cost of each compact model. Shared training reuses updates to the retained tensors across deployment tiers, while the independent family optimizes four separate parameter sets.

The full-size and smallest Transformer++ rates come from completed single-A100 training segments. The two intermediate rates come from matched 100-step probes with the same frozen architecture, global token batch, optimizer, and execution settings. We discard the first 20 steps and use the median of seven ten-step blocks over steps 20--90. The median rates are 91.98 and 70.85 ktok/s for the $T_4$ and $T_7$ targets, respectively. Shared ESSH accounting is defined in Appendix~\ref{app:comp-amort}.

\begin{table}[!htbp]
\centering\small
\setlength{\tabcolsep}{3pt}
\caption{Training cost for the four sizes in Table~\ref{tab:external-four-sizes}. All rates use one A100-80GB. Independent costs are estimated as 7.4B tokens divided by each model's steady-state rate. Shared ESSH includes the teacher computation and measured shared refinement stage. The reported costs are estimates of training work at the measured rates.}
\label{tab:external-training-cost}
\begin{tabular*}{\linewidth}{@{\extracolsep{\fill}}lrr@{}}
\toprule
Transformer++ target / total & Rate (ktok/s) & A100-hours \\
\midrule
$T_{1}$ & 141.32 & 14.55 \\
$T_{4}$ & 91.98 & 22.35 \\
$T_{7}$ & 70.85 & 29.01 \\
$T_{10}$ & 57.87 & 35.52 \\
\midrule
Four independent Transformer++ runs & -- & 101.43 \\
One shared ESSH run & -- & \textbf{53.90} \\
\bottomrule
\end{tabular*}
\end{table}

\section[Systems: Algorithms and Kernels]{Systems: Algorithms and Kernels}
\label{app:systems_method}

This section gives the two execution paths used by ESSH: a chunked parallel scan for training and a tokenwise recurrent step for decoding. Both evaluate the same selective recurrence, including the channel-local value blend, zero-lag correction, and tier-specific tensor slices.

\subsection[Chunked forward scan]{Chunked forward scan}
\label{app:sys-fwd}

The chunk decomposition follows the structured state-space duality view of sequence computation~\citep{dao2024mamba2}, specialized to the retained selective spectral channels.

For channel $k$ and mode $i$, let $\psi_{k,i}(\tau)$ denote the damped-mode response in Equation~\ref{eq:normal_kernel}. The single-mode causal interaction is
\begin{equation}
G_{k,i}(t,s)
=
\mathbf1[s\leq t]\,
\eta^{\mathrm r}_{k,i}(t)
\eta^{\mathrm w}_{k,i}(s)
\mathrm e^{-(q_{k,i}(t)-q_{k,i}(s))}
\psi_{k,i}(t-s).
\label{eq:chunk_factorization}
\end{equation}
Entries above the causal diagonal are zero and need not be evaluated.

Consider a chunk $[b,e]$. Let
\[
G_k[t,s]=\sum_{i=1}^{m}G_{k,i}(t,s),
\]
and stack the channel values as
\[
V_k=[v_k(b)^\top;\ldots;v_k(e)^\top].
\]
The contribution from the state entering the chunk is
\[
B_k[t,:]
=
\sum_{i=1}^{m}
\eta^{\mathrm r}_{k,i}(t)
\kappa_{k,i}^{\top}
\Pi_{k,i}(t,b-1)
H_{k,i}(b-1),
\qquad b\leq t\leq e.
\]
Thus
\[
G_k\in\mathbb R^{(e-b+1)\times(e-b+1)},
\qquad
V_k,B_k,Z_k^{\mathrm{core}}
\in\mathbb R^{(e-b+1)\times P}.
\]

Unrolling from the entering state gives
\begin{equation}
\begin{aligned}
H_{k,i}(t)
={}&
\Pi_{k,i}(t,b-1)H_{k,i}(b-1)\\
&+
\sum_{s=b}^{t}
\Pi_{k,i}(t,s)
\eta^{\mathrm w}_{k,i}(s)e_1v_k(s)^\top,\\
Z_k^{\mathrm{core}}
={}&
\left(\sum_iG_{k,i}\right)V_k+B_k
=
G_kV_k+B_k.
\end{aligned}
\label{eq:chunk_execution_main}
\end{equation}
The state passed to the next chunk is
\[
H_{k,i}(e)
=
\Pi_{k,i}(e,b-1)H_{k,i}(b-1)
+
\sum_{s=b}^{e}
\Pi_{k,i}(e,s)
\eta^{\mathrm w}_{k,i}(s)e_1v_k(s)^\top.
\]
The local injection at $s=b$ does not include $A_{k,i}(b)$, whereas the entering state does. Equation~\ref{eq:chunk_execution_main} therefore accounts for the boundary without double counting. Repeating the update over chunks reproduces the tokenwise recurrence, including a final partial chunk. Equation~\ref{eq:zero_lag} is then applied once at each valid position.

\paragraph{Interaction across a chunk boundary.}
Fix the realized controls. For a source $s<b$ and target $t\geq b$,
\begin{equation}
G_{k,i}(t,s)
=
\left[
\eta^{\mathrm r}_{k,i}(t)
\kappa_{k,i}^{\top}
\Pi_{k,i}(t,b-1)
\right]
\left[
\Pi_{k,i}(b-1,s)
\eta^{\mathrm w}_{k,i}(s)e_1
\right].
\label{eq:boundary_interaction_factors}
\end{equation}
Each mode therefore contributes a token-interaction factorization of inner dimension two. Concatenating the factors over $m$ modes gives rank at most $2m$ for the scalar token-interaction block crossing the boundary. The zero-lag correction contributes no cross-boundary term. With $P$ value coordinates, the corresponding recurrent state contains $2mP$ entries per channel. This rank statement applies to the cross-boundary interaction block, not to the complete causal matrix.

\paragraph{Shared values across modes.}
All modes in channel $k$ act on the same $V_k$, so
\[
\sum_iG_{k,i}V_k
=
\left(\sum_iG_{k,i}\right)V_k.
\]
The mode tiles can therefore be accumulated before the value multiplication. For a full $Q$-token chunk, the recurrent core requires
\[
O(K_TmQ^2)
+
O(K_TQ^2P)
+
O(K_TQmP)
\]
work for tile construction, shared-value multiplication, and boundary propagation. Across a length-$L$ sequence this becomes
\[
O\!\left(
LK_T[Q(m+P)+mP]
\right),
\]
up to padding of the final chunk. Projection costs are accounted for separately in Table~\ref{tab:complexity}.

Algorithms~\ref{alg:ms-ssd-forward} and~\ref{alg:ms-ssd-backward} use
\[
\Pi_{k,i}(t,s)
=
A_{k,i}(t)\cdots A_{k,i}(s+1),
\qquad
\Pi_{k,i}(t,t)=I_2.
\]
$H^{\mathrm{bd}}$ denotes the state entering a chunk, and $\mathcal H$ stores these boundary states for backward evaluation. Within chunk $j$, the local accumulated clock
\[
q^{(j)}_{k,i}(t)
=
\sum_{s=b_j}^{t}\Delta_{k,i}(s)
\]
produces the same within-chunk differences
$q^{(j)}(t)-q^{(j)}(s)=q(t)-q(s)$.
Batch indices are omitted.

\begin{algorithm}[H]
\small
\DontPrintSemicolon
\SetAlgoVlined
\caption{Chunked forward evaluation}
\label{alg:ms-ssd-forward}
\KwIn{$v_k(t),\eta^{\mathrm w}_{k,i}(t),\eta^{\mathrm r}_{k,i}(t),\Delta_{k,i}(t)$, $\rho,\theta,\kappa,a$, $L,Q,K_T$, flag $\mathrm{save}$}
\KwOut{corrected channel outputs $z_k(1{:}L)$ and boundary cache $\mathcal H$}

$\mathcal H\gets\varnothing$\;

\ForEach{$k=1,\ldots,K_T$ in parallel}{
  $H^{\mathrm{bd}}_{k,i}\gets0$ for $i=1,\ldots,m$\;

  \For{$j=1,\ldots,\lceil L/Q\rceil$}{
    $b\gets(j-1)Q+1$\;
    $e\gets\min(jQ,L)$\;
    $\mathcal C_j\gets\{b,\ldots,e\}$\;

    \If{$\mathrm{save}$}{
      $\mathcal H_{k,:,j}\gets H^{\mathrm{bd}}_{k,:}$\;
    }

    $V_k\gets[v_k(b)^\top;\ldots;v_k(e)^\top]$\;

    $q^{(j)}_{k,i}(t)
    \gets
    \sum_{s=b}^{t}\Delta_{k,i}(s)$
    for each $(i,t)$ with $t\in\mathcal C_j$\;

    $G_k[t-b+1,s-b+1]
    \gets
    \sum_{i=1}^{m}G_{k,i}(t,s)$
    using $q^{(j)}$ for $t,s\in\mathcal C_j$\;

    \If{$e-b+1<Q$}{
      $(G_k,V_k)\gets\operatorname{ZeroPad}(G_k,V_k;Q)$\;
    }

    $Z_k^{\mathrm{local}}\gets G_kV_k$\;

    \ForEach{$t\in\mathcal C_j$ in parallel}{
      $z_k^{\mathrm{core}}(t)
      \gets
      (Z_k^{\mathrm{local}}[t-b+1,:])^\top$\;

      $z_k^{\mathrm{core}}(t)
      \gets
      z_k^{\mathrm{core}}(t)
      +
      \sum_i
      \eta^{\mathrm r}_{k,i}(t)
      (H^{\mathrm{bd}}_{k,i})^\top
      \Pi_{k,i}(t,b-1)^\top
      \kappa_{k,i}$\;

      $z_k(t)
      \gets
      z_k^{\mathrm{core}}(t)
      -
      \sum_i
      a_{k,i}
      \eta^{\mathrm r}_{k,i}(t)
      \eta^{\mathrm w}_{k,i}(t)
      \kappa^{\mathrm c}_{k,i}
      v_k(t)$\;
    }

    \ForEach{$i=1,\ldots,m$ in parallel}{
      $H^{\mathrm{bd}}_{k,i}
      \gets
      \Pi_{k,i}(e,b-1)H^{\mathrm{bd}}_{k,i}$\;

      $H^{\mathrm{bd}}_{k,i}
      \gets
      H^{\mathrm{bd}}_{k,i}
      +
      \sum_{s=b}^{e}
      \Pi_{k,i}(e,s)
      \eta^{\mathrm w}_{k,i}(s)
      e_1v_k(s)^\top$\;
    }
  }
}
\Return{$z,\mathcal H$}\;
\end{algorithm}

\subsection[Backward pass with boundary adjoints]{Backward pass with boundary adjoints}
\label{app:sys-bwd}

The output projection gives the upstream channel gradient
\[
g_k(t)
=
K_T^{-1/2}
(W_k^{\mathrm{out}})^\top
\nabla_{\hat y^{(T)}(t)}\mathcal L.
\]
Fix one channel and suppress its index. For one mode, define the state adjoint backward from the final position:
\begin{equation}
E_L
=
\eta^{\mathrm r}_L\kappa g_L^\top,
\qquad
E_t
=
\eta^{\mathrm r}_t\kappa g_t^\top
+
A_{t+1}^\top E_{t+1},
\quad t<L.
\label{eq:adjoint}
\end{equation}

For one mode,
\[
\mathrm dH_t
=
A_t\,\mathrm dH_{t-1}
+
(\mathrm dA_t)H_{t-1}
+
e_1v_t^\top\,\mathrm d\eta^{\mathrm w}_t
+
\eta^{\mathrm w}_te_1(\mathrm dv_t)^\top.
\]
The transition differential is
\begin{equation}
\mathrm dA_t
=
\frac{A_t}{\rho}\,\mathrm d\rho
-
A_t\,\mathrm d\Delta_t
+
\rho\mathrm e^{-\Delta_t}
R'(\theta)\,\mathrm d\theta.
\label{eq:transition_differential}
\end{equation}

Using
\[
\langle E_t,A_t\,\mathrm dH_{t-1}\rangle_F
=
\langle A_t^\top E_t,\mathrm dH_{t-1}\rangle_F,
\]
the core derivatives are
\begin{equation}
\begin{aligned}
\nabla_{v_t}\mathcal L
&=
\sum_i
\eta^{\mathrm w}_{i,t}E_{i,t}^\top e_1,
&
\partial_{\eta^{\mathrm w}_t}\mathcal L
&=
e_1^\top E_tv_t,
\\
\partial_{\eta^{\mathrm r}_t}\mathcal L
&=
g_t^\top H_t^\top\kappa,
&
\partial_{\Delta_t}\mathcal L
&=
-\langle E_t,A_tH_{t-1}\rangle_F,
\\
\partial_{\rho}\mathcal L
&=
\sum_t
\langle E_t,A_tH_{t-1}\rangle_F/\rho,
&
\partial_{\theta}\mathcal L
&=
\sum_t
\rho\mathrm e^{-\Delta_t}
\langle E_t,R'(\theta)H_{t-1}\rangle_F,
\\
\nabla_\kappa\mathcal L
&=
\sum_t
\eta^{\mathrm r}_tH_tg_t.
\end{aligned}
\label{eq:core_gradients}
\end{equation}
Here
\[
\langle X,Y\rangle_F
=
\operatorname{tr}(X^\top Y).
\]
Because
\[
\kappa
=
(\kappa^{\mathrm c},-\kappa^{\mathrm s})^\top,
\]
the stored readout coefficients satisfy
\[
\frac{\partial\mathcal L}{\partial\kappa^{\mathrm c}}
=
[\nabla_\kappa\mathcal L]_1,
\qquad
\frac{\partial\mathcal L}{\partial\kappa^{\mathrm s}}
=
-[\nabla_\kappa\mathcal L]_2,
\]
before adding the zero-lag correction to the $\kappa^{\mathrm c}$ derivative.

The zero-lag correction is local and does not change the state adjoint. For one mode, define
\[
\chi_t=g_t^\top v_t,
\qquad
\ell_t^{\mathrm{corr}}
=
-a\eta_t^{\mathrm r}
\eta_t^{\mathrm w}
\kappa^{\mathrm c}\chi_t.
\]
Its nonzero derivatives are
\[
\begin{aligned}
\nabla_{v_t}\ell_t^{\mathrm{corr}}
&=
-a\eta_t^{\mathrm r}\eta_t^{\mathrm w}
\kappa^{\mathrm c}g_t,
&
\partial_a\ell_t^{\mathrm{corr}}
&=
-\eta_t^{\mathrm r}\eta_t^{\mathrm w}
\kappa^{\mathrm c}\chi_t,
\\
\partial_{\eta_t^{\mathrm w}}\ell_t^{\mathrm{corr}}
&=
-a\eta_t^{\mathrm r}\kappa^{\mathrm c}\chi_t,
&
\partial_{\eta_t^{\mathrm r}}\ell_t^{\mathrm{corr}}
&=
-a\eta_t^{\mathrm w}\kappa^{\mathrm c}\chi_t,
\\
\partial_{\kappa^{\mathrm c}}\ell_t^{\mathrm{corr}}
&=
-a\eta_t^{\mathrm r}\eta_t^{\mathrm w}\chi_t.
\end{aligned}
\]
The correction has no direct derivative with respect to
$\Delta$, $\rho$, $\theta$, or $\kappa^{\mathrm s}$.

For the raw decay parameter,
\[
\rho
=
\exp[-\mathrm{softplus}(\tilde\rho)],
\qquad
\frac{\partial\rho}{\partial\tilde\rho}
=
-\rho\,\mathrm{sigmoid}(\tilde\rho).
\]
For the affine logits
$s^\Delta_t,s^{\mathrm w}_t,s^{\mathrm r}_t$
from Appendix~\ref{app:mixer-proj},
\begin{equation}
\begin{aligned}
\partial_{\tilde\rho}\mathcal L
&=
-\operatorname{sigmoid}(\tilde\rho)
\sum_t
\langle E_t,A_tH_{t-1}\rangle_F,
\\
\partial_{s^\Delta_t}\mathcal L
&=
\operatorname{sigmoid}(s^\Delta_t)
\partial_{\Delta_t}\mathcal L,
\\
\partial_{s^{\mathrm w}_t}\mathcal L
&=
\eta^{\mathrm w}_t
(1-\eta^{\mathrm w}_t/2)
\partial_{\eta^{\mathrm w}_t}\mathcal L,
\\
\partial_{s^{\mathrm r}_t}\mathcal L
&=
\eta^{\mathrm r}_t
(1-\eta^{\mathrm r}_t/2)
\partial_{\eta^{\mathrm r}_t}\mathcal L.
\end{aligned}
\label{eq:raw_parameter_chain_rules}
\end{equation}
The gate derivatives in Equation~\ref{eq:raw_parameter_chain_rules} include the local zero-lag terms above.

\paragraph{Stable factorization of accumulated decay.}
For one mode,
\begin{equation}
-\log
\|A_t\cdots A_{s+1}\|_2
=
-(t-s)\log\rho
+
q(t)-q(s).
\label{eq:accumulated_decay}
\end{equation}
For positions $t$ in a chunk beginning at $b$, define the zero-based within-chunk coordinate $t-b$ and
\[
\ell(t)=(t-b)\gamma+\sum_{u=b}^{t}\Delta(u).
\]
For any center $\ell_\star$,
\[
\rho^{t-s}
\mathrm e^{-(q(t)-q(s))}
=
\mathrm e^{-(\ell(t)-\ell_\star)}
\mathrm e^{\ell(s)-\ell_\star}.
\]
The rotation factor separates similarly:
\[
R(\theta(t-s))
=
R(\theta(t-b))
R(\theta(s-b))^\top.
\]
These identities allow endpoint-factorized evaluation without changing the mode response.

Centering reduces the dynamic range of the endpoint factors. The factorized branch is used only when its centered factors remain inside the configured fp32 range. Otherwise the kernel evaluates the causal within-chunk interaction sums directly using a masked $Q\times Q$ block. With largest and smallest positive normal fp32 values $F_{\max}$ and $F_{\min}$, define
\[
E_{\mathrm{safe}}
=
\min\{
\log(F_{\max}\epsilon),
-\log(F_{\min}/\epsilon)
\},
\qquad
\epsilon=2\times10^{-3}.
\]
The factorized branch admits a mode when
\[
(Q-1)\gamma_{k,i}
+
\max_j
\sum_{t\in\mathcal C_j}
\Delta_{k,i}(t)
\leq
2E_{\mathrm{safe}}.
\]
Intrinsic decay is centered at $(Q-1)/2$, and each chunk's accumulated clock is centered at half its final sum. The maximum is taken across chunks and batch sequences, giving one admission decision per channel and mode. Modes outside this range use the direct branch.

Algorithm~\ref{alg:ms-ssd-backward} organizes the reverse pass through the chunk boundaries. Its direct branch is written as a reverse recurrence to specify the gradients. The kernel evaluates the equivalent masked within-chunk sums. Let $\Lambda_{k,i,j}$ denote the gradient from chunks after $j$ with respect to the mode state at $e_j$. Each chunk defines
\[
\mathcal P_{k,i,j}
=
\Pi_{k,i}(e_j,b_j-1)
\]
and a local contribution
\[
\mathcal B_{k,i,j}
=
\sum_{t=b_j}^{e_j}
\eta^{\mathrm r}_{k,i}(t)
\Pi_{k,i}(t,b_j-1)^\top
\kappa_{k,i}g_k(t)^\top.
\]
The boundary adjoints then satisfy
\[
\Lambda_{k,i,j-1}
=
\mathcal B_{k,i,j}
+
\mathcal P_{k,i,j}^\top
\Lambda_{k,i,j}.
\]

\begin{algorithm}[H]
\small
\DontPrintSemicolon
\SetAlgoVlined
\caption{Backward evaluation with boundary adjoints}
\label{alg:ms-ssd-backward}
\KwIn{$g_k(t)=\partial\mathcal L/\partial z_k(t)$, forward inputs and parameters, $\mathcal H$, chunks $\mathcal C_j=[b_j,e_j]$, $j=1,\ldots,J$}
\KwOut{gradients of values, gate/clock logits, and modal parameters}

\ForEach{$(k,i,j)$ in parallel}{
  $\mathcal P_{k,i,j}\gets\Pi_{k,i}(e_j,b_j-1)$\;

  $\mathcal B_{k,i,j}
  \gets
  \sum_{t=b_j}^{e_j}
  \eta^{\mathrm r}_{k,i}(t)
  \Pi_{k,i}(t,b_j-1)^\top
  \kappa_{k,i}
  g_k(t)^\top$\;
}

$\Lambda_{k,i,J}\gets0$ for all $(k,i)$\;

\For{$j=J,J-1,\ldots,1$}{
  $\Lambda_{k,i,j-1}
  \gets
  \mathcal B_{k,i,j}
  +
  \mathcal P_{k,i,j}^\top
  \Lambda_{k,i,j}$
  for all $(k,i)$\;
}

\ForEach{$(k,i,j)$ in parallel}{
  $H_{k,i}(b_j-1{:}e_j)
  \gets
  \operatorname{Recompute}
  (\mathcal H_{k,i,j},v,\eta^{\mathrm w},A)$\;

  \eIf{
    $\operatorname{Admit}
    (\rho_{k,i},\{\Delta_{k,i}[\mathcal C_c]\}_{c=1}^{J},\mathrm{dtype})$
  }{
    $\mathcal D_{k,i,j}
    \gets
    \operatorname{FactorizedGrad}
    (H,\Lambda_{k,i,j},g;\mathcal C_j)$\;
  }{
    $E\gets\Lambda_{k,i,j}$\;
    $\mathcal D_{k,i,j}\gets0$\;

    \For{$t=e_j,e_j-1,\ldots,b_j$}{
      $E
      \gets
      E
      +
      \eta^{\mathrm r}_{k,i}(t)
      \kappa_{k,i}
      g_k(t)^\top$\;

      $\mathcal D_{k,i,j}
      \gets
      \mathcal D_{k,i,j}
      +
      \operatorname{CoreGrad}
      (t,H_t,H_{t-1},E;
      g,\eta^{\mathrm w},\eta^{\mathrm r},A,\kappa)$\;

      $E\gets A_{k,i}(t)^\top E$\;
    }
  }

  $\mathcal D_{k,i,j}
  \gets
  \mathcal D_{k,i,j}
  +
  \operatorname{CorrectionGrad}
  (g,v,\eta^{\mathrm w},\eta^{\mathrm r},
  a,\kappa^{\mathrm c};\mathcal C_j)$\;
}

$\nabla\mathcal L
\gets
\operatorname{ReduceShared}
(\{\mathcal D_{k,i,j}\})$\;

$\nabla\mathcal L
\gets
\operatorname{ChainRules}
(\nabla\mathcal L;
\mathrm{sigmoid},\mathrm{softplus},\rho,\kappa)$\;

\Return{$\nabla\mathcal L$}\;
\end{algorithm}

\subsection[Fused single-token decoder]{Fused single-token decoder}
\label{app:sys-decode}

The tokenwise decoder uses the same recurrence as the chunked parallel scan, together with the channel-local value blend and zero-lag correction. Values and the write, read, and clock logits are produced by one concatenated projection, followed by the recurrent update and output projection. The identity
\[
\exp[-\mathrm{softplus}(x)]
=
\mathrm{sigmoid}(-x)
\]
gives the clock contribution to the transition directly.

The rotation matrices $R(\theta_{k,i})$ and other fixed modal constants are cached when a tier is loaded. The timed fused decoder includes the trained value blend and zero-lag correction.

Algorithm~\ref{alg:decode-step} receives the recurrent states and window KV caches for positions $1,\ldots,t-1$ and processes token $t$. $\operatorname{Project}$ computes the retained channel values and gate/clock logits, $\operatorname{Blend}$ is Equation~\ref{eq:value_blend}, and $\operatorname{Attn}$ denotes causal multi-head attention over the retained window.

\begin{algorithm}[H]
\small
\DontPrintSemicolon
\SetAlgoVlined
\caption{One decoding step of a standalone export}
\label{alg:decode-step}
\KwIn{token $\mathrm{id}_t$, $\Theta_T$, states $H$, window KV caches $(\mathcal K,\mathcal V)$, attention-layer set $\mathcal A$, window $w$, position $t$}
\KwOut{next-token logits and updated $H,\mathcal K,\mathcal V,t$}

$x\gets E_{\mathrm{tok}}[\mathrm{id}_t,:]^\top$\;

\For{$\ell=1,\ldots,N_{\mathrm{layers}}$}{
  $u\gets\operatorname{RMSNorm}_{\ell,1}(x)$\;

  \eIf{$\ell\notin\mathcal A$}{
    $(v^0,s^{\mathrm w},s^{\mathrm r},s^\Delta)
    \gets
    \operatorname{Project}_{\ell,T}(u)$\;

    \For{$k=1,\ldots,K_T$}{
      $v_k\gets\operatorname{Blend}_{\beta_k}(v_k^0)$\;
      $z_k\gets0$\;

      \For{$i=1,\ldots,m$}{
        $(\eta^{\mathrm w},\eta^{\mathrm r})
        \gets
        (
        2\,\mathrm{sigmoid}(s^{\mathrm w}_{k,i}),
        2\,\mathrm{sigmoid}(s^{\mathrm r}_{k,i})
        )$\;

        $\Delta
        \gets
        \mathrm{softplus}(s^\Delta_{k,i})$\;

        $A
        \gets
        \rho_{k,i}
        \mathrm e^{-\Delta}
        R(\theta_{k,i})$\;

        $H_{\ell,k,i}
        \gets
        AH_{\ell,k,i}
        +
        \eta^{\mathrm w}e_1v_k^\top$\;

        $z_k
        \gets
        z_k
        +
        \eta^{\mathrm r}
        \left(
        H_{\ell,k,i}^\top\kappa_{k,i}
        -
        a_{k,i}
        \eta^{\mathrm w}
        \kappa^{\mathrm c}_{k,i}
        v_k
        \right)$\;
      }
    }

    $\hat y
    \gets
    K_T^{-1/2}
    \sum_kW_k^{\mathrm{out}}z_k$\;
  }{
    $(\mathbf q,\mathbf k,\mathbf v)
    \gets
    \operatorname{QKV}_{\ell}(u)$\;

    $(\mathbf q,\mathbf k)
    \gets
    \operatorname{RoPE}_t(\mathbf q,\mathbf k)$\;

    $(\mathcal K_\ell,\mathcal V_\ell)
    \gets
    \operatorname{Append}
    (\mathcal K_\ell,\mathcal V_\ell;
    \mathbf k,\mathbf v)$\;

    \If{$|\mathcal K_\ell|>w$}{
      $(\mathcal K_\ell,\mathcal V_\ell)
      \gets
      (\mathcal K_\ell[-w{:}],\mathcal V_\ell[-w{:}])$\;
    }

    $\hat y
    \gets
    \operatorname{OutProj}_{\ell}
    \left(
    \operatorname{Attn}_{\ell}
    (\mathbf q,\mathcal K_\ell,\mathcal V_\ell)
    \right)$\;
  }

  $x\gets x+\hat y$\;
  $u\gets\operatorname{RMSNorm}_{\ell,2}(x)$\;
  $n\gets n_{\mathrm{ff},T}$\;

  $z^{\mathrm{ff}}
  \gets
  \mathrm{SiLU}
  (W^{\mathrm{gate}}_{\ell,1:n,:}u)
  \odot
  (W^{\mathrm{up}}_{\ell,1:n,:}u)$\;

  $x
  \gets
  x+
  W^{\mathrm{down}}_{\ell,:,1:n}
  z^{\mathrm{ff}}$\;
}

$\mathrm{logits}
\gets
E_{\mathrm{tok}}
\operatorname{RMSNorm}_{\mathrm{final}}(x)$\;

$t\gets t+1$\;

\Return{$\mathrm{logits},H,\mathcal K,\mathcal V,t$}\;
\end{algorithm}

\subsection[Device-adaptive autotuning]{Device-adaptive autotuning}
\label{app:sys-autotune}

Kernel configurations are selected separately for each device and shape class. Candidate warp counts, pipeline stages and band-kernel variants are timed at the fixed chunk size under the same workload and precision, and the selected configuration is checked against the reference implementation for forward and gradient agreement.

\subsection[Numerical correctness checks]{Numerical correctness checks}
\label{app:sys-gates}

Independent fp64 CPU implementations compare direct recurrence, explicit interaction matrices, and chunked execution under varying clocks, signed sine coefficients, partial chunks, nonzero entering states, and zero-lag corrections. Analytic gradients are compared with centered finite differences. Separate checks verify standalone slicing against the same-tier shared model and compare the fused GPU decoder with both the chunked forward path and a reference tokenwise stepper.

\begin{table}[htbp]
\centering
\footnotesize
\setlength{\tabcolsep}{3pt}
\caption{Numerical verification of the recurrent and export implementations. CPU entries report fp64 maximum absolute error with random seed 2333; centered finite differences use step $10^{-6}$. GPU entries report relative error on the recorded reduced test shapes.}
\label{tab:verification-new}
\label{tab:verification-added}
\begin{tabular*}{\linewidth}{@{\extracolsep{\fill}}p{.58\linewidth}lr@{}}
\toprule
Check & Error metric & Maximum \\
\midrule
CPU recurrence, chunk boundaries and zero-lag identities
& absolute & $3.55\times10^{-15}$ \\
CPU recurrence gradients versus centered finite differences
& absolute & $2.28\times10^{-9}$ \\
CPU toy export versus same-tier shared model
& absolute & $0$ \\
\midrule
GPU fused decoder versus chunked forward
& relative & $1.37\times10^{-6}$ \\
GPU fused decoder versus reference stepper
& relative & $2.11\times10^{-5}$ \\
\bottomrule
\end{tabular*}
\end{table}

\section[Systems Evaluation]{Systems Evaluation}
\label{app:systems_results}

The systems study reports whole-model training, prefill, decoding, and memory. \revr{Decoding and prefill follow the common inference protocol in Appendix~\ref{app:setup-hw}.} Shape benchmarks measure the executable architectures at the stated parameter counts, precision, and execution settings. Table~\ref{tab:new-quality-latency-1p5b}(b) places trained-checkpoint quality beside the corresponding tier-size benchmarks. The A100 study is complemented by H200 and B300 inference measurements.

\subsection[Training throughput per device]{Training throughput per device}
\label{app:sys-train}

Table~\ref{tab:sys-train} reports full-capacity ESSH throughput at context 2048 on one A100, H200 and B300, using steps 20--35 of 40-step probes and the listed micro-batch and accumulation settings\revr{, with block compilation}.

The family-cost measurements in Appendices~\ref{app:comp-amort} and~\ref{app:comp-external-cost} include capacity mixing and teacher computation. Their pretraining costs are estimated from the rate under that workload, while shared refinement uses measured elapsed time.

\begin{table}[htbp]
\centering\small
\caption{Full-capacity ESSH training throughput on one GPU at context 2048\revr{, with block compilation}. Params include tied embeddings. $B/A$ denote micro-batch size/accumulation, and rates use steps 20--35 of 40-step probes. ESSH remains in full-capacity warm-up, before capacity mixing and teacher forwards.}
\label{tab:sys-train}
\label{tab:sys-train-paired}
\begin{tabular*}{\linewidth}{@{\extracolsep{\fill}}llrrrr@{}}
\toprule
Device & Model & Params (M) & $B$ & $A$ & ktok/s $\uparrow$ \\
\midrule
A100 & ESSH 370M & 370.9 & 8 & 16 & 41.7 \\
H200 & ESSH 370M & 370.9 & 8 & 16 & 75.5 \\
B300 & ESSH 370M & 370.9 & 8 & 16 & \revr{79.3} \\
\midrule
A100 & ESSH 1.53B & 1531.1 & 4 & 8 & 13.5 \\
H200 & ESSH 1.53B & 1531.1 & 4 & 8 & \revr{21.7} \\
B300 & ESSH 1.53B & 1531.1 & 4 & 8 & \revr{26.2} \\
\bottomrule
\end{tabular*}
\end{table}

\subsection[Prefill]{Prefill}
\label{app:sys-prefill}

The A100 prefill curves separate the cost of adding tokens from the cost of increasing model capacity. Within each tier, throughput rises slightly as context grows from 2K to 128K. Full ESSH increases from \revr{34.8 to 40.8} ktok/s. Across tiers, smaller retained shapes are faster. The recurrence uses fixed-size mode states, and attention uses a fixed window. Their work grows linearly with sequence length.

The advantage of a compact tier therefore persists for long inputs: it removes spectral-channel and FFN work at every position. Peak allocation also contains the full vocabulary-logit tensor returned by this benchmark, so it follows a different curve from throughput and recurrent state. The 2K latency measures the prefill component of time to first token. At matched full-model size, Mamba-2 has the highest measured prefill throughput. ESSH overtakes Transformer++ as the input grows: their 128K rates are \revr{40.8 and 12.1} ktok/s, respectively. This crossing follows the different growth of recurrent/windowed processing and full attention. It is distinct from the recurrent decode comparison.

\begin{table}[htbp]
\begin{revblockr}
\centering
\small
\setlength{\tabcolsep}{4pt}
\caption{A100 batch-one full-logit prefill with torch.compile for every model (the latencies of Table~\ref{tab:rev-sym-prefill}). Throughput is ktok/s and latency ms. Peak allocations are reported in Table~\ref{tab:sys-memory}. Rows are session medians, each session with three warm-ups and five timed calls. Configurations are in Table~\ref{tab:sys-matched-config}. Bold compares the parameter-matched full models.}
\label{tab:sys-prefill}
\begin{tabular}{@{}lrccccc@{}}
\toprule
 &  & \multicolumn{4}{c}{prefill ktok/s $\uparrow$} &  \\
\cmidrule(lr){3-6}
Model & Params (B) & 2K & 8K & 32K & 128K & Prefill ms (2K) $\downarrow$ \\
\midrule
\ours{}, $T_1$ & 0.474 & 57.0 & 59.2 & 61.1 & 61.5 & 35.9 \\
\ours{}, $T_4$ & 0.828 & 45.9 & 50.2 & 52.1 & 52.4 & 44.7 \\
\ours{}, $T_7$ & 1.168 & 39.8 & 44.7 & 45.7 & 46.0 & 51.5 \\
\midrule
\ours{}, $T_{10}$ & 1.531 & 34.8 & 39.4 & 40.6 & 40.8 & 58.8 \\
Mamba-2 + FFN & 1.526 & \textbf{55.3} & \textbf{65.7} & \textbf{66.7} & \textbf{66.4} & \textbf{37.0} \\
Mamba-3 SISO + FFN & 1.535 & 38.2 & 43.3 & 44.5 & 44.7 & 53.6 \\
Mamba-3 MIMO + FFN & 1.527 & 30.5 & 33.7 & 34.3 & 34.5 & 67.1 \\
Transformer++ & 1.536 & 53.9 & 52.4 & 32.1 & 12.1 & 38.0 \\
\bottomrule
\end{tabular}\par
\end{revblockr}
\end{table}

\subsection[Decode latency, throughput and memory]{Decode latency, throughput and memory}
\label{app:sys-decode-res}

Decode latency is the time for one synchronous step across the entire batch. Aggregate throughput is the number of sequences divided by that step time. \revr{With the same compile policy for every model, ESSH and Mamba-2 are level on A100 at batch one (3.03 and 3.04\,ms). From batch 16 on, ESSH has the lowest latency, 32.58 against 36.43\,ms for Mamba-2 at batch 512.} Step time grows much more slowly than batch size, allowing the same model weights and operations to serve more generated tokens per invocation. \revr{Across batch sizes, Mamba-3 takes 1.15--1.63 times ESSH's latency, and Transformer++ takes 1.12 times as long at batch one and 4.24 times at batch 128.}

Within ESSH, $T_1$ roughly halves batch-one latency relative to $T_{10}$, but the benefit is smaller at batch 128. Truncation removes spectral-channel and FFN computation. The layer sequence, residual width, embeddings, and window attention remain. These retained operations set a common cost across tiers, so the 69\% parameter reduction produces a smaller latency reduction. The measured family is consequently useful for choosing both model size and response time.

The memory decomposition makes the same resource allocation explicit. Recurrent state shrinks from 3.28 to 0.82\,MB, but the window KV cache remains 50.33\,MB. Most model-storage savings from truncation therefore come from weights, while the fixed cache reserves direct context access at every tier. The fixed window KV cache supplies the same local access route as spectral-channel count and feed-forward width decrease. 

{\color{revred}
At the same 1.531B shape with both paths compiled, the fused decoder takes 3.04\,ms versus 3.64\,ms on A100 at batch one and a 2K cache (Table~\ref{tab:sys-decode}). This is a $1.20\times$ speedup under the common compilation policy.
\par}

\begin{table}[htbp]
\centering\footnotesize
\setlength{\tabcolsep}{3pt}
\caption{A100 whole-model decode latency (ms/step), with state/cache sized for a 2K context. \revr{Every model compiles its whole decode step with torch.compile and replays it with CUDA graphs, with the same merged gate/up FFN projection. Full-model rows are the three-session medians of Table~\ref{tab:rev-sym-decode}, and compact tiers use the same compiled ESSH step.} ESSH weights/KV are bf16 and recurrent state fp32. Configurations: Table~\ref{tab:sys-matched-config}. Bold compares parameter-matched full models and the paired fusion control; OOM denotes out of memory. \revr{The fused-versus-unfused control compiles both execution paths of the same 1.531B model.}}
\label{tab:sys-decode}
\label{tab:exp-efficiency}
\begin{tabular*}{\linewidth}{@{\extracolsep{\fill}}lrrrrr@{}}
\toprule
Model & Params (B) & $B=1$ & 16 & 128 & 512 \\
\midrule
\ours{}, $T_{1}$ & 0.474 & \revr{1.47} & \revr{2.76} & \revr{7.03} & \revr{23.56} \\
\ours{}, $T_{2}$ & 0.575 & \revr{1.68} & \revr{2.69} & \revr{7.27} & \revr{25.08} \\
\ours{}, $T_{3}$ & 0.716 & \revr{1.98} & \revr{3.02} & \revr{8.06} & \revr{25.79} \\
\ours{}, $T_{4}$ & 0.828 & \revr{2.18} & \revr{3.14} & \revr{7.97} & \revr{27.27} \\
\ours{}, $T_{5}$ & 0.931 & \revr{2.21} & \revr{3.21} & \revr{8.17} & \revr{28.23} \\
\ours{}, $T_{6}$ & 1.076 & \revr{2.42} & \revr{3.40} & \revr{8.86} & \revr{28.94} \\
\ours{}, $T_{7}$ & 1.168 & \revr{2.66} & \revr{3.47} & \revr{9.11} & \revr{30.62} \\
\ours{}, $T_{8}$ & 1.291 & \revr{2.75} & \revr{3.68} & \revr{9.72} & \revr{31.58} \\
\ours{}, $T_{9}$ & 1.416 & \revr{2.97} & \revr{4.00} & \revr{9.28} & \revr{32.21} \\
\midrule
\ours{}, $T_{10}$ & 1.531 & \revr{\textbf{3.03}} & \revr{\textbf{3.92}} & \revr{\textbf{10.14}} & \revr{\textbf{32.58}} \\
Mamba-2 + FFN & 1.526 & \revr{3.04} & \revr{4.19} & \revr{10.74} & \revr{36.43} \\
Mamba-3 SISO + FFN & 1.535 & \revr{3.48} & \revr{5.06} & \revr{14.64} & \revr{50.37} \\
Mamba-3 MIMO + FFN & 1.527 & \revr{3.74} & \revr{5.41} & \revr{15.67} & \revr{53.19} \\
Transformer++ & 1.536 & \revr{3.40} & \revr{9.59} & \revr{43.03} & OOM \\
\midrule
\multicolumn{6}{l}{\revr{Fused-versus-unfused control, both paths compiled}}\\
\revr{ESSH fused} & \revr{1.531} & \revr{\textbf{3.04}} & \revr{\textbf{3.92}} & \revr{\textbf{10.14}} & \revr{\textbf{32.52}} \\
\revr{ESSH unfused} & \revr{1.531} & \revr{3.64} & \revr{4.58} & \revr{10.18} & \revr{33.44} \\
\bottomrule
\end{tabular*}
\end{table}

\begin{table}[htbp]
\centering
\footnotesize
\setlength{\tabcolsep}{1.7pt}
\caption{A100 memory at the stated model sizes. State/cache totals exclude weights; prefill peaks include weights, logits, and temporaries. ESSH state is fp32 and cache bf16. MB/GB are decimal. Full-model allocations use the matched A100 measurements; compact rows use the tier sweep. Bold compares state+KV totals among the parameter-matched full models.}
\label{tab:sys-memory}
\begin{tabular*}{\linewidth}{@{\extracolsep{\fill}}lrcccccccc@{}}
\toprule
 &  & \multicolumn{3}{c}{$n_{\mathrm b}{=}1$ (MB) $\downarrow$} & total $n_{\mathrm b}{=}128$ $\downarrow$ & \multicolumn{4}{c}{prefill peak (GB) $\downarrow$} \\
\cmidrule(lr){3-5}\cmidrule(lr){7-10}
Model & Params (B) & state & KV & total & (GB) & 2K & 8K & 32K & 128K \\
\midrule
\ours{}, $T_1$ & 0.474 & 0.82 & 50.33 & 51.15 & 6.55 & 1.5 & 3.1 & 9.6 & 35.7 \\
\ours{}, $T_4$ & 0.828 & 1.84 & 50.33 & 52.17 & 6.68 & 2.2 & 3.8 & 10.3 & 36.4 \\
\ours{}, $T_7$ & 1.168 & 2.56 & 50.33 & 52.89 & 6.77 & 3.0 & 4.6 & 11.1 & 37.1 \\
\midrule
\ours{}, $T_{10}$ & 1.531 & 3.28 & 50.33 & 53.61 & 6.86 & 3.6 & 5.2 & 11.7 & 37.8 \\
Mamba-2 + FFN & 1.526 & 30.33 & 0.00 & \textbf{30.33} & \textbf{3.88} & 3.6 & 5.3 & 11.8 & 37.8 \\
Mamba-3 SISO + FFN & 1.535 & 59.64 & 0.00 & 59.64 & 7.63 & 3.6 & 5.3 & 11.8 & 37.8 \\
Mamba-3 MIMO + FFN & 1.527 & 61.01 & 0.00 & 61.01 & 7.81 & 3.6 & 5.2 & 11.7 & 37.8 \\
Transformer++ & 1.536 & 0.00 & 469.99 & 469.99 & 60.16 & 3.6 & 5.2 & 11.8 & 37.8 \\
\bottomrule
\end{tabular*}
\end{table}

\revsubsection{Decode timing under one compile policy}{app:rev-sym-decode}
\begin{revblockr}
Table~\ref{tab:rev-sym-decode} gives the A100 comparison of the 1.53B configurations under that policy: torch.compile and CUDA graphs, the trained SWA layout (blocks 15/22/27), a 2K cache, and the same merged gate/up FFN projection. Each cell is the median of three sessions, and the other settings follow Table~\ref{tab:sys-matched-config}.

Under this policy, ESSH and Mamba-2 take 3.03 and 3.04 ms per step at batch one. From batch 16 on, ESSH has the lowest step time, 32.58 against 36.43 ms for Mamba-2 at batch 512. Mamba-3 is slower at every batch size, and Transformer++ rises from 3.40 ms at batch one to 43.03 ms at batch 128 and runs out of memory at batch 512. The compiled kernel counts of the two recurrent models are also similar (339 for ESSH and 309 for Mamba-2). Merging the gate/up projection accounts for 1--8\% of the baseline step at batch 16 and below and for under 1\% at larger batches (lower block of the table). Moving the window-attention layers from blocks 9/19/27 to the trained blocks 15/22/27 changes A100 latency by less than 0.1\% at every tested batch size.
\end{revblockr}

\begin{table}[htbp]
\begin{revblockr}
\centering\footnotesize
\setlength{\tabcolsep}{3pt}
\revcaption{A100 decode latency (ms/step) of the 1.53B configurations with torch.compile and CUDA graphs for every model, trained SWA layout 15/22/27, and a 2K cache, with medians of three sessions. Kernels are compiled kernel launches per step at batch 1. OOM: out of memory.}{tab:rev-sym-decode}
\begin{tabular*}{\linewidth}{@{\extracolsep{\fill}}lrrrrr@{}}
\toprule
& \multicolumn{4}{c}{ms per step, lower is better} & \\
\cmidrule(lr){2-5}
Model & $B=1$ & 16 & 128 & 512 & Kernels \\
\midrule
\multicolumn{6}{l}{Matched protocol, merged gate/up FFN projection for every model}\\
\ours{} & 3.03 & 3.92 & 10.14 & 32.58 & 339 \\
Mamba-2 + FFN & 3.04 & 4.19 & 10.74 & 36.43 & 309 \\
Mamba-3 SISO + FFN & 3.48 & 5.06 & 14.64 & 50.37 & 617 \\
Mamba-3 MIMO + FFN & 3.74 & 5.41 & 15.67 & 53.19 & 675 \\
Transformer++ & 3.40 & 9.59 & 43.03 & OOM & 338 \\
\midrule
\multicolumn{6}{l}{Matched protocol, baselines with unmerged gate/up projections}\\
Mamba-2 + FFN & 3.20 & 4.53 & 10.81 & 36.57 & -- \\
Mamba-3 SISO + FFN & 3.66 & 5.38 & 14.69 & 50.52 & -- \\
Mamba-3 MIMO + FFN & 3.96 & 5.76 & 15.78 & 53.31 & -- \\
Transformer++ & 3.52 & 9.69 & 43.16 & OOM & -- \\
\bottomrule
\end{tabular*}
\end{revblockr}
\end{table}

\begin{revblockr}
Prefill under the same compile policy (Table~\ref{tab:rev-sym-prefill}) ranks the models differently from decoding. The full ESSH model prefills faster than Mamba-3 MIMO at every length and faster than Transformer++ from 32K tokens on, while Mamba-2 and Mamba-3 SISO prefill faster. The $T_1$ export prefills 2K tokens in about the time Mamba-2 needs.
\end{revblockr}

\begin{table}[htbp]
\begin{revblockr}
\centering\footnotesize
\setlength{\tabcolsep}{4pt}
\revcaptionr{A100 prefill latency (ms, batch one, a forward pass that returns all positions and the full vocabulary) of the 1.53B configurations with torch.compile for every model, with session medians. All rows are shape benchmarks with randomly initialized weights. ESSH uses the trained 1.5B configuration and its tier shapes, with parameter-matched baselines.}{tab:rev-sym-prefill}
\begin{tabular*}{\linewidth}{@{\extracolsep{\fill}}lrrrr@{}}
\toprule
& \multicolumn{4}{c}{prompt length} \\
\cmidrule(lr){2-5}
Model & 2K & 8K & 32K & 128K \\
\midrule
\ours{} $T_1$ (0.474B) & 35.91 & 138.49 & 536.20 & 2131.95 \\
\ours{} $T_4$ (0.828B) & 44.65 & 163.35 & 629.24 & 2500.77 \\
\ours{} $T_7$ (1.168B) & 51.46 & 183.24 & 716.43 & 2849.87 \\
\ours{} $T_{10}$ (1.531B) & 58.78 & 207.85 & 806.56 & 3210.96 \\
Mamba-2 + FFN (1.526B) & 37.04 & 124.72 & 491.05 & 1974.51 \\
Mamba-3 SISO + FFN (1.535B) & 53.60 & 189.35 & 736.08 & 2933.52 \\
Mamba-3 MIMO + FFN (1.527B) & 67.09 & 242.85 & 955.76 & 3801.60 \\
Transformer++ (1.536B) & 37.97 & 156.34 & 1021.37 & 10838.97 \\
\bottomrule
\end{tabular*}
\end{revblockr}
\end{table}

\begin{revblockr}
Under the same protocol, ESSH has the lowest measured decode latency on H200 and B300 at every tested batch size (Table~\ref{tab:rev-xdev-decode}). Its advantage over Mamba-2 grows with batch size, from $1.05$ to $1.15\times$ on H200 and from $1.12$ to $1.45\times$ on B300. Transformer++ falls behind as its KV cache grows ($5.1\times$ at batch 128 on H200, $6.6\times$ at batch 512 on B300).
\end{revblockr}

\begin{table}[htbp]
\begin{revblockr}
\centering\footnotesize
\setlength{\tabcolsep}{3pt}
\revcaptionr{H200 and B300 decode latency (ms/step) of the 1.53B configurations under the protocol of Table~\ref{tab:rev-sym-decode}: torch.compile and CUDA graphs for every model, merged gate/up FFN projection, trained SWA layout 15/22/27, a 2K cache, cuDNN attention disabled for every model. Medians of at least three sessions. Kernels are compiled kernel launches per step at batch 1. OOM: out of memory.}{tab:rev-xdev-decode}
\begin{tabular*}{\linewidth}{@{\extracolsep{\fill}}lrrrrr@{}}
\toprule
Model & $B=1$ & 16 & 128 & 512 & Kernels \\
\midrule
\multicolumn{6}{l}{H200}\\
\ours{} & 1.65 & 2.13 & 5.45 & 17.63 & 367 \\
Mamba-2 + FFN & 1.73 & 2.39 & 6.13 & 20.20 & 366 \\
Mamba-3 SISO + FFN & 2.13 & 2.94 & 6.87 & 21.75 & 675 \\
Mamba-3 MIMO + FFN & 2.35 & 3.24 & 7.69 & 24.05 & 731 \\
Transformer++ & 1.99 & 4.62 & 27.96 & OOM & 366 \\
\midrule
\multicolumn{6}{l}{B300}\\
\ours{} & 1.35 & 1.87 & 4.52 & 12.21 & 367 \\
Mamba-2 + FFN & 1.52 & 2.16 & 5.63 & 17.71 & 338 \\
Mamba-3 SISO + FFN & 1.94 & 2.52 & 5.53 & 15.80 & 647 \\
Mamba-3 MIMO + FFN & 2.19 & 2.83 & 5.84 & 16.30 & 703 \\
Transformer++ & 1.62 & 4.39 & 21.43 & 80.78 & 366 \\
\bottomrule
\end{tabular*}
\end{revblockr}
\end{table}

\begin{revblockr}
The same protocol was applied at 370M (Table~\ref{tab:rev-370-eff}). In the 22-layer, width-512 shell of Appendix~\ref{app:rev-essm}, ESSH decodes faster than Mamba-3 + SWA and Transformer++ at every batch size on A100, H200 and B300. The released ES-SSM implementation evaluates fixed 2048-tap filters by FFT convolution and provides no recurrent decoding interface. Among the elastic families at full size, ESSH decodes faster than MatFormer and MatMamba. The head-nested Mamba-3 + SWA of Table~\ref{tab:rev-aligned-m3} is slower than ESSH at matched sizes while holding more state.
\end{revblockr}

\begin{table}[htbp]
\begin{revblockr}
\centering\footnotesize
\setlength{\tabcolsep}{3pt}
\revcaptionr{370M decode under the protocol of Table~\ref{tab:rev-xdev-decode}: torch.compile and CUDA graphs for every model, a 2K cache and session medians, with single-session batch-128 entries for the elastic families on B300. (a) Models at matched size and common depth 22 and width 512, with window attention at layers 12, 17 and 21 for the recurrent hybrids and full attention for Transformer++. The released ES-SSM implementation uses FFT convolution and provides no recurrent decoding interface. (b) Elastic families at full size and the head-nested Mamba-3 + SWA at matched sizes, in the shell of Table~\ref{tab:main-370-tiers}. State+KV is per sequence at 2K. Cells give A100 / H200 / B300.}{tab:rev-370-eff}
\begin{tabular*}{\linewidth}{@{\extracolsep{\fill}}lrrrr@{}}
\toprule
& & \multicolumn{2}{c}{Decode (ms)} & \\
Model & Params (M) & $B{=}1$ & $B{=}128$ & State+KV (MB) \\
\midrule
\multicolumn{5}{@{}l}{(a) Four mixers in one shell} \\
\ours{} & 370.8 & 1.35 / 0.84 / 0.80 & 3.17 / 1.80 / 1.54 & 13.21 \\
Mamba-3 + SWA & 371.4 & 1.68 / 1.17 / 1.17 & 5.04 / 2.87 / 2.27 & 22.70 \\
Transformer++ & 371.1 & 1.52 / 1.00 / 0.94 & 11.07 / 6.42 / 5.09 & 92.27 \\
ES-SSM (original mixer) & 370.5 & \multicolumn{2}{c}{not implemented in released code} & -- \\
\midrule
\multicolumn{5}{@{}l}{(b) Elastic families and the head-nested control} \\
\ours{}, $T_{10}$ & 370.9 & 1.34 / 0.88 / 0.83 & 4.10 / -- / 2.05 & 23.11 \\
MatFormer, full & 365.6 & 1.51 / 1.04 / 1.01 & 18.41 / -- / 9.01 & 161.56 \\
MatMamba, full & 369.9 & 1.52 / 1.45 / 1.42 & 7.91 / -- / 4.53 & 26.05 \\
Head-nested M3 + SWA, $T_{10}$ & 372.3 & 1.63 / 1.26 / 1.24 & -- & 39.73 \\
\ours{}, $T_{1}$ & 92.1 & 0.94 / -- / 0.71 & 3.38 / -- / 1.78 & 22.29 \\
Head-nested M3 + SWA, $T_{1}$ & 92.3 & 1.23 / 1.00 / 1.09 & -- & 23.92 \\
\bottomrule
\end{tabular*}
\end{revblockr}
\end{table}

\revsubsection{ES-SSM in the same shell and recipe}{app:rev-essm}
\begin{revblockr}
We compare ESSH, the original ES-SSM mixer and Mamba-3 at a common depth of 22 layers, residual width 512, and sliding-window attention layout at layers 12, 17 and 21. A Transformer++ reference uses full attention at every layer. FFN widths are adjusted to match total parameter counts within 0.3\%. All models use 3.7B tokens, the same data and optimizer, and a common evaluator. The released ES-SSM mixer applies fixed 2048-tap Hankel filters by FFT convolution. Table~\ref{tab:rev-essm-dense} reports seed 197 for every model and seed 313 where available. In the shared seed, ESSH has 2.15 lower PPL than ES-SSM, with higher seven-task mean and full-key accuracy. 

\editblue{ESSH modulates retention, writing and readout with the input, whereas the original ES-SSM mixer applies a fixed temporal filter. The joint LTI control in Table~\ref{tab:rev-lti} supports the contribution of this selectivity within the ESSH architecture.}

\end{revblockr}

\begin{table}[htbp]
\begin{revblockr}
\centering\footnotesize
\setlength{\tabcolsep}{5pt}
\revcaptionr{Dense models at matched size (22 layers, width 512, adjusted FFN widths, 3.7B tokens, raw end-of-pretraining checkpoints, common evaluator). The recurrent hybrids use window attention at layers 12, 17 and 21. Transformer++ uses full attention at every layer. PPL, seven-task mean (\%) and teacher-forced full-key accuracy on the 2K NIAH test. Seeds 197 / 313, with one seed for ES-SSM.}{tab:rev-essm-dense}
\begin{tabular*}{\linewidth}{@{\extracolsep{\fill}}lrccc@{}}
\toprule
Mixer & Params (M) & PPL $\downarrow$ & Seven-task $\uparrow$ & Full-key $\uparrow$ \\
\midrule
\ours{} & 370.8 & 16.95 / 16.87 & 40.7 / 40.3 & 0.20 / 0.31 \\
ES-SSM (original mixer) & 370.5 & 19.10 & 39.0 & 0.13 \\
Mamba-3 + SWA & 371.4 & 16.77 / 16.68 & 40.8 / 41.0 & 0.14 / 0.22 \\
Transformer++ & 371.1 & 18.85 / 18.74 & 39.2 / 39.0 & 0.37 / 0.30 \\
\bottomrule
\end{tabular*}
\end{revblockr}
\end{table}

\revsubsection{Ranked post-hoc recovery at $T_1$ under an equal budget}{app:rev-t1-recovery}
\begin{revblockr}
Table~\ref{tab:rev-t1-recovery} compares ranked post-hoc recovery with the jointly trained family at $T_1$ under one recovery budget. Each arm uses the same 370M architecture and 3.7B-token pretraining budget. The two post-hoc arms cut the dense parent to the $T_1$ shape (92.1M active parameters) by Taylor joint-importance ranking or by FFN-activation ranking, and the third arm takes the $T_1$ tier of the jointly trained ESSH model before shared refinement. All three then receive the same 0.5B tokens of $T_1$-only recovery with distillation from their own full-size parent. The ranked cuts reach PPL 26.78 and 27.73, and the jointly trained tier reaches 22.97, 3.8--4.8 PPL lower with the same budget. Without any $T_1$-specific recovery, the ESSH tier after the standard all-tier refinement is already at 25.32--25.38, below either ranked cut after its recovery.
\end{revblockr}

\begin{table}[htbp]
\begin{revblockr}
\centering\footnotesize
\setlength{\tabcolsep}{4pt}
\revcaptionr{Ranked post-hoc cuts of a dense 370M parent and the jointly trained ESSH tier at the $T_1$ shape (92.1M active parameters), after the same 0.5B tokens of $T_1$-only recovery with distillation from the full-size parent (3.7B-token parents, common evaluator, seed 197).}{tab:rev-t1-recovery}
\begin{tabular*}{\linewidth}{@{\extracolsep{\fill}}lcc@{}}
\toprule
$T_1$ model & PPL $\downarrow$ & Seven-task (\%) $\uparrow$ \\
\midrule
Dense parent, Taylor-ranked cut & 26.78 & 36.2 \\
Dense parent, FFN-activation-ranked cut & 27.73 & 35.6 \\
Jointly trained \ours{} tier & \textbf{22.97} & \textbf{37.1} \\
\bottomrule
\end{tabular*}
\end{revblockr}
\end{table}

\subsection[Where the decode advantage comes from]{Where the decode advantage comes from}
\label{app:sys-why}

\revr{Under one compile policy, the decode comparisons show different execution costs across architectures. Against Transformer++, the full-context KV cache grows with batch size. On B300, its latency ratio to ESSH grows from 1.20 at batch one to 4.74 at batch 128 and 6.62 at batch 512. Against Mamba-3, the compiled ESSH step launches fewer kernels, 367 per step against 647 for SISO and 703 for MIMO on B300, and runs 1.2--1.6 times faster.}

\revr{ESSH and Mamba-2 contain 1.531B and 1.526B parameters, and their compiled steps launch a similar number of kernels on B300, 367 and 338 per step. ESSH is still faster at every batch size there, by 1.12 times at batch one and 1.45 times at batch 512. Counts are GPU compute-kernel events in a 20-step profile of the compiled step function, divided by 20, and graph latency is timed separately.} Shared channel values are reused across mode updates, and the structured transitions allow the resulting operations to be fused. \revr{With both execution paths compiled and graph-replayed, the fused decoder is 1.20, 1.17 and 1.03 times faster than the unfused path at batch 1, 16 and 512, and level with it at batch 128 (Table~\ref{tab:sys-decode}).}

Table~\ref{tab:sys-matched-config} records the comparison configurations: depth, residual width, vocabulary, and embedding tying are held fixed, and baseline FFN widths are selected before timing to match total parameters within 0.34\%. These architecture timings use randomly initialized weights at the stated tensor shapes. Quality is evaluated from trained models. Table~\ref{tab:sys-small-devices} gives the separate near-size comparison at 370M.

\begin{table}[htbp]
\centering\small
\caption{Configurations for the A100, H200, and B300 comparisons. Parameter counts include tied embeddings. All use vocabulary 128256, residual width 2048, and 28 blocks. Baseline FFN widths are matched in multiples of 64; official Mamba mixer settings stay fixed. Gap is relative to ESSH.}
\label{tab:sys-matched-config}
\begin{tabular*}{\linewidth}{@{\extracolsep{\fill}}lrrrr@{}}
\toprule
Model & Params (M) & Gap (\%) & FFN width & SWA layers\\
\midrule
ESSH & 1531.1 & +0.00 & 5632 & 3 \\
Mamba-2 + FFN & 1526.0 & -0.33 & 3136 & 0 \\
Mamba-3 SISO + FFN & 1534.9 & +0.25 & 3136 & 0 \\
Mamba-3 MIMO + FFN & 1526.7 & -0.29 & 2816 & 0 \\
Transformer++ & 1536.3 & +0.34 & 4672 & 0 \\
\bottomrule
\end{tabular*}
\end{table}

\subsection[Cross-device comparison]{Cross-device comparison}
\label{app:sys-portability}

Appendix~\ref{app:cross-device-final} reports A100, H200, and B300 measurements, with the consolidated efficiency table at the end of the appendix.

\section[Reproducibility]{Reproducibility}
\label{app:reproducibility}

This section collects the configurations, export procedure, evaluation environment, and compute accounting needed to rerun the experiments. The supplementary material contains the experimental code.

\subsection[Configurations and hyperparameters]{Configurations and hyperparameters}
\label{app:repro-config}

Table~\ref{tab:repro-hparams} collects the recipe constants. The launch configurations are in Appendix~\ref{app:setup-config}. Ablation tables state deviations.

\begin{table}[htbp]
\centering
\small
\setlength{\tabcolsep}{3pt}
\caption{Constants used in the main recipe and the experiment sections that evaluate alternative settings.}
\label{tab:repro-hparams}
\begin{tabular}{@{}p{0.23\linewidth}p{0.25\linewidth}p{0.18\linewidth}p{0.25\linewidth}@{}}
\toprule
{\raggedright Constant\par} & {\raggedright 370M\par} & {\raggedright 1.5B\par} & {\raggedright fixed by\par} \\
\midrule
{\raggedright $\bar K$, $m$\par} & {\raggedright 32, 8\par} & {\raggedright 32, 8\par} & {\raggedright divisibility, Appendices~\ref{app:sel-kbar}, \ref{app:init-r}\par} \\
{\raggedright window layers, window\par} & {\raggedright 3 late, 2048\par} & {\raggedright 3 late, 2048\par} & {\raggedright Appendix~\ref{app:setup-config}\par} \\
{\raggedright channel / feed-forward exponents\par} & {\raggedright 0.5 / 0.8\par} & {\raggedright same\par} & {\raggedright Appendix~\ref{app:el-nesting}\par} \\
{\raggedright full share $f$, slice scale $\alpha_{\mathrm{ff}}$\par} & {\raggedright 0.75, 0.5\par} & {\raggedright 0.75, 0.5\par} & {\raggedright Appendix~\ref{app:el-constants}\par} \\
{\raggedright distillation objectives and coefficients\par} & {\raggedright KL at 0.5 (pretraining), TV at 5 (final stage)\par} & {\raggedright same\par} & {\raggedright Appendix~\ref{app:el-constants}\par} \\
{\raggedright pretraining capacities\par} & {\raggedright $1/16$, $3/32$, $1/8$, $5/32$, $3/16$, $1/4$, $3/8$, $1/2$, $3/4$, $1$\par} & {\raggedright same\par} & {\raggedright design\par} \\
{\raggedright ordering period, window\par} & {\raggedright 500 steps, until $80\%$\par} & {\raggedright same\par} & {\raggedright Appendix~\ref{app:elastic-sort}\par} \\
{\raggedright clock bias, gate init\par} & {\raggedright $-3$, gates at one\par} & {\raggedright same\par} & {\raggedright Appendix~\ref{app:mixer-proj}\par} \\
{\raggedright zero-lag coefficient initialization\par} & {\raggedright 0.5, trainable\par} & {\raggedright same\par} & {\raggedright Appendix~\ref{app:mixer-decay}\par} \\
{\raggedright peak lr, warm-up, floor\par} & {\raggedright $1.2\times10^{-3}$, $1\%$, $10\%$\par} & {\raggedright $1.0\times10^{-3}$, $1\%$, $10\%$\par} & {\raggedright Appendix~\ref{app:el-constants}\par} \\
{\raggedright AdamW betas, wd, clip\par} & {\raggedright (0.9, 0.95), 0.1, 1.0\par} & {\raggedright same\par} & {\raggedright Mamba-2 recipe\par} \\
{\raggedright tokens per step, context\par} & {\raggedright 262,144, 2048\par} & {\raggedright 524,288, 2048\par} & {\raggedright design\par} \\
{\raggedright final stage\par} & {\raggedright TV 5, 2060 steps, 131,072 tokens/step (0.27B), lr $6\times10^{-5}$\par} & {\raggedright TV 5, 4120 steps, same tokens/step (0.54B), same lr\par} & {\raggedright training configuration\par} \\
{\raggedright SR capacities\par} & {\raggedright $\{2,4,8,16,24\}/32$, one per step\par} & {\raggedright $T_1$--$T_9$, shuffled across micro-batches\par} & {\raggedright Appendix~\ref{app:elastic-export}\par} \\
\bottomrule
\end{tabular}
\end{table}

The shared refinement objective is defined in Appendix~\ref{app:elastic-export}. Unlike the full-tier distillation step, this stage includes the full model's label loss as well as the reduced-capacity model's label and distillation losses.

\subsection[Export procedure]{Export procedure}
\label{app:repro-export}

Given a trained full model, instantiate each tier using Table~\ref{tab:el-map}. Slice its tensors as described in Appendix~\ref{app:elastic-slice} and load them into the standalone model with strict key matching. Verify the parameter count and compare its outputs with those of the shared model at the same tier.

\subsection[Evaluation harness and environment]{Evaluation harness and environment}
\label{app:repro-env}

\revr{The matched A100 and H200 inference measurements use PyTorch 2.11.0+cu128 and CUDA 12.8. The B300 measurements use PyTorch 2.11.0+cu130 and CUDA 13.0. All three use Triton 3.6.0~\citep{paszke2019pytorch,tillet2019triton}. The seven-task evaluator and metric selection are specified in Appendix~\ref{app:setup-eval}.}

\label{app:repro-seeds}
Training-seed ranges and item-cluster bootstrap intervals describe different uncertainty sources. The former compare independently trained models. The latter resample the key pairs within a fixed needle evaluation. Neither substitutes for the other.

\subsection[Compute accounting]{Compute accounting}
\label{app:repro-compute}

\revr{Table~\ref{tab:repro-compute} itemizes 4509 raw GPU-hours for the reported experiments, summed across devices without hardware normalization.}

\begin{table}[!htbp]
\centering
\footnotesize
\setlength{\tabcolsep}{4pt}
\caption{Reconstructed raw GPU-hours for \revr{the itemized experiments}. Totals include all runs for these experiments and exclude unreported exploration. Device hours are summed without hardware normalization and rounded to integers.}
\label{tab:repro-compute}
\begin{tabular*}{\linewidth}{@{\extracolsep{\fill}}p{0.39\linewidth}p{0.43\linewidth}r@{}}
\toprule
Stage & Devices & GPU-hours \\
\midrule
1.5B main training & A100, H100, H200, B200 & 1808 \\
1.5B final stage, export and evaluation & H200, RTX 5090/3090 & 47 \\
370M shared recipe, 7.4B & A100, H100, RTX 5090/3090 & 104 \\
370M comparisons, 7.4B & A100, H200, RTX 5090/3090 & 772 \\
370M ablations, 3.7B & A100, RTX 5090/3090 & 1633 \\
Synthetic tasks and DNA & A100, RTX 5090/3090 & 129 \\
Systems measurements & A100, H200, B300 & 16 \\
\midrule
Total & & 4509 \\
\bottomrule
\end{tabular*}
\end{table}

\FloatBarrier
\section[Cross-Device Efficiency]{Cross-Device Efficiency}
\label{app:cross-device-final}

The 1.53B systems comparisons use 28-layer, width-2048 configurations on A100-SXM4-80GB, NVIDIA H200, and NVIDIA B300. \revr{The ESSH decode and prefill comparisons on all three GPUs use the trained window-attention layout at blocks 15, 22 and 27 (Table~\ref{tab:setup-config}).}

\revr{Every model is compiled with torch.compile, with the same merged gate/up FFN projection.} Decode uses CUDA graphs after a 2048-token context, with 50 warm-ups and 200 timed steps per session. Matched full-model results on all three devices and B300 compact batch-one results are medians of \revr{at least} three sessions. Prefill uses three warm-ups and five timed calls per session. Tables identify single-session compact measurements where retained. The matched H200/B300 runs use the same attention-backend policy for all models, with cuDNN SDPA disabled.

Ratios in the text use unrounded measurements.

\subsection[H200 single-GPU measurements]{H200 single-GPU measurements}
\label{app:sys-h200}

\revr{The parameter-matched H200 comparison repeats the decode result on a second GPU generation. At batch one, ESSH takes 1.65\,ms, against 1.73 for Mamba-2, 2.13 for Mamba-3 SISO, 2.35 for Mamba-3 MIMO, and 1.99 for Transformer++. ESSH is the fastest full model at every batch size, 1.05--1.15 times faster than Mamba-2 and 1.23--1.52 times faster than Mamba-3. Transformer++ takes 1.20 times as long at batch one and 5.13 times at batch 128, and it runs out of memory at batch 512.}

The prefill ordering changes with input length. \revr{Transformer++ is faster at 2K and 8K, while ESSH sustains 76.0 ktok/s at 128K versus 19.2, a $3.96\times$ advantage.} The chunked parallel scan and fixed-window attention require linear work as context length grows, consistent with this sustained throughput. Mamba-2 reaches \revr{154.0} ktok/s at 128K, retaining its strength in parallel prefill, while ESSH has the lowest recurrent decode latency. These measurements distinguish the execution benefits of the two workloads.

Cache size becomes a feasibility constraint at large batches. Transformer++ encounters allocation failure at batch 512 in all three sessions: its 2K KV cache would require 240.6\,GB, exceeding the 150.0\,GB (139.7\,GiB) reported by the driver. ESSH completes that setting at 17.63\,ms with 27.45\,GB of state+KV. Keeping attention in three window layers therefore reserves a substantially smaller per-sequence cache while the remaining layers use recurrent state. Capacity-dependent timing is reported for A100 and B300 under the protocols specified in their tables.

\begin{table}[htbp]
\begin{revblockr}
\centering\footnotesize
\setlength{\tabcolsep}{3pt}
\caption{H200 decode latency (ms/step), with a 2K cache. Every model compiles its whole decode step with torch.compile and replays it with CUDA graphs. All five models are parameter-matched, use the same merged gate/up FFN projection and disable cuDNN attention, with medians of three sessions. ESSH state is fp32 and weights/KV bf16, and baselines retain native state precision. OOM: out of memory. Bold ranks within each group.}
\label{tab:sys-h200}
\begin{tabular*}{\linewidth}{@{\extracolsep{\fill}}lrrrrr@{}}
\toprule
Model & Params (B) & $B=1$ & 16 & 128 & 512 \\
\midrule
\multicolumn{6}{l}{Parameter-matched full models}\\
\ours{}, $T_{10}$ & 1.531 & \textbf{1.65} & \textbf{2.13} & \textbf{5.45} & \textbf{17.63} \\
Mamba-2 + FFN & 1.526 & 1.73 & 2.39 & 6.13 & 20.20 \\
Mamba-3 SISO + FFN & 1.535 & 2.13 & 2.94 & 6.87 & 21.75 \\
Mamba-3 MIMO + FFN & 1.527 & 2.35 & 3.24 & 7.69 & 24.05 \\
Transformer++ & 1.536 & 1.99 & 4.62 & 27.96 & OOM \\
\bottomrule
\end{tabular*}
\end{revblockr}
\end{table}

\begin{table}[htbp]
\begin{revblockr}
\centering\footnotesize
\setlength{\tabcolsep}{3pt}
\caption{H200 batch-one full-logit prefill with torch.compile for every model. Throughput is in ktok/s, latency in ms and peak allocation in decimal GB. All five models use three sessions with three warm-ups and five timed calls per session. cuDNN SDPA is disabled for every model. Bold compares throughput and latency among the parameter-matched full models.}
\label{tab:sys-h200-prefill}
\begin{tabular*}{\linewidth}{@{\extracolsep{\fill}}lrrrrrrrr@{}}
\toprule
 &  & \multicolumn{4}{c}{Prefill ktok/s $\uparrow$} & ms $\downarrow$ & \multicolumn{2}{c}{Peak GB $\downarrow$} \\
Model & Params (B) & 2K & 8K & 32K & 128K & 2K & 2K & 128K \\
\midrule
\multicolumn{9}{l}{Parameter-matched full models}\\
\ours{}, $T_{10}$ & 1.531 & 72.9 & 75.7 & 76.6 & 76.0 & 28.1 & 3.64 & 37.79 \\
Mamba-2 + FFN & 1.526 & 108.5 & \textbf{156.3} & \textbf{155.4} & \textbf{154.0} & 18.9 & 3.65 & 37.80 \\
Mamba-3 SISO + FFN & 1.535 & 94.7 & 104.9 & 106.1 & 106.5 & 21.6 & 3.65 & 37.80 \\
Mamba-3 MIMO + FFN & 1.527 & 55.6 & 56.7 & 57.3 & 56.1 & 36.8 & 3.64 & 37.79 \\
Transformer++ & 1.536 & \textbf{142.4} & 113.5 & 59.6 & 19.2 & \textbf{14.4} & 3.64 & 37.27 \\
\bottomrule
\end{tabular*}
\end{revblockr}
\end{table}

\subsection[B300 decoding across the ten tiers]{B300 decoding across the ten tiers}
\label{app:sys-b300-decode}

\revr{At near-equal parameter counts and with one compile policy, ESSH has the lowest decode latency at every measured batch size (Table~\ref{tab:sys-b300-decode}). Batch-one latency is 1.35\,ms, compared with 1.52 for Mamba-2, 1.94 for Mamba-3 SISO, 2.19 for Mamba-3 MIMO, and 1.62 for Transformer++.} All use the same depth, residual width, vocabulary, and timing code, and parameter counts differ from ESSH by at most 0.34\%.

The gain is present even though ESSH stores slightly more weights than Mamba-2, 3.062 versus 3.052\,GB. \revr{The two compiled steps also launch a similar number of kernels, 367 and 338 per step, while the Mamba-3 steps launch 647 (SISO) and 703 (MIMO).} The spectral computation shares each projected channel value across its modes and combines state updates and readouts in the fused recurrent path.

\revr{The relative advantage grows with batch size. ESSH is 1.12 times faster than Mamba-2 at batch one and 1.45 times faster at batch 512, where it generates 41.9 ktok/s against 32.4 for the fastest baseline, Mamba-3 SISO.} Batching shares fixed model work across more generated tokens, so the comparison moves from short-response latency toward sustained aggregate throughput. ESSH remains favorable in both regimes.

\revr{Within the ESSH family, batch-one latency stays between 1.13 and 1.53\,ms across the ten tiers, and removing about 69\% of the parameters changes it from 1.35\,ms at $T_{10}$ to 1.25\,ms at $T_1$. At batch 512 the step time falls from 12.21 to 10.76\,ms.} The unchanged backbone and attention operations remain in every tier, so parameter savings exceed the measured latency savings.

\revr{The $T_7$ and $T_{10}$ profiles both record 367 compute kernels per step at batch one, but use different matrix-multiplication tilings. This is consistent with shape-dependent kernel efficiency at small batch sizes. At batches 128 and 512, $T_7$ is faster than $T_{10}$ (Table~\ref{tab:sys-b300-decode}).}

\begin{table}[htbp]
\begin{revblockr}
\centering\footnotesize
\setlength{\tabcolsep}{3pt}
\caption{B300 decode latency (ms/step), after a 2K context. Every model compiles its whole decode step with torch.compile and replays it with CUDA graphs. Full-model entries and compact $B=1$ entries are medians of at least three sessions. Compact entries at batch 16 and above are single-session measurements at $T_1$, $T_4$ and $T_7$. ESSH state is fp32 and weights/KV are bf16. Mamba rows use the official mixers with a residual FFN block and the same merged gate/up projection. Bold compares the parameter-matched full models.}
\label{tab:sys-b300-decode}
\begin{tabular*}{\linewidth}{@{\extracolsep{\fill}}lrrrrr@{}}
\toprule
Model & Params (B) & $B=1$ & 16 & 128 & 512 \\
\midrule
\ours{}, $T_{1}$ & 0.474 & 1.25 & 1.52 & 3.56 & 10.76 \\
\ours{}, $T_{2}$ & 0.575 & 1.13 & -- & -- & -- \\
\ours{}, $T_{3}$ & 0.716 & 1.30 & -- & -- & -- \\
\ours{}, $T_{4}$ & 0.828 & 1.34 & 1.63 & 3.79 & 11.43 \\
\ours{}, $T_{5}$ & 0.931 & 1.31 & -- & -- & -- \\
\ours{}, $T_{6}$ & 1.076 & 1.38 & -- & -- & -- \\
\ours{}, $T_{7}$ & 1.168 & 1.53 & 1.90 & 4.25 & 11.90 \\
\ours{}, $T_{8}$ & 1.291 & 1.48 & -- & -- & -- \\
\ours{}, $T_{9}$ & 1.416 & 1.47 & -- & -- & -- \\
\midrule
\ours{}, $T_{10}$ & 1.531 & \textbf{1.35} & \textbf{1.87} & \textbf{4.52} & \textbf{12.21} \\
Mamba-2 + FFN & 1.526 & 1.52 & 2.16 & 5.63 & 17.71 \\
Mamba-3 SISO + FFN & 1.535 & 1.94 & 2.52 & 5.53 & 15.80 \\
Mamba-3 MIMO + FFN & 1.527 & 2.19 & 2.83 & 5.84 & 16.30 \\
Transformer++ & 1.536 & 1.62 & 4.39 & 21.43 & 80.78 \\
\bottomrule
\end{tabular*}
\end{revblockr}
\end{table}

\FloatBarrier
\subsection[B300 prefill across tiers]{B300 prefill across tiers}
\label{app:sys-b300-prefill}

The prefill comparison changes with context length. \revr{Transformer++ leads at 2K and remains faster than ESSH at 8K, but its throughput falls to 26.0 ktok/s at 128K while ESSH sustains 113.6, a $4.36\times$ advantage.} Both models keep the same parameter count along their curves. The changed ordering therefore occurs as input length increases the full-attention workload, while the chunked parallel scan and fixed-window attention require work that grows linearly with context length.

The recurrent baselines also sustain their rates, with Mamba-2 reaching \revr{226.0} ktok/s at 128K. Prefill computes many positions together, whereas decode repeatedly advances one state. ESSH's lowest decode latency and Mamba-2's highest prefill throughput describe different strengths under these two execution patterns.

Compact ESSH tiers remove channel and FFN work at every input position, but prefill peak memory decreases by a smaller proportion. The returned bf16 logits alone occupy 33.6\,GB at 128K, independently of tier. Against that common output cost, the $T_{10}$--$T_1$ peak difference of \revr{2.09}\,GB closely follows their 2.11\,GB difference in weight storage. Keeping output allocation separate from state+KV explains why throughput, model size, and prefill peak memory have different relative changes.

\begin{table}[htbp]
\begin{revblockr}
\centering\small
\setlength{\tabcolsep}{3pt}
\caption{B300 batch-one full-logit prefill with torch.compile for every model. Throughput is in ktok/s, latency in ms and peak allocation in decimal GB. Full models use at least three sessions, with three warm-ups and five timed calls per session. Compact rows are single-session measurements. Full-model parameter counts differ by at most $0.34\%$ from ESSH. Bold compares throughput and latency among the parameter-matched full models.}
\label{tab:sys-b300-prefill}
\begin{tabular*}{\linewidth}{@{\extracolsep{\fill}}lrrrrrrrr@{}}
\toprule
 &  & \multicolumn{4}{c}{Prefill ktok/s $\uparrow$} & ms $\downarrow$ & \multicolumn{2}{c}{Peak GB $\downarrow$} \\
\cmidrule(lr){3-6}\cmidrule(lr){8-9}
Model & Params (B) & 2K & 8K & 32K & 128K & 2K & 2K & 128K \\
\midrule
\ours{}, $T_{1}$ & 0.474 & 127.6 & 133.2 & 136.8 & 135.1 & 16.0 & 1.52 & 35.68 \\
\ours{}, $T_{4}$ & 0.828 & 113.6 & 125.1 & 128.4 & 126.5 & 18.0 & 2.24 & 36.39 \\
\ours{}, $T_{7}$ & 1.168 & 107.4 & 117.5 & 121.7 & 120.1 & 19.1 & 2.90 & 37.05 \\
\midrule
\ours{}, $T_{10}$ & 1.531 & 102.0 & 112.0 & 114.7 & 113.6 & 20.1 & 3.62 & 37.77 \\
Mamba-2 + FFN & 1.526 & 131.5 & \textbf{224.6} & \textbf{232.3} & \textbf{226.0} & 15.6 & 3.63 & 37.78 \\
Mamba-3 SISO + FFN & 1.535 & 107.3 & 119.8 & 122.5 & 121.5 & 19.1 & 3.63 & 37.78 \\
Mamba-3 MIMO + FFN & 1.527 & 57.4 & 55.9 & 56.5 & 56.2 & 35.7 & 3.61 & 37.76 \\
Transformer++ & 1.536 & \textbf{224.1} & 180.6 & 83.3 & 26.0 & \textbf{9.1} & 3.62 & 37.24 \\
\bottomrule
\end{tabular*}
\end{revblockr}
\end{table}
\FloatBarrier
\subsection[370M elastic families across devices]{370M elastic families across devices}
\label{app:sys-small-devices}

The 370M results test whether the deployment advantage persists at a smaller model scale. \revr{With one compile policy for every family, full ESSH decodes faster than the near-size full MatFormer and MatMamba models on A100, H200 and B300. At batch one it takes 1.335, 0.881 and 0.831\,ms on the three devices, against 1.507, 1.043 and 1.014 for MatFormer and 1.522, 1.450 and 1.419 for MatMamba. At batch 128 the gap widens, to 4.10 against 18.41 and 7.91\,ms on A100 and 2.05 against 9.01 and 4.53\,ms on B300.} ESSH also uses less state+KV, 23.11 versus 161.56 and 26.05\,MB (Table~\ref{tab:sys-small-devices}(b)). \revr{Each recorded shape uses at least three timing sessions at batch one after a 2K context, with 50 warm-ups and 200 timed steps per session.}

The models devote similar parameter budgets to different computations. MatFormer uses attention in every block, MatMamba has 48 recurrent blocks, and ESSH uses 22 blocks with three window layers and a fused recurrent path. Fewer recurrent invocations and the fused update are part of the complete ESSH deployment design. The common-depth, width-512 comparison in Table~\ref{tab:rev-370-eff}(a) separately controls for these backbone differences.

Within a family, truncation reduces parameters faster than latency because embeddings, residual dimensions, and the layer sequence remain. The tables therefore pair model size with measured response time at each tier. \revr{The tier sweeps and the H200 full-model comparison compile every family in the same way.}

\begin{table}[!htb]
\begin{revblockr}
\centering\scriptsize
\caption{Batch-one and batch-128 decode at the 370M scale, with a 2K cache. Every family compiles its whole decode step with torch.compile and replays it with CUDA graphs. (a) A100 and B300 tier sweeps followed by the full-model comparison. Batch-128 entries on B300 are single-session measurements. (b) H200 full-model repeat at the stated parameter counts. Other values are medians of at least three sessions. ESSH state is fp32 and KV bf16, and baselines retain native precision. Bold compares full models within each device and batch size. Equal tier labels across families need not have equal sizes.}
\label{tab:sys-small-devices}
\begin{tabular*}{\linewidth}{@{\extracolsep{\fill}}lrrrrrr@{}}
\toprule
\multicolumn{7}{l}{\textbf{(a) Tier sweeps on A100 and B300}}\\
 & & & \multicolumn{2}{c}{Graph ms, $B=1$ $\downarrow$} & \multicolumn{2}{c}{Graph ms, $B=128$ $\downarrow$}\\
\cmidrule(lr){4-5}\cmidrule(lr){6-7}
Family & Tier & Params (M) & A100 & B300 & A100 & B300 \\
\midrule
ESSH & $T_{1}$ & 92.1 & 0.94 & 0.71 & 3.38 & 1.78 \\
ESSH & $T_{2}$ & 118.9 & 0.97 & 0.63 & -- & -- \\
ESSH & $T_{3}$ & 154.7 & 1.00 & 0.79 & -- & -- \\
ESSH & $T_{4}$ & 185.3 & 1.04 & 0.80 & 3.60 & 1.88 \\
ESSH & $T_{5}$ & 214.5 & 1.13 & 0.80 & -- & -- \\
ESSH & $T_{6}$ & 248.8 & 1.10 & 0.86 & -- & -- \\
ESSH & $T_{7}$ & 274.3 & 1.13 & 0.86 & 4.11 & 2.22 \\
ESSH & $T_{8}$ & 308.6 & 1.27 & 0.82 & -- & -- \\
ESSH & $T_{9}$ & 337.9 & 1.27 & 0.84 & -- & -- \\
\midrule
MatFormer & $T_{1}$ & 130.9 & 1.17 & 0.90 & 17.93 & 8.86 \\
MatFormer & $T_{2}$ & 146.0 & 1.23 & 0.93 & -- & -- \\
MatFormer & $T_{3}$ & 168.7 & 1.29 & 0.93 & -- & -- \\
MatFormer & $T_{4}$ & 191.5 & 1.30 & 0.95 & 18.06 & 8.90 \\
MatFormer & $T_{5}$ & 214.2 & 1.35 & 0.97 & -- & -- \\
MatFormer & $T_{6}$ & 248.2 & 1.42 & 0.96 & -- & -- \\
MatFormer & $T_{7}$ & 270.9 & 1.39 & 0.97 & 18.23 & 8.96 \\
MatFormer & $T_{8}$ & 301.2 & 1.42 & 0.99 & -- & -- \\
MatFormer & $T_{9}$ & 331.5 & 1.48 & 1.00 & -- & -- \\
\midrule
MatMamba & $T_{1}$ & 83.3 & 1.18 & 0.92 & 1.86 & 1.51 \\
MatMamba & $T_{2}$ & 102.4 & 1.14 & 1.04 & -- & -- \\
MatMamba & $T_{3}$ & 131.1 & 1.16 & 1.11 & -- & -- \\
MatMamba & $T_{4}$ & 159.7 & 1.37 & 1.15 & 3.91 & 2.65 \\
MatMamba & $T_{5}$ & 188.4 & 1.31 & 1.18 & -- & -- \\
MatMamba & $T_{6}$ & 226.6 & 1.40 & 1.31 & -- & -- \\
MatMamba & $T_{7}$ & 255.3 & 1.45 & 1.34 & 5.77 & 3.48 \\
MatMamba & $T_{8}$ & 293.5 & 1.50 & 1.33 & -- & -- \\
MatMamba & $T_{9}$ & 331.7 & 1.56 & 1.24 & -- & -- \\
\midrule
ESSH & $T_{10}$ & 370.9 & \textbf{1.34} & \textbf{0.83} & \textbf{4.10} & \textbf{2.05} \\
MatFormer & $T_{10}$ & 365.6 & 1.51 & 1.01 & 18.41 & 9.01 \\
MatMamba & $T_{10}$ & 369.9 & 1.52 & 1.42 & 7.91 & 4.53 \\
\bottomrule
\end{tabular*}
\par\smallskip
\begin{tabular*}{\linewidth}{@{\extracolsep{\fill}}lrrr@{}}
\toprule
\multicolumn{4}{l}{\textbf{(b) H200 full models}}\\
Model & Params (M) & Graph ms $\downarrow$ & State+KV (MB) $\downarrow$\\
\midrule
ESSH & 370.9 & \textbf{0.88} & \textbf{23.11} \\
MatFormer & 365.6 & 1.04 & 161.56 \\
MatMamba & 369.9 & 1.45 & 26.05 \\
\bottomrule
\end{tabular*}
\end{revblockr}
\end{table}
\FloatBarrier
\subsection[Consolidated efficiency measurements]{Consolidated efficiency measurements}
\label{app:sys-consolidated}

The consolidated table compares full models at the stated parameter budgets across three GPUs (Table~\ref{tab:sys-portability}). The 1.53B comparison fixes depth and residual width and matches parameter counts within 0.34\%. The 370M comparison uses each trained architecture at a similar parameter count.

\revr{With one compile policy for every model, ESSH has the lowest decode latency in these full-model comparisons on H200 and B300 at every batch size. On A100 it is level with Mamba-2 at batch one and the fastest from batch 16 on.} Prefill has a different ordering: Mamba-2 is strongest among the recurrent models, while ESSH sustains higher long-context throughput than full attention. State+KV reflects another trade-off: ESSH's fixed window KV cache makes its total larger than Mamba-2's recurrent state, but substantially smaller than Transformer++'s full-context cache.

\begin{table}[htbp]
\begin{revblockr}
\centering\footnotesize
\caption{Cross-device efficiency at the stated parameter counts. Every model is compiled with torch.compile in decoding and prefill, and decoding replays the compiled step with CUDA graphs. Graph decode uses a 2K cache and prefill uses batch one. Full-model rows use session medians at matched parameter counts. OOM is measured allocation failure. Bold ranks within each group and device. Memory is decimal. ESSH state is fp32 and KV bf16, with native baseline state precision.}
\label{tab:sys-portability}
\begin{tabular*}{\linewidth}{@{\extracolsep{\fill}}lrrrr@{}}
\toprule
Measurement / model & Params (M) & A100 & H200 & B300 \\
\midrule
\multicolumn{5}{l}{Full models near 370M: graph decode, batch one (ms $\downarrow$)}\\
ESSH & 370.9 & \textbf{1.335} & \textbf{0.881} & \textbf{0.831} \\
MatFormer & 365.6 & 1.507 & 1.043 & 1.014 \\
MatMamba & 369.9 & 1.522 & 1.450 & 1.419 \\\midrule
\multicolumn{5}{l}{Full models near 1.53B: graph decode, batch 1 (ms $\downarrow$)}\\
\ours{}, $T_{10}$ & 1531.1 & \textbf{3.03} & \textbf{1.65} & \textbf{1.35} \\
Mamba-2 + FFN & 1526.0 & 3.04 & 1.73 & 1.52 \\
Mamba-3 SISO + FFN & 1534.9 & 3.48 & 2.13 & 1.94 \\
Mamba-3 MIMO + FFN & 1526.7 & 3.74 & 2.35 & 2.19 \\
Transformer++ & 1536.3 & 3.40 & 1.99 & 1.62 \\\midrule
\multicolumn{5}{l}{Full models near 1.53B: graph decode, batch 128 (ms $\downarrow$)}\\
\ours{}, $T_{10}$ & 1531.1 & \textbf{10.14} & \textbf{5.45} & \textbf{4.52} \\
Mamba-2 + FFN & 1526.0 & 10.74 & 6.13 & 5.63 \\
Mamba-3 SISO + FFN & 1534.9 & 14.64 & 6.87 & 5.53 \\
Mamba-3 MIMO + FFN & 1526.7 & 15.67 & 7.69 & 5.84 \\
Transformer++ & 1536.3 & 43.03 & 27.96 & 21.43 \\\midrule
\multicolumn{5}{l}{Full models near 1.53B: graph decode, batch 512 (ms $\downarrow$)}\\
\ours{}, $T_{10}$ & 1531.1 & \textbf{32.58} & \textbf{17.63} & \textbf{12.21} \\
Mamba-2 + FFN & 1526.0 & 36.43 & 20.20 & 17.71 \\
Mamba-3 SISO + FFN & 1534.9 & 50.37 & 21.75 & 15.80 \\
Mamba-3 MIMO + FFN & 1526.7 & 53.19 & 24.05 & 16.30 \\
Transformer++ & 1536.3 & OOM & OOM & 80.78 \\\midrule
\multicolumn{5}{l}{Full-model prefill at 2K (ktok/s $\uparrow$)}\\
\ours{}, $T_{10}$ & 1531.1 & 34.8 & 72.9 & 102.0 \\
Mamba-2 + FFN & 1526.0 & \textbf{55.3} & 108.5 & 131.5 \\
Mamba-3 SISO + FFN & 1534.9 & 38.2 & 94.7 & 107.3 \\
Mamba-3 MIMO + FFN & 1526.7 & 30.5 & 55.6 & 57.4 \\
Transformer++ & 1536.3 & 53.9 & \textbf{142.4} & \textbf{224.1} \\\midrule
\multicolumn{5}{l}{Full-model prefill at 128K (ktok/s $\uparrow$)}\\
\ours{}, $T_{10}$ & 1531.1 & 40.8 & 76.0 & 113.6 \\
Mamba-2 + FFN & 1526.0 & \textbf{66.4} & \textbf{154.0} & \textbf{226.0} \\
Mamba-3 SISO + FFN & 1534.9 & 44.7 & 106.5 & 121.5 \\
Mamba-3 MIMO + FFN & 1526.7 & 34.5 & 56.1 & 56.2 \\
Transformer++ & 1536.3 & 12.1 & 19.2 & 26.0 \\\midrule
\multicolumn{5}{l}{Persistent state+KV at batch one, 2K (MB $\downarrow$)}\\
\ours{}, $T_{10}$ & 1531.1 & 53.61 & 53.61 & 53.61 \\
Mamba-2 + FFN & 1526.0 & \textbf{30.33} & \textbf{30.33} & \textbf{30.33} \\
Mamba-3 SISO + FFN & 1534.9 & 59.64 & 59.64 & 59.64 \\
Mamba-3 MIMO + FFN & 1526.7 & 61.01 & 61.01 & 61.01 \\
Transformer++ & 1536.3 & 469.99 & 469.99 & 469.99 \\
\bottomrule
\end{tabular*}
\end{revblockr}
\end{table}
\FloatBarrier

\end{document}

%% file: math_commands.tex
\usepackage{amsmath,amsfonts,bm}

\def\1{\bm{1}}

\DeclareMathAlphabet{\mathsfit}{\encodingdefault}{\sfdefault}{m}{sl}
\SetMathAlphabet{\mathsfit}{bold}{\encodingdefault}{\sfdefault}{bx}{n}

\newcommand{\KL}{D_{\mathrm{KL}}}

